\documentclass[journal]{IEEEtran}
\usepackage{color,xcolor}
\usepackage{graphicx}
\usepackage{amsmath}
\usepackage{amssymb}
\usepackage{subfig}
\usepackage{cite}
\usepackage{makecell}
\usepackage{threeparttable}
\usepackage{multirow}
\usepackage{colortbl}

\def\ie{\textit{i.e.}}
\def\eg{\textit{e.g.}}
\def\etc{\textit{etc.}}
\def\et{\textit{et al.}}

\definecolor{mygray}{gray}{.9}
\definecolor{mypink}{rgb}{.99,.91,.95}
\definecolor{mycyan}{cmyk}{.3,0,0,0}
\definecolor{myyellow}{rgb}{.99,.9,0}

\ifCLASSINFOpdf
\else
\fi
\begin{document}
%
\title{Prior-Induced Information Alignment for Image Matting}
%
%
%

\author{Yuhao~Liu$^{*}$, Jiake~Xie$^{*}$,  Yu~Qiao, and~Yong~Tang$^{\dagger}$, Xin~Yang$^{\dagger}$

\thanks{This work was supported in part by the National Natural Science Foundation of China under Grant 91748104, Grant 61972067, the Innovation Technology Funding of Dalian (Project No. 2018J11CY010, 2020JJ26GX036) and PicUP.Ai project of the Winroad Holdings Ltd . }
\thanks{Yuhao Liu, Yu Qiao and Xin Yang are with the School of Computer Science and Technology, Dalian University of Technology, Dalian 116024, China.~E-mail: yuhaoLiu7456@gmail.com, qiaoyu2017@mail.dlut.edu.cn, xinyang@dlut.edu.cn.}
\thanks{Jiake~Xie and Yong~Tang are Computer Vision Algorithm Engineer in Winroad Holdings Limited. E-mail: jxie@win.road, yt@win.road.}
\thanks{Yuhao Liu and Jiake Xie  are the joint first authors. Xin~Yang and Yong~Tang  are the joint corresponding authors.}}

\maketitle

\begin{abstract}
	Image matting is an ill-posed problem that aims to estimate the opacity of foreground pixels in an image. 
	However, most existing deep learning-based methods still suffer from the coarse-grained details.
   In general, these algorithms are incapable of felicitously distinguishing the  degree of exploration  between deterministic domains (\eg~certain FG and BG pixels) and undetermined domains (\eg~uncertain in-between pixels),
   or inevitably lose information in the continuous sampling process, leading to a sub-optimal result.	
	In this paper, we propose a novel network named Prior-Induced Information Alignment Matting Network (PIIAMatting), which can efficiently model the distinction of pixel-wise response maps and the correlation of layer-wise feature maps. It mainly consists of a Dynamic Gaussian Modulation mechanism (DGM) and an Information Alignment strategy (IA).
	Specifically, the DGM can dynamically acquire a pixel-wise domain response map learned from the  prior distribution. The response map can present the relationship between the opacity variation and the convergence process during training.
	On the other hand, the IA comprises an Information Match Module (IMM) and an Information Aggregation Module (IAM),  jointly scheduled to  match and aggregate the adjacent layer-wise features adaptively. 
	Besides, we also develop a Multi-Scale Refinement (MSR) module to integrate multi-scale receptive field information at the refinement stage to recover the fluctuating appearance details. 
Extensive quantitative and qualitative evaluations demonstrate that the proposed PIIAMatting performs favourably against state-of-the-art image matting methods on the Alphamatting.com, Composition-1K and Distinctions-646 dataset.
\end{abstract}


%
\IEEEpeerreviewmaketitle

\section{Introduction}
%
%
%
%
\IEEEPARstart{T}{he} digital  matting is one of the essential tasks in computer vision, which aims to accurately estimate the opacity of foreground objects in images and video sequences. It has a wide range of applications, especially in the field of digital image editing and film production. Formally, the input image is modelled as a linear combination of foreground and background colours, as shown below:
\begin{equation}
\label{image_synthesis}
\textit{I}_{i} = \alpha_{i}F_{i}+(1 - \alpha_{i})B_{i}, \quad \alpha_{i}\in[0,1]	
\end{equation}
\begin{figure}[t]
	\centering
	\subfloat[Image]{
		\label{fig:image}
		\begin{minipage}[t]{0.15\textwidth}
			\centering
			\setlength\parindent{-4mm}                     
			\includegraphics[scale=0.064]{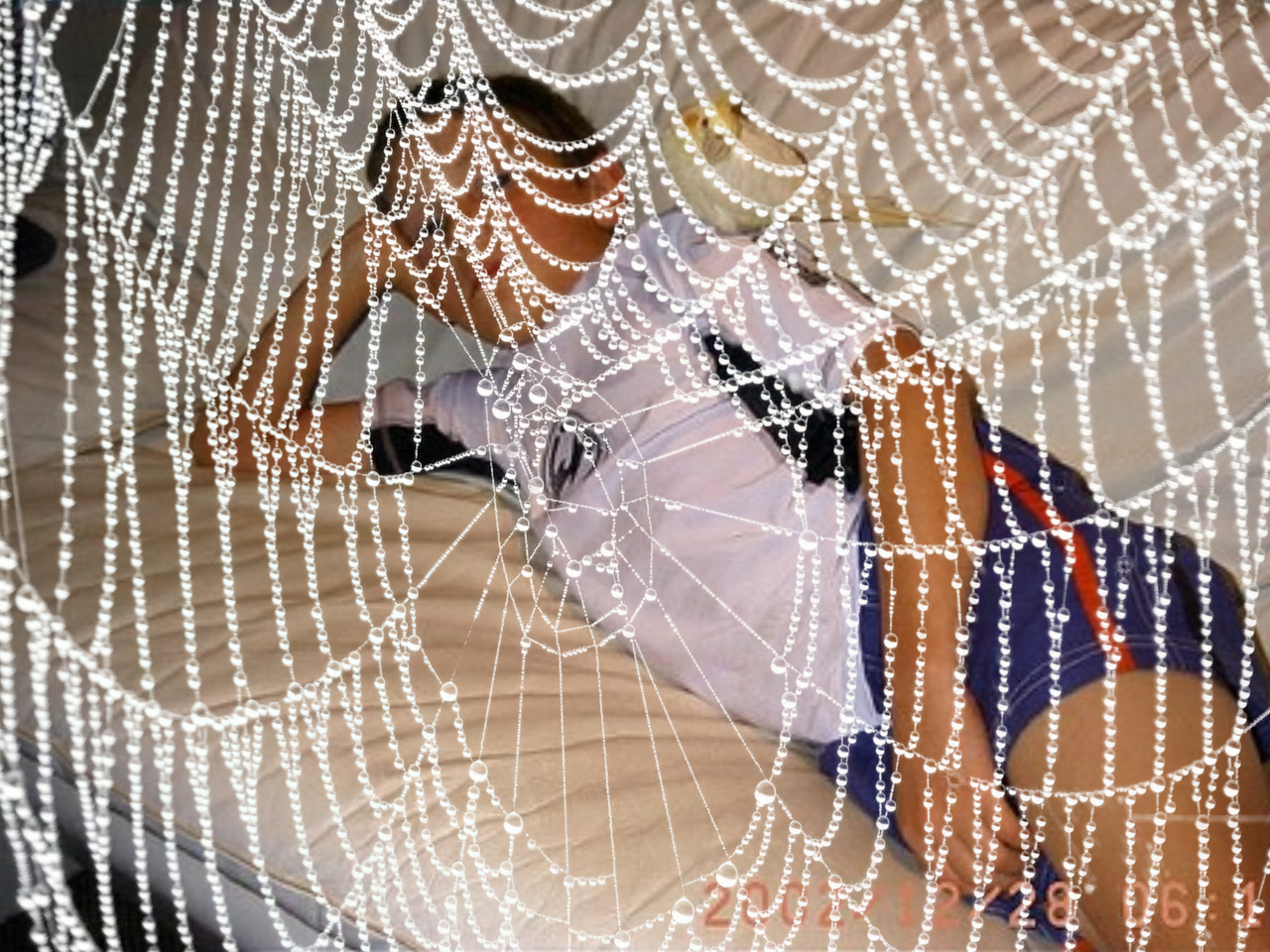}
		\end{minipage}
	}
	\subfloat[Trimap]{
		\label{fig:trimap}
		\begin{minipage}[t]{0.15\textwidth}
			\centering
			\setlength\parindent{-4mm}
			\includegraphics[scale=0.064]{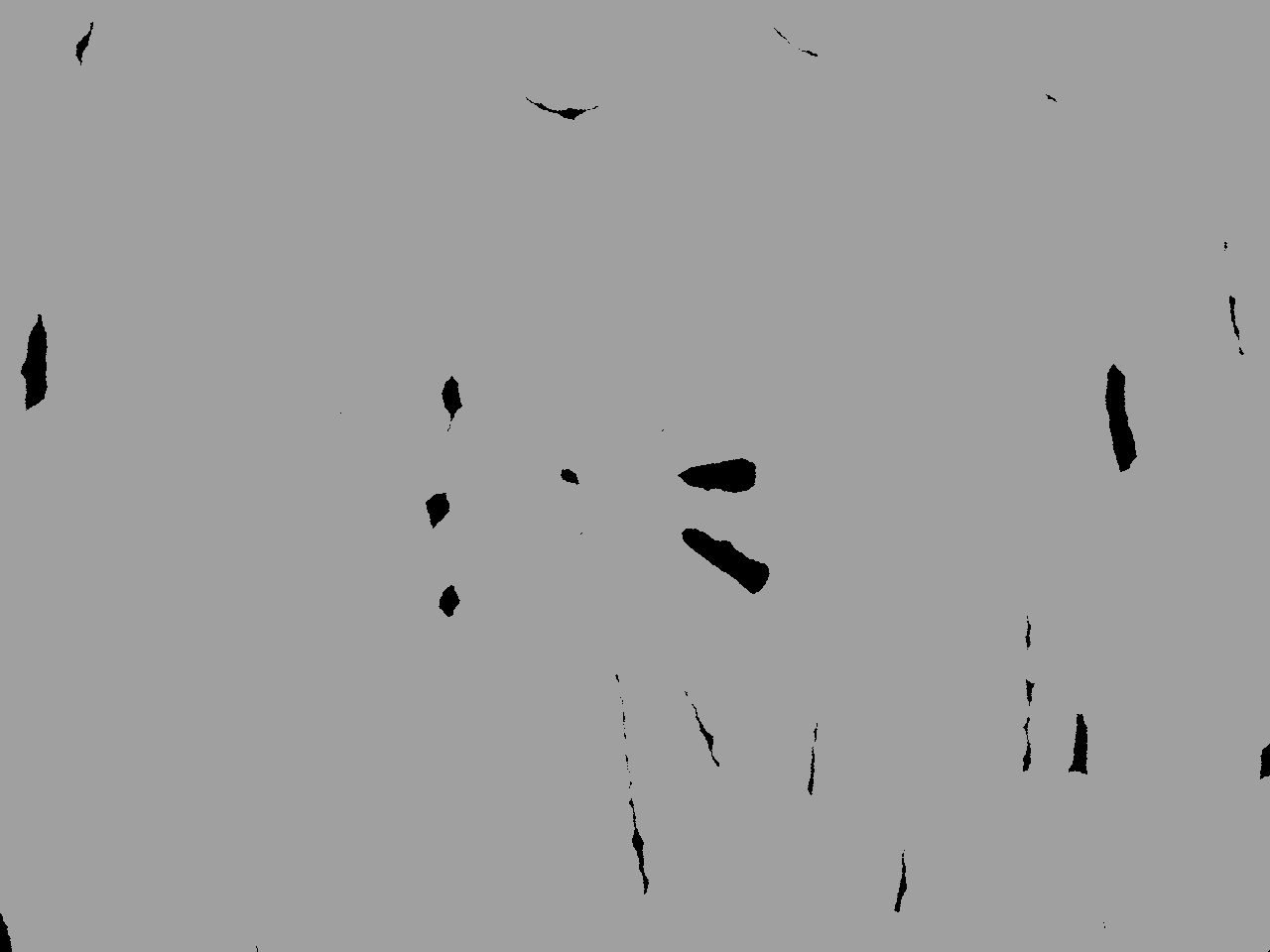}
		\end{minipage}
	}
	\subfloat[DIM\cite{Xu2017Deep}]{
		\label{fig:dim}
		\begin{minipage}[t]{0.15\textwidth}
			\centering
			\setlength\parindent{-4mm}
			\includegraphics[scale=0.064]{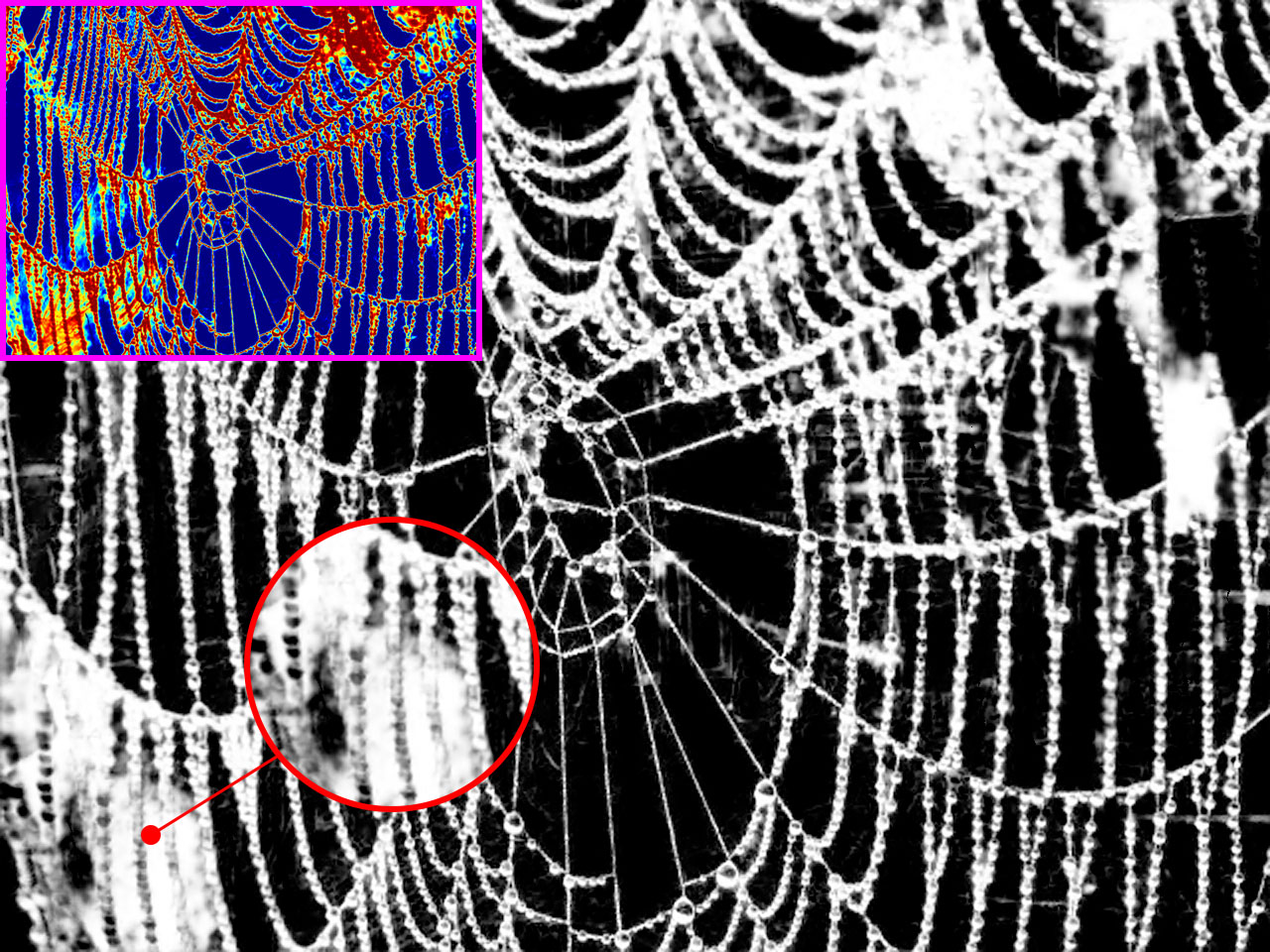}
		\end{minipage}
	}
	\vspace{-1mm}
	\subfloat[IndexNet\cite{lu2020index}]{
		\label{fig:indexnet}
		\begin{minipage}[t]{0.15\textwidth}
			\centering
			\setlength\parindent{-2mm}
			\includegraphics[scale=0.064]{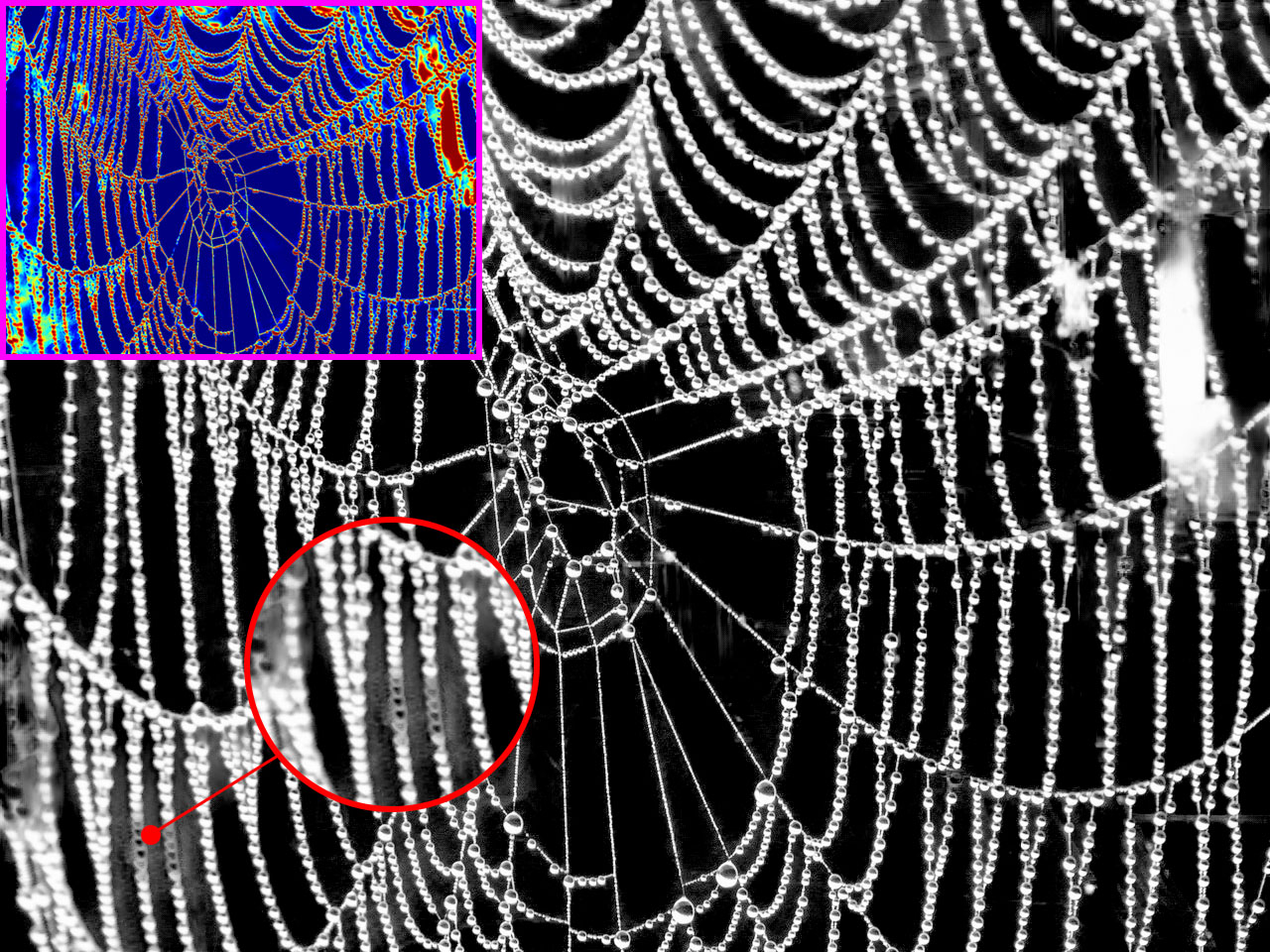}
		\end{minipage}
	}
	\subfloat[GCA\cite{li2020natural}]{
		\label{fig:gca}
		\begin{minipage}[t]{0.15\textwidth}
			\centering
			\setlength\parindent{-2mm}
			\includegraphics[scale=0.064]{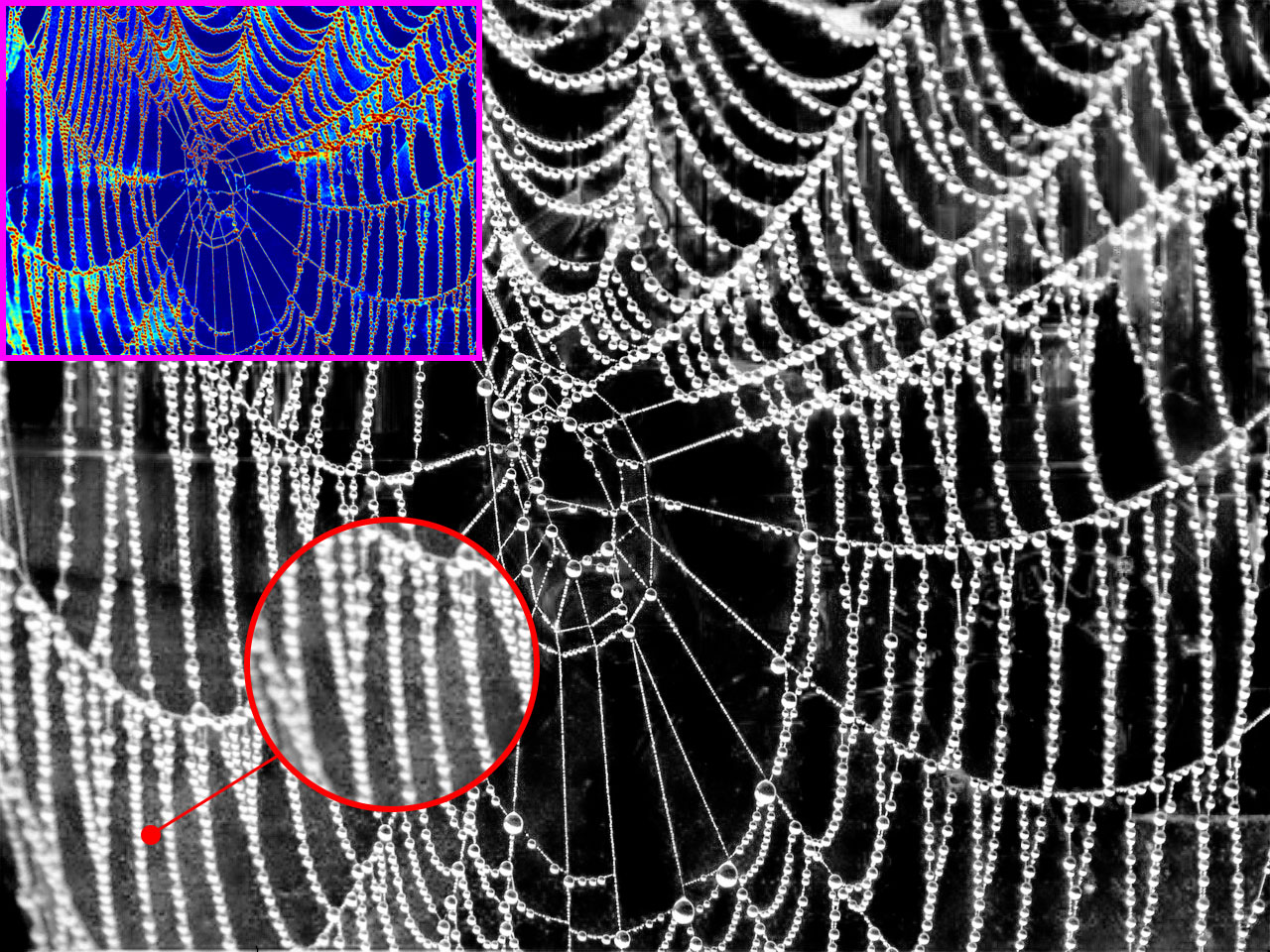}
		\end{minipage}
	}
	\subfloat[PIIAMatting (Ours)]{
		\label{fig:our}
		\begin{minipage}[t]{0.15\textwidth}
			\centering
			\setlength\parindent{-2mm}
			\includegraphics[scale=0.064]{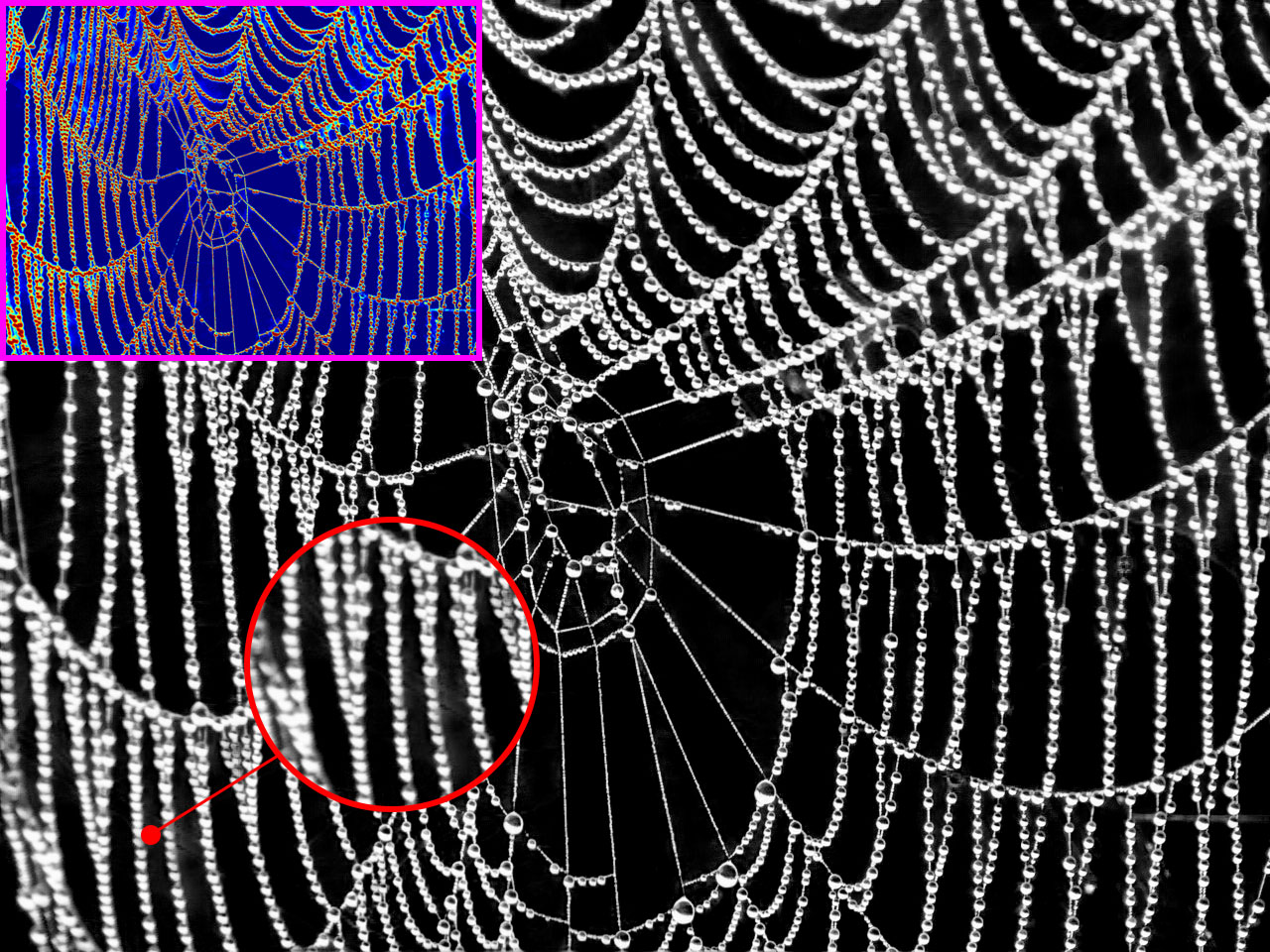}
		\end{minipage}
	}
	\vspace{-1mm}
	\caption{Sample results of our method compared with others. {\color{purple}{Purple}} box: each alpha matte is accompanied by an attention map on the left corner. {\color{red}{Red}} circle: the partial details zoom in.}
	\label{fig:teaser}
	\vspace{-3mm}
\end{figure}
where $i$ refers to the pixel position in the input image $\textit{I}$. $\textit{F}_{i}$, $\textit{B}_{i}$ and $\alpha_{i}$ denote the foreground, background colour, and alpha matte separately. Given an input image I, the image matting algorithms intend to solve
$F$, $B$ and $\alpha$ simultaneously. Based on Eq.~\ref{image_synthesis}, for a pixel of a typical 3-channel (\eg~RGB) input image, 7 unknowns (\ie~3 $F$ values, 3 $B$ values and 1 $alpha$ value)
need to be solved, but there are only 3 known quantities (3 $I$ values). Therefore, the problem is highly ill-posed.

Natural image matting is essentially the computation of the opacity of each pixel in an image
\cite{rhemann2009perceptually,Xu2017Deep}. According to Eq.~\ref{image_synthesis}, the pixels in an image can be divided into three groups where $\alpha_{i}=1$ and $\alpha_{i}=0$ refer to the foreground and background regions, respectively, while $\alpha_{i}\in(0,1)$ 
represent the transition regions (usually the area that needs to be solved in image matting) which can be decoupled to the \textbf{deterministic}  domains (certain foreground  and background pixels) and \textbf{undetermined}  domains (uncertain in-between pixels).
In general, some user-specified constraint information is utilized, such as trimap and scribble, which are helpful to reduce the solution space of the ill-posed problem for solving Eq.~\ref{image_synthesis}. The trimap is composed of three parts, white, black, and grey, indicating the foreground, background, and transition regions separately. While the scribble can be regarded as a simplified version of trimap, which specifies the foreground and background regions using few sparse scribbles. With the trimap as assistance, early works attempted to solve Eq.~\ref{image_synthesis} using the colour distribution, leading to blurred or chunky artefacts\cite{Xu2017Deep}.

Motivated by the success of deep learning, the digital image matting methods based on Deep Convolutional Neural Networks (DCNNs)
 have been proposed. Cho \et \cite{cho2018deep} took the results of \cite{chen2013knn} and \cite{levin2007closed} combined with the RGB colours as input to predict an alpha matte by using the CNN. Then, Deep Image Matting (DIM)\cite{Xu2017Deep} proposed the first large-scale image matting dataset and  trained a model in an end-to-end fashion.
Most of the following methods\cite{tang2019learning, hou2019context, cai2019disentangled, qiao2020attention} endeavoured to design a more complicated network to implicitly constrain Eq.~\ref{image_synthesis} for obtaining high-quality alpha mattes. However, all the above methods suffer two possible limitations: 1). the disproportion between pixel domains. 2). the information discrepancy between the process of sampling. A visual comparison of different methods can be seen in Fig.~\ref{fig:teaser}.

To address the above issue-1, the LFM\cite{zhang2019late} combined two classification networks and a fusion network under the supervision of a hybrid loss to jointly predict alpha mattes with a single RGB image as input. The hybrid loss function intends to calculate the different regions in varied manners, especially  L1 loss for  transition regions and L2 loss for the rest of the regions. 
While this setting can attenuate the disproportion of pixels at the different domains to some extent, it neglects the internal distribution inbalance of the transition regions and  can only be regarded as a static pre-defined partition that does not properly accommodate varieties in opacity.  As for the information discrepancy, the index map generated in indexNet\cite{lu2020index} is similar to an attention map, which can learn the index from specific feature maps and propagate the crucial information to the corresponding upsampling operation in the decoder stage, thereby preventing the loss of details. However, it ignores the complementarity between adjacent layers in information selection and ensemble. 

\begin{figure}[t]
	\setlength{\tabcolsep}{1pt}\small{
		\begin{tabular}{ccc}
			\setlength\parindent{1pt}
			\includegraphics[scale=0.040]{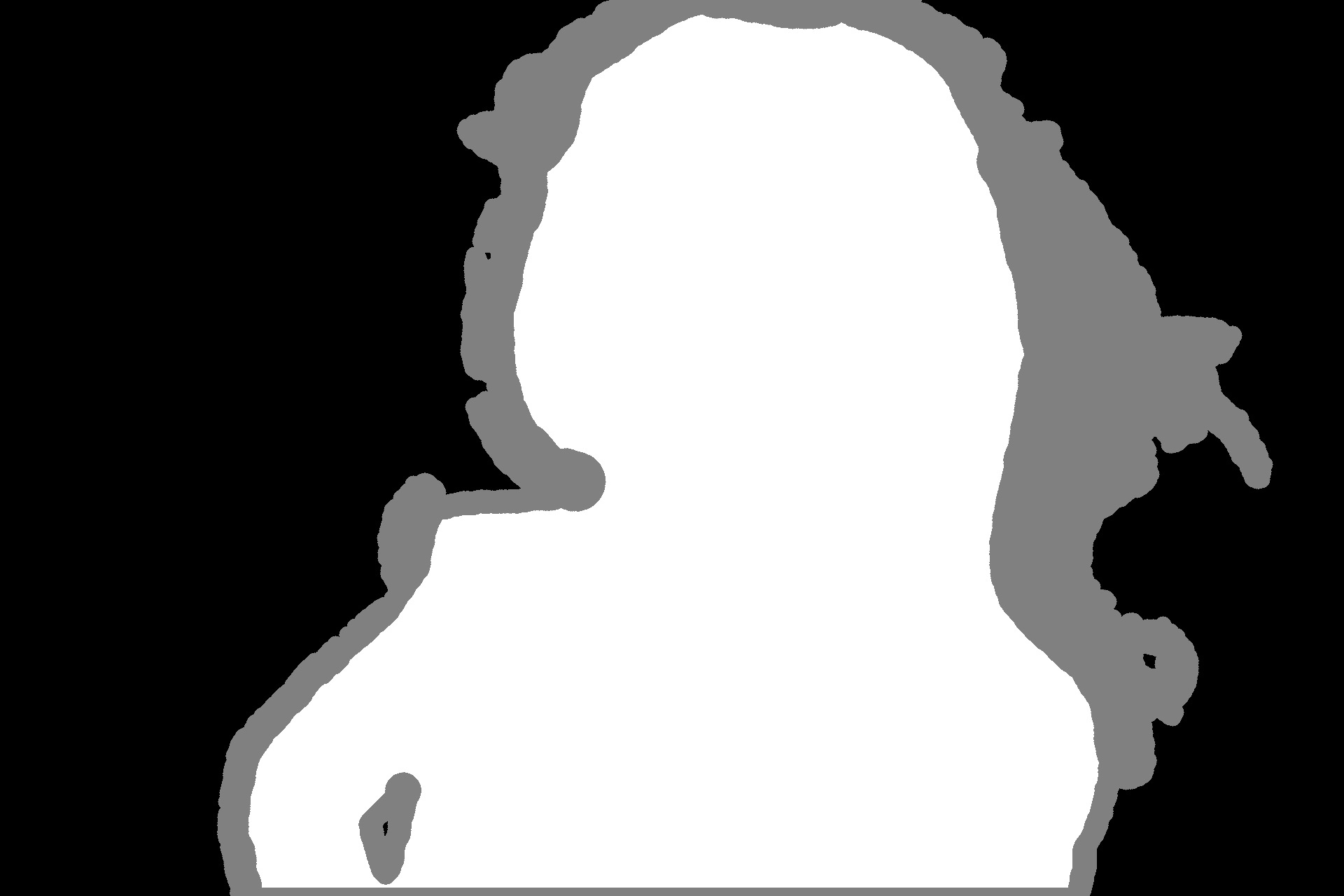} &
			\includegraphics[scale=0.040]{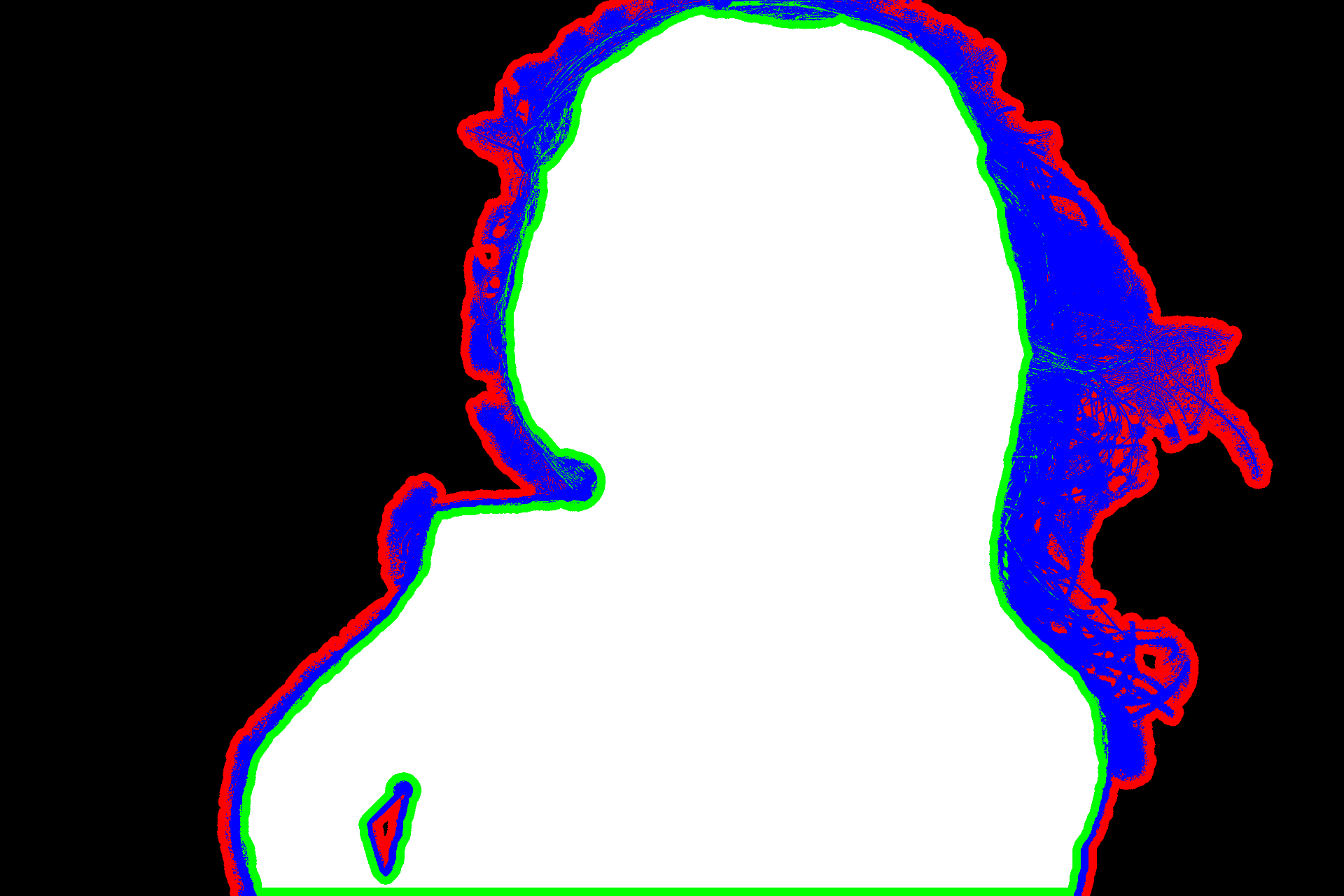} &
			\includegraphics[scale=0.040]{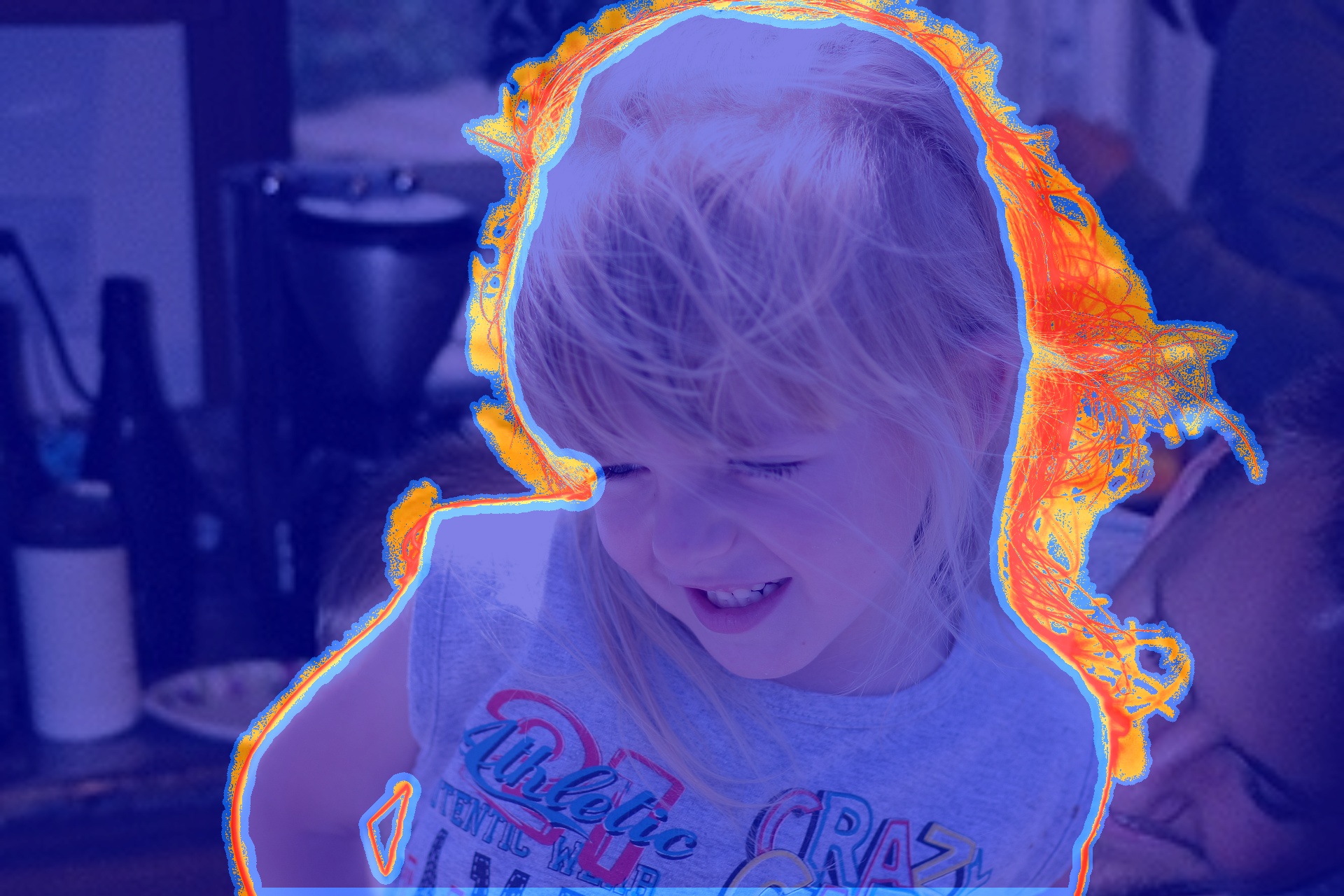}
			\\
			\setlength\parindent{1pt}
			\includegraphics[scale=0.040]{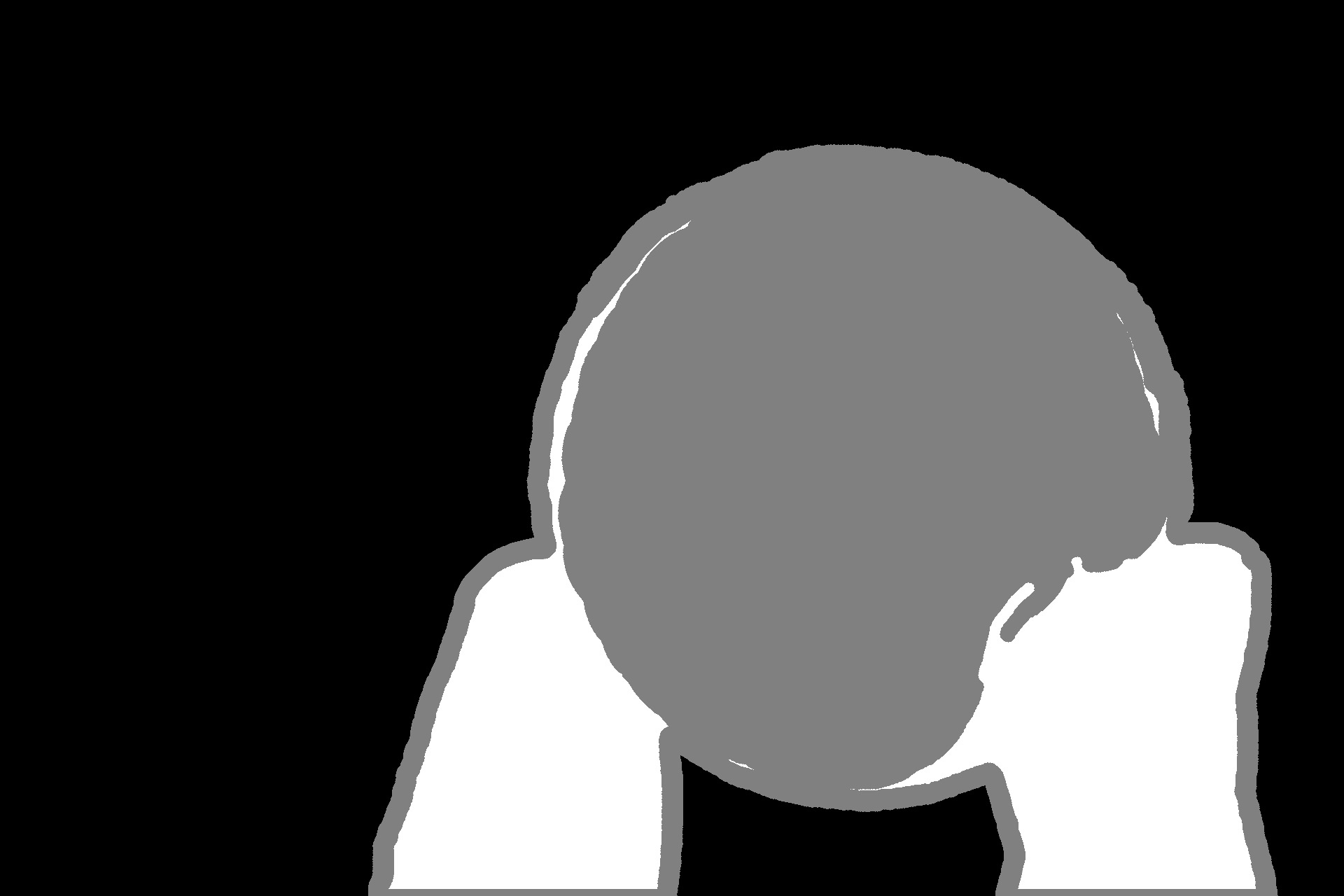} &
			\includegraphics[scale=0.040]{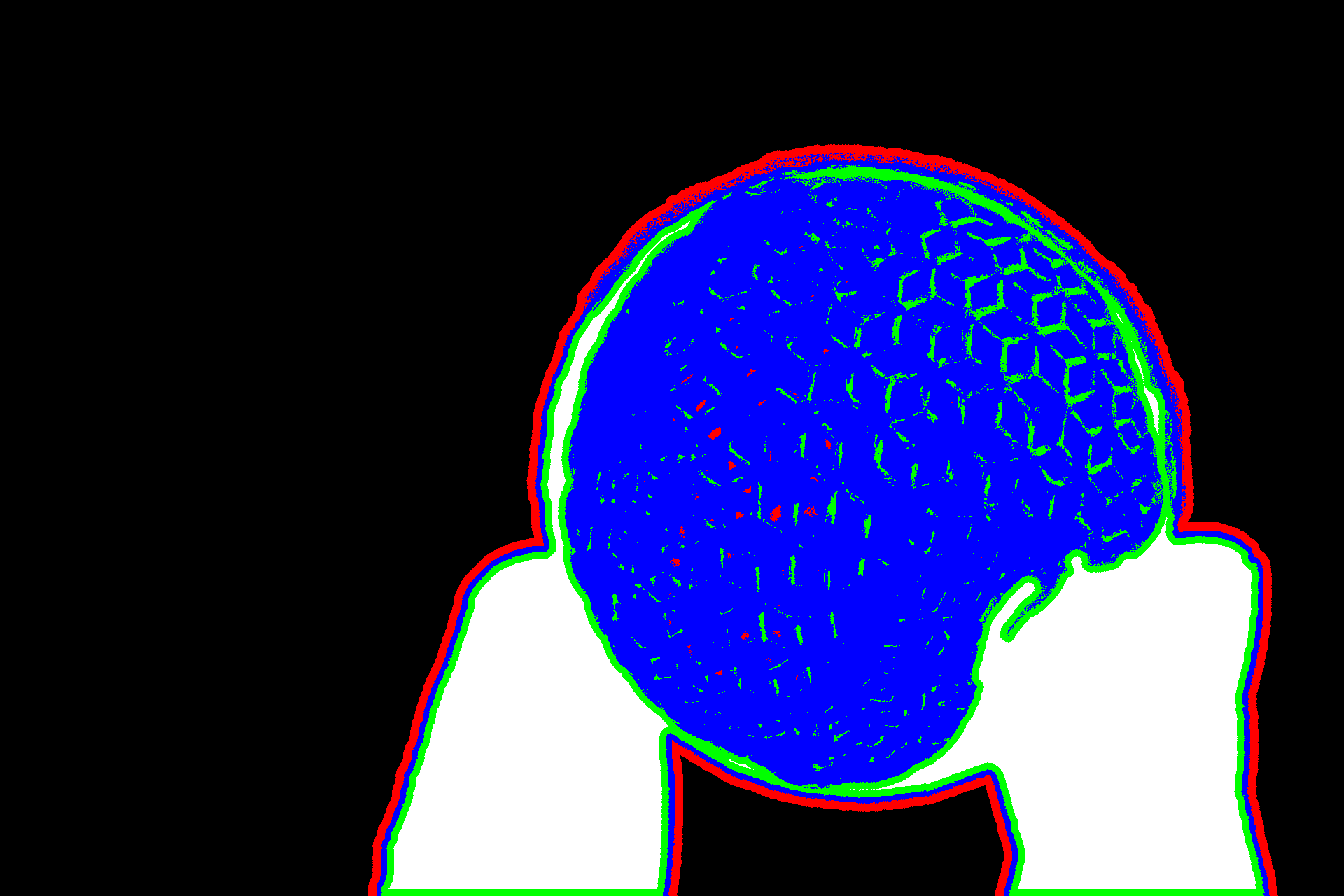} &
			\includegraphics[scale=0.040]{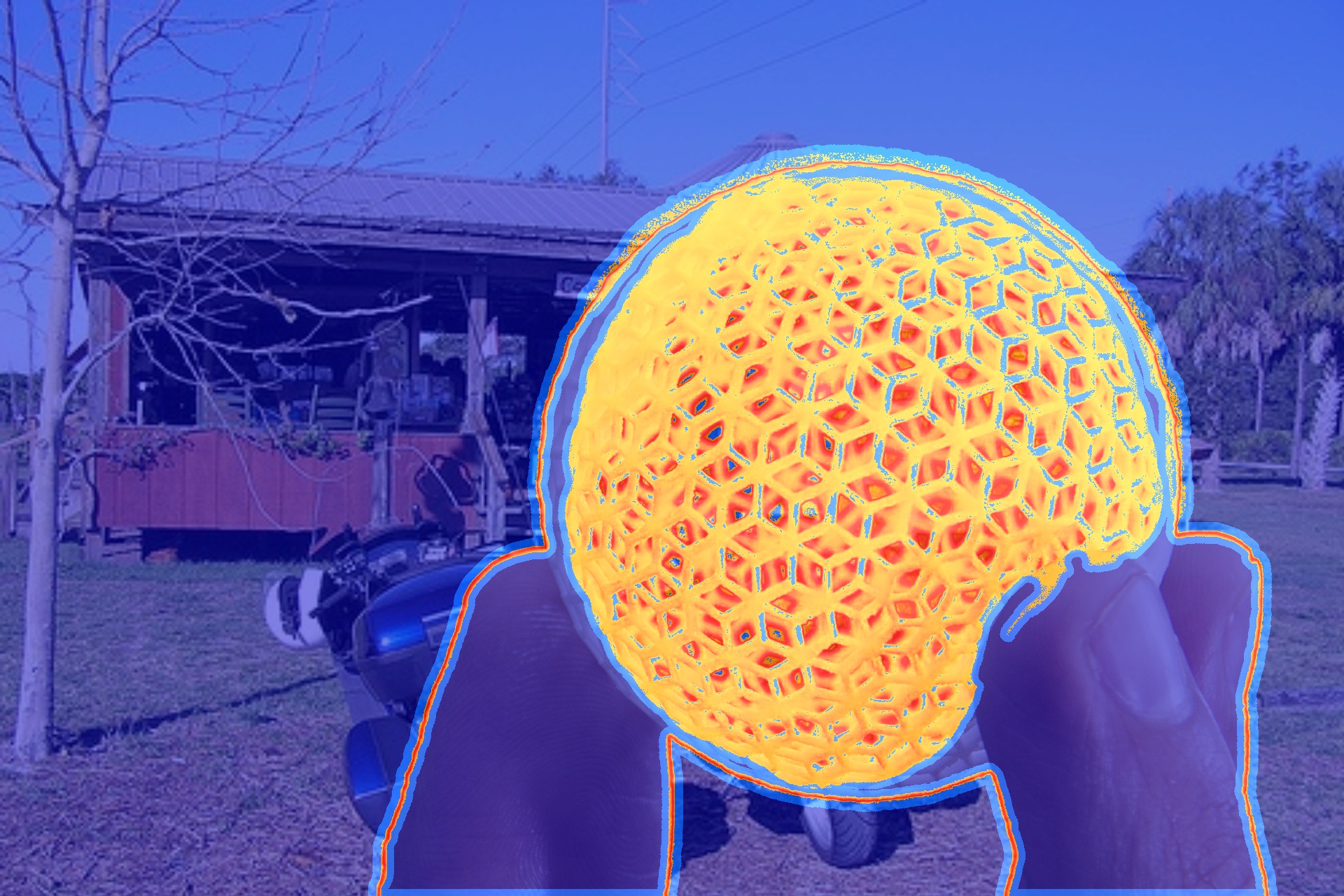}
			\\
			
	\end{tabular}}
	\vspace{-2mm}
	\caption{From left to right: Input trimap, Visualization of area allocation in the unknown regions in trimap, and Visualization of domain response map.  In the second column, {\color{red}{red}} and {\color{green}{green}} represent the deterministic domain, {\color{blue}{blue}} refers to the undetermined domain.  In the third column, the varied colour represents the distinct response degree. The deeper the colour, the stronger the response (from {\color{blue}{blue}} to {\color{orange}{orange}}).}
	\vspace{-3mm}
	\label{fig:visual_trimap}
\end{figure}

Overall, the domain disproportion  and the information discrepancy  can bring necessary details lost for image matting.
 On the one hand,  the pixel distribution of unknown regions is unbalanced (see the second column of Fig.~\ref{fig:visual_trimap}), the number of pixels in the deterministic domains  (\eg~foreground  (\textbf{FG}) pixels ({\color{green}{green} colour}) and background (\textbf{BG}) pixels ({\color{red}{red} colour})) and the undetermined  domains (\textbf{uncertain} in-between pixels ({\color{blue}{blue} colour})) is highly biased. Meanwhile, due to the constraint of trimap, the opacity of large amount of foreground and background pixels are already given. Hence the information dissemination is more conducive to the deterministic domains in unknown regions according to the local smoothness assumption\cite{he2011global}.
 On the other hand, the continuous sampling operation is likely to produce information discrepancy and deteriorate the details of the boundary.
Consequently, we argue that both biased pixel distribution and information discrepancy may produce a negative influence on pixel regression and sampling propagation, resulting in deficient alpha mattes.

\begin{figure*}[h]
  \setlength{\abovecaptionskip}{0.4cm}
  \setlength{\belowcaptionskip}{-0.4cm}
    \centering
    \begin{minipage}[t]{1\linewidth}
    \centering
    \includegraphics[width=6.5in]{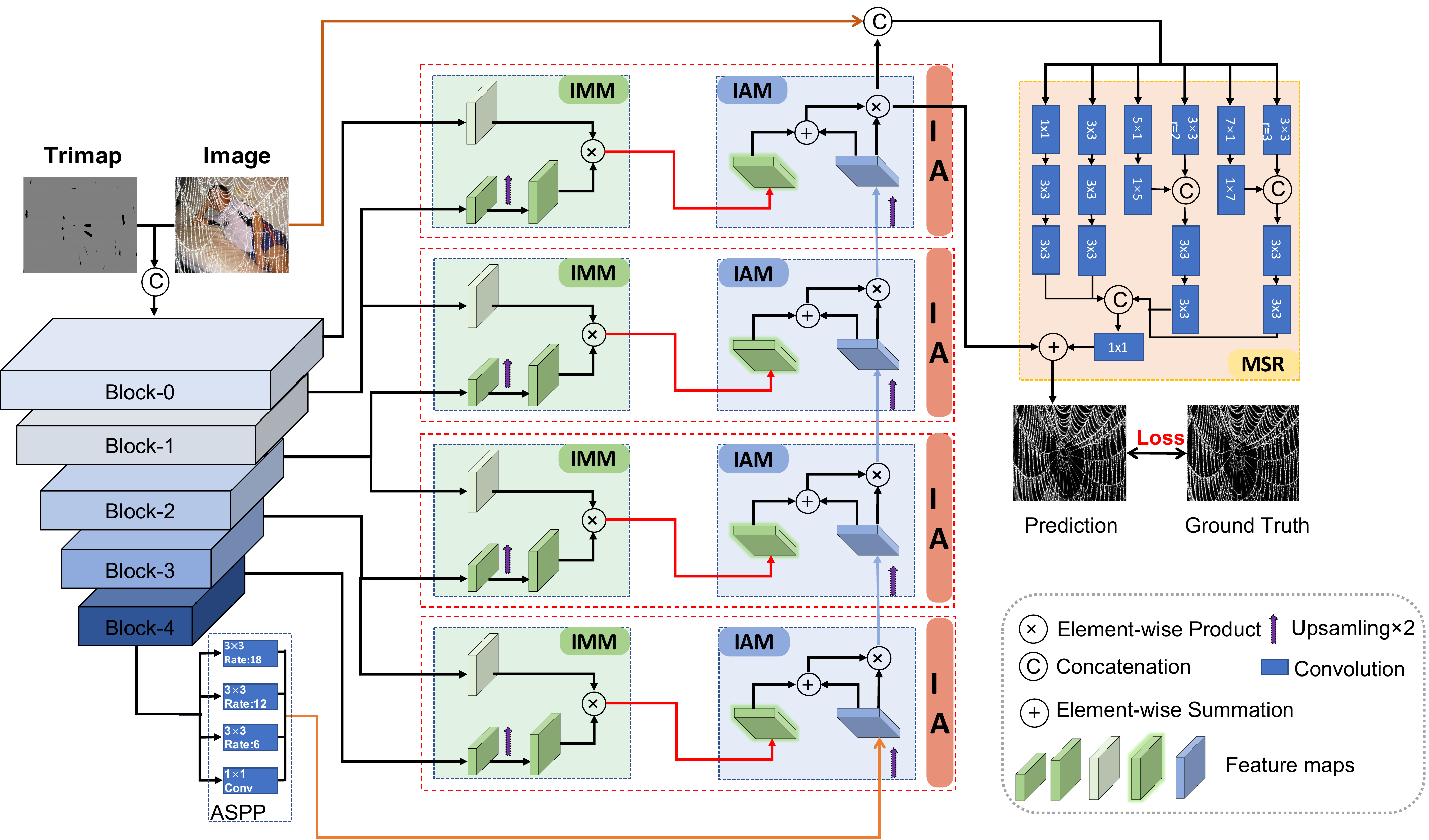}

    \end{minipage}
\centering
\caption{\textbf{Schematic illustration of the proposed PIIAMatting}.
We employ the ResNet-50\cite{he2016deep} as the backbone and utilize the ASPP\cite{chen2018encoder} to extract multi-scale contextual information. According to the depth of the network from shallow to deep, we split the  ResNet-50 into five blocks from block0 to block4. 
Information Match Module (IMM) and Information Aggregation Module (IAM) are concrete implementations of the Information Alignment strategy (IA), represented by the  {\color{green}{green box}} and {\color{blue}{blue box}}, respectively. The {\color{myyellow}{yellow box}} denotes the Multi-Scale Refinement module (MSR).}
\label{fig:pipeline}
\end{figure*}

To address the limitations mentioned above, we propose a Prior-Induced Information Alignment network for image matting (PIIAMatting) as depicted in Fig.~\ref{fig:pipeline}. As for the disproportion of the pixel domains, we propose the Dynamic Gaussian Modulation mechanism (DGM) that can adjust the degree of exploration of pixels at different positions by a domain response map learned from the prior distribution in Ground Truth. The response map can better adapt to the opacity variation, thereby facilitating the entire convergence process during training. To further alleviate the information discrepancy, we propose an Information Alignment strategy (IA) to sufficiently explore the complementarity between adjacent layer-wise features.
It is decoupled into Information Match Module (IMM) and Information Aggregation Module (IAM), which performs guided information matching and information ensemble.
Moreover, a Multi-Scale Refinement (MSR) module is exploited to integrate multi-scale information to improve the alpha matte in a parallel residual fashion. Finally, extensive experiments on three datasets demonstrate that our method can achieve consistently superior performance over recent state-of-the-art approaches.

Our contributions are summarized as follows:
\begin{enumerate}
    \item We propose a Dynamic Gaussian Modulation mechanism (DGM), which can distribute the adaptive responses to each pixel according to the domain response map learned from the prior distribution. This mechanism is very efficient for the highly translucent pixels and can stabilize the process of training.
    \item We propose an Information Alignment strategy consisting of an Information Matching Module (IMM) and an Information Aggregation Module (IAM). Both of them are combined to commit to preserving details by matching and aggregating the features of two adjacent layer-wise features.
    \item Experimental results demonstrate that the proposed PIIAMatting can achieve  state-of-the-art performance on three datasets,  proving the effectiveness and superiority of the proposed method.
  \end{enumerate}

  \section{RELATED WORKS}
  In this section, we will briefly review the image matting from the following three categories: sampling-based, affinity-based, and deep learning-based approaches.
  
  \textbf{Sampling-based} approaches collect a set of known foreground and background samples to find candidate colours for the foreground and background of a given pixel. According to the local smoothness assumption\cite{he2011global} on the image statistics, which requires these sampling colours should be ”close” to the true foreground and background colours. Once the foreground and background colours are determined, the corresponding alpha values can be computed based on the Eq.~\ref{image_synthesis}. Bayesian Matting\cite{chuang2001bayesian}, Cluster Matting\cite{feng2016cluster}, Shared Matting \cite{gastal2010shared}, K-L divergence Matting\cite{karacan2015image}, Global Matting\cite{he2011global}, Comprehensive Matting\cite{shahrian2013improving}, Robust Matting\cite{wang2007optimized}, Iterative Matting\cite{wang2005iterative} are some typical representative methods follow this assumption.

  \textbf{Affinity-based} methods take advantage of affinities of neighboring pixels to propagate the known alpha value from the known regions into unknown regions. ClosedForm Matting\cite{levin2007closed}, Spectral Matting\cite{Levin2008Spectral}, Information-Flow Matting\cite{aksoy2017designing}, KNN Matting\cite{chen2013knn}, Random Walk Matting\cite{grady2005random}, Non-Local Matting\cite{lee2011nonlocal}, Poisson Matting\cite{sun2004poisson} 
  \etc~are some of the known propagation methods introduced in this direction. 
  The most representative method is ClosedForm Matting\cite{levin2007closed}, which is derived from the matting Laplacian and acquires the globally optimal alpha matte by solving a sparsely linear system of equations.
  While the affinity-based approaches can obtain successful alpha matte, they typically encounter the difficulty of high computational complexity and memory limitations.

  \textbf{Deep learning-based} algorithms have recently received considerable attention. Beneﬁting from the effective feature representation of the Deep Convolutional Neural Networks (DCNNs), a few approaches have obtained promising performance. 
  Shen \et \cite{shen2016deep} proposed the ﬁrst automatic matting method for portrait photos in an end-to-end manner.
  DCNN matting\cite{cho2016automatic} combined the results of \cite{chen2013knn} and \cite{levin2007closed} with the input image to predict the final alpha matte. 
  Deep image matting \cite{Xu2017Deep} firstly introduced a large-scale dataset and utilized the Seg-Net\cite{badrinarayanan2017segnet} with a refinement to jointly estimate the alpha mattes. 
  AlphaGAN\cite{lutz2018alphagan} firstly introduced the generative adversarial network (GAN)\cite{goodfellow2014generative,han2019asymmetric} to generate alpha mattes using a discriminator,  indicating the GAN is capable to process well for pixel-wise regression tasks.
  SampleNet\cite{tang2019learning} and GCA Matting\cite{li2020natural} learned from image context to leverage the background information to guide the foreground prediction.  
  IndexNet matting\cite{lu2020index} introduced the idea of indexing information to integrate upsampling operators for improving the ability to retain details. 
  AdaMating\cite{cai2019disentangled} divided the matting task into two branches (i.e., trimap adaption and natural image matting), which can work together to jointly refine the alpha matte.
  Similarly, the Context-Aware matting \cite{hou2019context} dismantled the image matting into foreground  and alpha estimation,  then pushed them to generate the alpha matte and foreground image simultaneously.
 However, almost all the above methods turned to a static optimization without taking the opacity variation into account, leading to some obvious artifacts.
Moreover, LFM  matting\cite{zhang2019late} combined two classification networks and a fusion network under the supervision of a hybrid loss to jointly predict alpha mattes with single RGB image as input. Similar to LFM without auxiliary assistance, the HAttMatting\cite{qiao2020attention} employed spatial and channel-wise attention to integrate appearance cues and advanced semantics to achieve better results.
Besides, there are some other deep learning-based approaches\cite{chen2019learning, chen2018semantic, yang2018active, wang2018deep, aksoy2018semantic, liu2020boosting, sengupta2020background, he2017discovering,li2008faceseg,zhu2016multiple,msia,Yang2020Smart} to solve image matting and the field-specific tasks.

\section{Method}
In this section, we ﬁrst describe the approach overview brieﬂy in \ref{ssec:overview}. Then, we discuss the interrelation between different loss functions and introduce our Dynamic Gaussian Modulation mechanism (DGM) in \ref{ssec:gaussian}. Subsequently, we give a detailed depiction of our network structure in \ref{ssec:model} and elaborate on the loss function we use during the training in \ref{ssec:loss}. The overall architecture is shown in Fig.~\ref{fig:pipeline}.

\subsection{Overview}
\label{ssec:overview}
The deep learning-based image matting methods have shown their advantages compared with the original colour-based methods. However, most of them either treat pixels at different domains equally or lose potential boundary details during the sampling operation, resulting in the sub-optimal alpha matte. Consequently, we propose a PIIAMatting to model the pixel-wise domain modulation and the information alignment within a single network in an end-to-end manner. On the one hand, we argue that the pixels in different regions (such as the $\alpha_i\in(0,1)$ and the $\alpha_i=0$ or 1) should be treated as distinguished so that the model can be driven to focus more on the hard-to-mine samples. 
On the other hand, a well-balanced model should also explore layer-wise features to enhance the ability to match and integrate fine-grained details, which is critical in image matting, especially in boundary areas.

\subsection{Dynamic Gaussian Modulation}
\label{ssec:gaussian}

With the trimap as assistance, the target of image matting is to predict the opacity of each pixel within the unknown regions. 
 As shown in the second column of Fig.~\ref{fig:visual_trimap}, typically, the pixels in  unknown regions could be divided into the deterministic  and undetermined domains. The deterministic domains can be split into certain \textbf{FG} and \textbf{BG} pixels, while the undetermined  domains can be denoted as highly uncertain in-between pixels directly. 
As a common practice, it is solved as a regression problem using L1 or L2 Loss, but it  has limitations to a certain extent.
On the one hand, the distribution of the pixels is highly biased between the  deterministic and undetermined domains. Normally, the proportion of the  FG and Bg pixels is far less than the in-between pixels. This disproportion may cause instability during model learning and potentially deteriorate the quality of the alpha matte.
On the other hand, the exploration of opacity is varied according to the positions of pixels domains. 
Based on the convolution slide window and local smoothness assumption, the pixels at the deterministic domains ({\color{green}{green}} and {\color{red}{red}} colour) are directly connected to the known regions. Thus the opacity solution process on the deterministic domains is more accessible than the undetermined domains ({\color{blue}{blue}} colour).

 Without regard to varying network structures\cite{Xu2017Deep, hou2019context, zhang2019late}, the solving process of the pixel opacity is mainly optimized  by the objective function. In addition to the L1 and L2 loss, composition loss\cite{Xu2017Deep} is also widely used in the RGB domain to improve the alpha mattes. Although the blended loss function can improve the result, it does not solve the problem of disproportion essentially. 
 
 Another solution was proposed in LFM\cite{zhang2019late}, which employs distinct losses to supervise different areas: 
\begin{figure}[t]
\setlength{\abovecaptionskip}{0.4cm}
\centering
\begin{minipage}[t]{1\linewidth}
\centering
\includegraphics[width=2.8in]{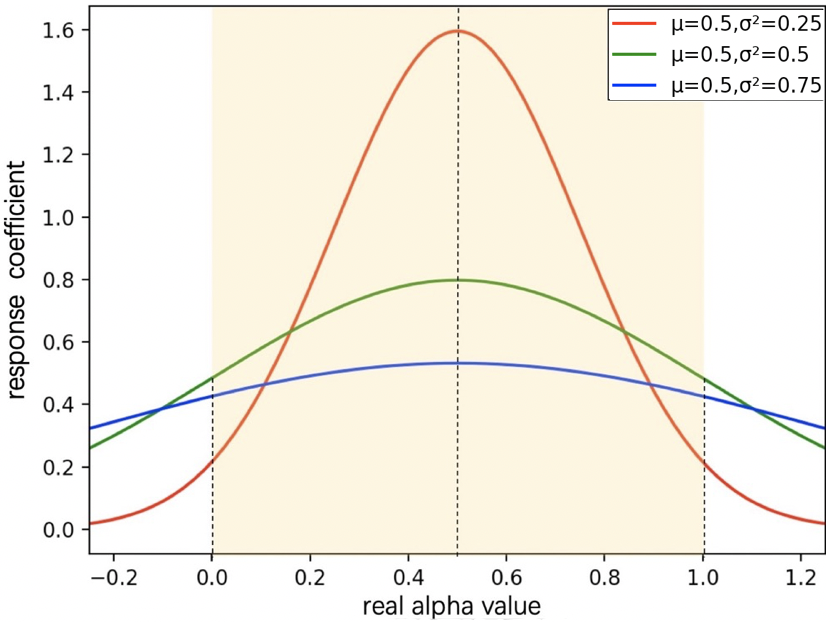}

\end{minipage}
\centering
\caption{The response coefficient variation based on the Dynamic Gaussian Modulation mechanism. Where $\mu=0.5$, the {\color{red}{red}}, {\color{green}{green}}, and {\color{blue}{blue}} lines represent the response coefficient for different pixels when the $\sigma^2$ is 0.25, 0.5, and 0.75. The diverse $\sigma$ represents varied exploration schemes, particularly the varied degree of excavation for the undetermined domains.}
\label{fig:weight_gaussian}
\end{figure}

\begin{equation}
  \mathcal{L}^{i}_{\alpha}=\left\{
  \begin{aligned}
    \left | \alpha_{p}^{i}-\alpha_{g}^{i}  \right |,\qquad  0< \alpha_{g}^{i} < 1 .\\
      (\alpha_{p}^{i}-\alpha_{g}^{i})^{2},\qquad \alpha_{g}^{i} \in [0,1]. \\
  \end{aligned}
  \right.
\end{equation}
where  $\alpha_{p}^{i}$ and $\alpha_{g}^{i}$ refer to  the predicted alpha matte and ground truth alpha matte at pixel i separately.
However, since the transparency  is continuously distributed, this solution can only be expressed as a static pre-defined partition, which does not completely solve the dynamically varied opacity. Furthermore, it is difficult to draw a clear demarcation line to separate the deterministic and undetermined domains.

Based on the above observations, we propose our Dynamic Gaussian Modulation mechanism (DGM), which naturally fits the task of image matting. As shown in Fig.~\ref{fig:weight_gaussian},  
the DGM  is evolved  from the standard Gaussian distribution for which $\mu=0$, $\sigma^{2}=1$. The different colours represent the distinct  response scheme. From top to bottom, the response coefficient for the undetermined domains is getting smaller and smaller.
Specifically, we denote the Ground Truth as  $
 \boldsymbol{F}_{ground-truth} \in \mathbb{R}^{1 \times H \times W}$, and the output feature map holds the same size as the Ground Truth. 
 Since the distribution of pixels in the transition regions varies for different images, we resort to the case-specific prior distribution in the corresponding Ground Truth to acquire the domain response map.
\begin{equation}
	R=f^{transform}\left({F}_{ground-truth}\right)=\frac{1}{\sqrt{2\pi\sigma^{2}}}e^{-\frac{(\alpha^{i}_{g}-\mu)^{2}}{2\sigma^{2}}}
\end{equation}
where $i$ denotes the position of pixels, $g$ refer to the Ground Truth, and the $\alpha_{g}^{i}$ is the ground truth alpha matte at pixel i. The mean $\mu$ and  variance $\sigma^{2}$ of the Gaussian distribution are adjustable parameters, and they were set to $0.5$ and $0.25$ unless otherwise specified. 
The  Gaussian Distribution will assign slighter responses to the deterministic domains ({\color{green}{foreground}} and {\color{red}{background}} pixels) and stronger responses to the undetermined domains (uncertain {\color{blue}{in-between}} pixels). The higher the uncertainty level of pixels, the stronger the responses. In general, the higher the uncertainty for pixels with opacity closer to 0.5.
In this way, every pixel at every image can attain an  opacity-adaptive pixel-wise domain response coefficient  and contribute to the network to explore more valuable information.

Furthermore, as the training progresses, the  adaptability of our model to unknown regions is continuously enhanced. In order to better match the opacity variation according to the fitting ability of the  model  during the training process, we further propose to adjust the response map dynamically.  Intuitively, we can gradually reduce the $\sigma^2$ as the number of training iterations  progresses so as to modulate the responsiveness of the deterministic  and undetermined domains. The specific formulation is as follows:
\begin{equation}
{D}(\sigma^{2}) = \sigma^{2} + 0.005 \qquad while \; iteration \% 2000= 0
\end{equation}
the $\sigma^2$ is initialized to 0.25 and increased by 0.005 every 2000 iterations.
As the iteration progresses, the response coefficient assigned to the undetermined  domains gradually decreases.
 Hence, we can achieve domain-specific weight distribution through the Dynamic Gaussian Modulation mechanism induced by the prior information, which can stabilize the convergence process while exploring information evenly. 
 Some examples of the domain response map obtained from our DGM are shown in the third column in Fig.~\ref{fig:visual_trimap}. Since the pixels are continuously distributed inside the unknown regions, the domain response coefficient varies according to the different opacity of pixels. From {\color{blue}{blue}} to {\color{orange}{orange}}, the deeper the colour, the stronger the response.

\subsection{Model Architecture}
\label{ssec:model}

As mentioned above, we  explicitly assign the dynamic responses to different pixels  for exploring various types of regions. 
However, as the network deepens, the resolution of the feature map decreases gradually. During this process, the down-sampling operation inevitably results in information loss, especially the average pooling that discard the high-frequency information and only retain the low-frequency information\cite{jpeg_transform}, leading to ambiguity in the boundary regions. Same as average pooling, the other two prevalent options for down-sampling are max-pooling and convolution with stride larger than 2, which also incur different degrees of information loss. 
Therefore, the information discrepancy will occur in the corresponding up-sampling stage in the decoder, which will bring accuracy loss\cite{lu2020index} and potentially affect the convergence of the  model.
Consequently, we argue that designing a strategy to match valid information for aggregation effectively can improve the model's overall performance.

  In the following, we firstly describe the overall network structure. Then, the Information Alignment strategy is introduced detailedly in both the Information Match Module and Information Aggregation Module. At last come to introduce our MSR.

\textbf{Overall network structure: }
 Fig.~\ref{fig:pipeline} shows the pipeline of our model, which uses the original image concatenated with  trimap as input. 
 We use ResNet-50\cite{he2016deep} as our backbone, and it is split into five blocks from block0 to block4 according to the depth of the network from shallow to deep. In order to fuse multi-scale contextual information and enhance the feature representation capability, we extract the features from the block4 and transmit them to the ASPP\cite{chen2018encoder} to aggregate features of different receptive ﬁelds. 
 Specifically, considering the computation overhead, we exploit three parallel atrous convolution layers with different rates and an average pooling layer followed by a convolution operation with a kernel size of 1 $\times$1  to capture multi-scale information. The dilation rates are set to 1, 8 and 16, respectively. Then, the features of the four branch are concatenated and fed into a convolution operation with a kernel size of 3$\times$3 followed by a BatchNorm and ReLu.

 To ensure the integrity and validity of the sampled features, we utilize the IMM and IAM to jointly perform the adjacent layer-wise information match  and  aggregation. As the information alignment has been decoupled into information match and aggregation, the information aggregation is automatically fulfilled once the information match is complete.
  Besides, the MSR applied dilated convolutions with different rates in a parallel residual manner to complement more details and generate high-quality results.

\textbf{Information Alignment strategy:}
Skip-Connection\cite{ronneberger2015u} is a compelling feature enhancement method in different kinds of vision tasks\cite{yang2018drfn,chen2018encoder,xu2020learning}, which bridge the information gap by concatenating or adding features between encoder and decoder at each down-sampling and up-sampling stage. 

Although it can attenuate the information loss to some extent, it uses only the peer-to-peer level features and ignores the relationship between adjacent layer-wise features (\eg~features from block0 and block1), thus inevitably introducing superfluous bias during the constant sampling operation.
Therefore, it is only  direct reuse of features without discrimination and does not radically address the information discrepancies. 
Moreover, the low-level  and high-level features are intrinsically different but complementary to each other, one for the details preserved while the other for semantic retaining, then jointly contribute to the generation of alpha mattes. In the IA, we perform the IMM to match the valuable information at each stage in the encoder between adjacent layer-wise layers during down-sampling. As shown in Tab.~\ref{tab:ablation_networks},  the experimental analysis also confirms the effectiveness of our Information Alignment strategy.

\textit{Information Match Module:~}
\label{ssec:imm}

We feed two inputs to the IMM: $F_{low}\in R^{H*W*C}$ and  $F_{high}\in R^{\frac{H}{2}*\frac{W}{2}*C}$. H, W and C represent the height, width and number of channels respectively.
Firstly, we transfer $F_{high}$ to the transposed convolution for up-sampling.
Then, the product of $F_{high}$ and $F_{low}$ is used to match the valuable details, clear foreground, and background features into the IAM module. The implementation process is shown in the green box in Fig.~\ref{fig:pipeline} and specific operations can be depicted as follows:
\begin{equation}
  \mathcal{F}^{i}_{imm}=
    \mathcal{F}_{low}^{i-1}*\mathcal{T}(\mathcal{F}_{high}^{i})
\end{equation}
where i refers to the index of the block in ResNet, $T(\cdot)$ is the upsampling operation and $*$ denotes the element-wise product.

\begin{table*}[]
\caption{
 Comparison of our method with eight top-performing algorithms on the Alphamatting.com dataset\cite{rhemann2009perceptually}. “O” represents overall rank, “S”, “L”, and “U” represent performance corresponding to the trimaps with three difficulty levels: small, large, and user.
 Please see Alphamatting.com for more results.}
\label{tab:performance_comparison}
\centering
\begin{tabular}{l|cccc|cccc|cccc}
                  \hline
\multirow{2}{*}{Methods} & \multicolumn{4}{c}{MSE} & \multicolumn{4}{c}{SAD}  & \multicolumn{4}{c}{Gradient} \\
                 
& Overall  & S    & L    & U    & Overall & S    & L    & U    & Overall   & S      & L     & U     \\
\hline
\rowcolor{mygray}
Our & \textbf{7.6} & \textbf{3.8} & 9.8 & 9.4 & \textbf{4.9} & \textbf{3.1}  & \textbf{6.0} & \textbf{5.6} & \textbf{8.0} & 5.3 & 7.8 & 11.0 \\
AdaMatting\cite{cai2019disentangled} & 8.5 & 6.5 & \textbf{7.8} & 11.4 & 7.5     & 6.8 & 6.5 & 9.4 & 8.1 & \textbf{5.0} & \textbf{5.6} & 13.6 \\
SampleNet\cite{tang2019learning} & 9.7 & 6.6 & 9.6 & 12.8 & 8.4 & 6.6 & 8.0 & 10.6 & 9.7 & 5.9 & 7.4 & 15.8 \\
GCA\cite{li2020natural} & 10.0 & 10.0 & 8.6 & 11.3 & 9.3 & 10.1 & 6.8 & 11.0 & \textbf{8.0} & 8.1 & 6.6 & 9.3 \\
Context-Aware\cite{hou2019context} & 12.2 & 15.8 & 13.4 & \textbf{7.5} & 18.1 & 22.0 & 16.0 & 16.3 & 9.3 & 10.6 & 9.9 & \textbf{7.3} \\
IndexNet\cite{lu2020index} & 17.8 & 20.4 & 16.1 & 16.8 & 14.1 & 16.5 & 12.8 & 13.1 & 13.3 & 12.3 & 11.8 & 16 \\
Information-Flow\cite{aksoy2017designing} & 14.5 & 17.3 & 13.8 & 12.5 & 13.1    & 14.3 & 13.8 & 11.4 & 20.1 & 23.0 & 18.8 & 18.6 \\
DIM\cite{Xu2017Deep} & 13.8 & 12.5 & 12.5 & 16.5 & 10.9 & 12.3 & 10.1 & 10.3 & 18.3 & 15.4 & 14.9 & 24.6 \\
AlphaGAN\cite{lutz2018alphagan} & 18.8 & 19.3 & 19.9 & 17.4 & 15.7 & 16.5 & 15.8 & 14.8 & 18.1 & 17.0 & 15.9 & 21.4 
\\
\hline
\end{tabular}
\end{table*}
  
\begin{figure*}[h]
	\setlength{\tabcolsep}{1pt}\small{
		\begin{tabular}{ccccc}			
			\includegraphics[scale=0.508]{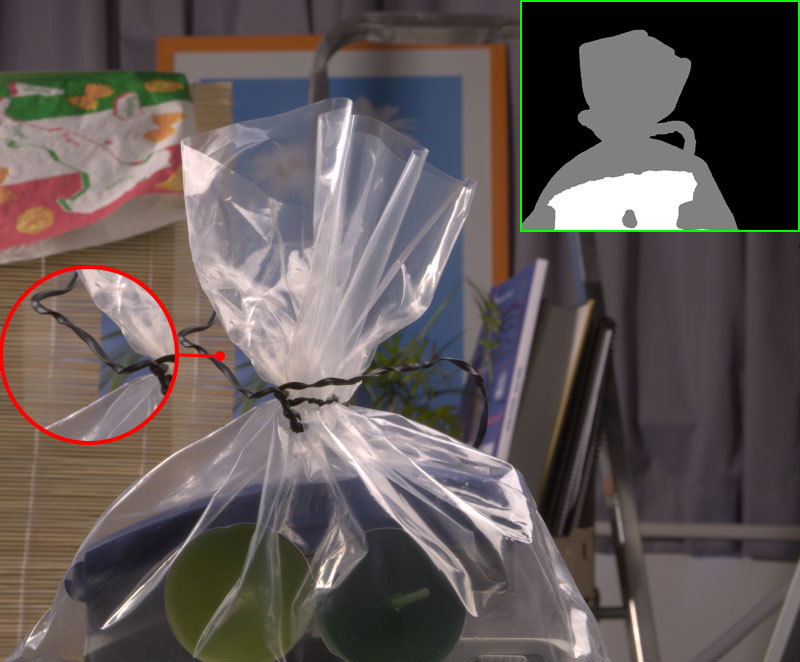} &
			\includegraphics[scale=0.122]{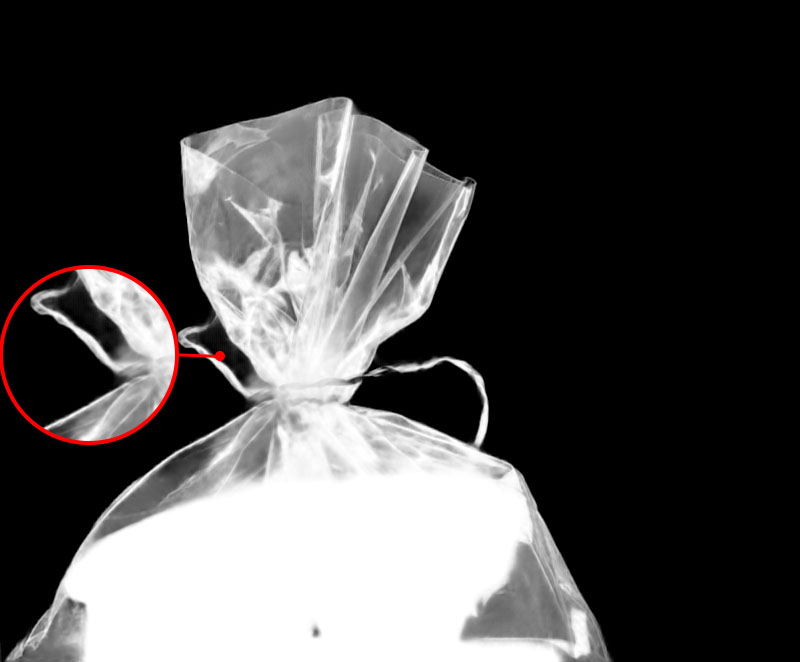} &			
			\includegraphics[scale=0.122]{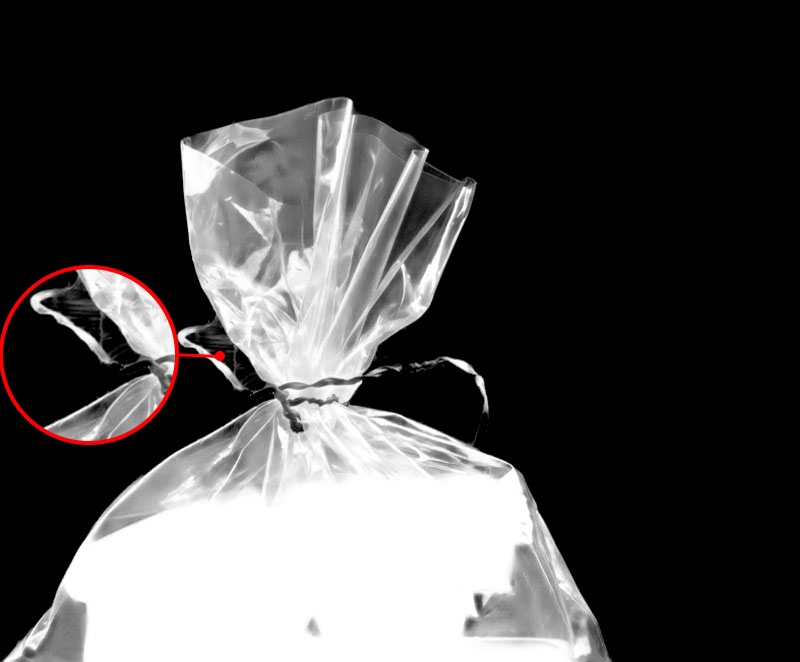} &
			\includegraphics[scale=0.122]{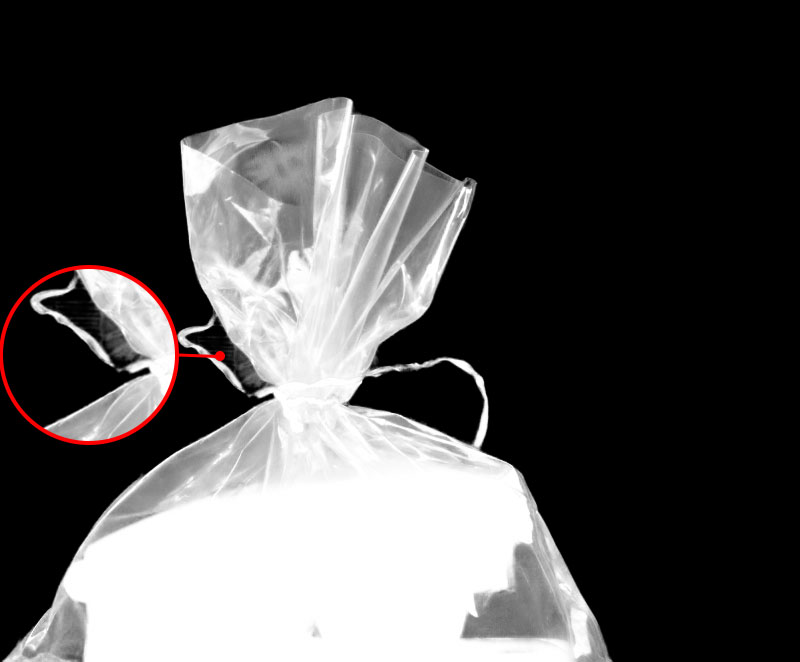} &
			\includegraphics[scale=0.122]{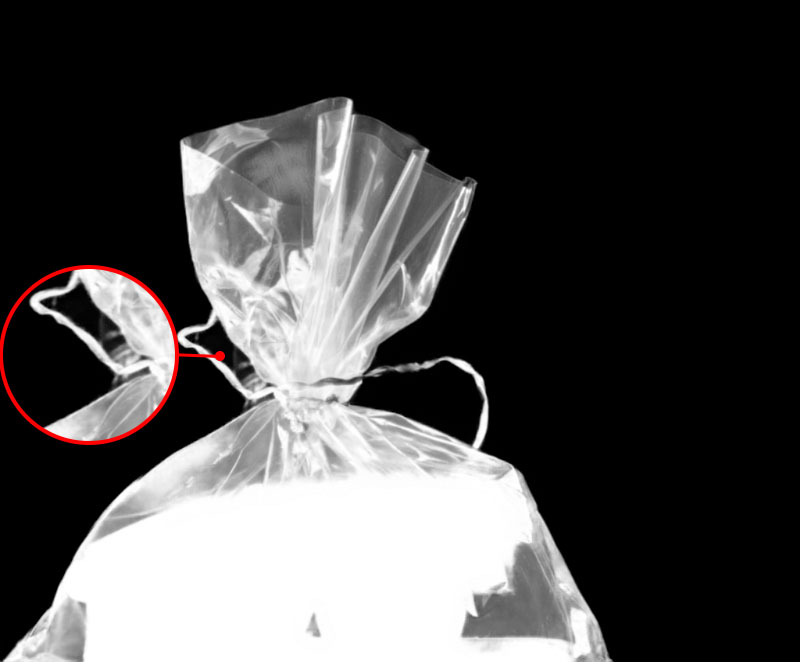} \\
	
			\includegraphics[scale=0.508]{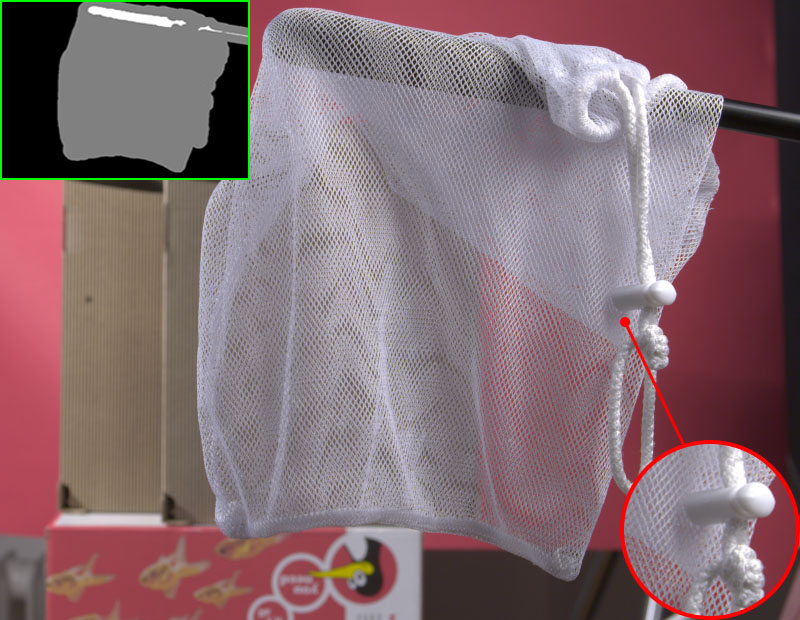} &
			\includegraphics[scale=0.122]{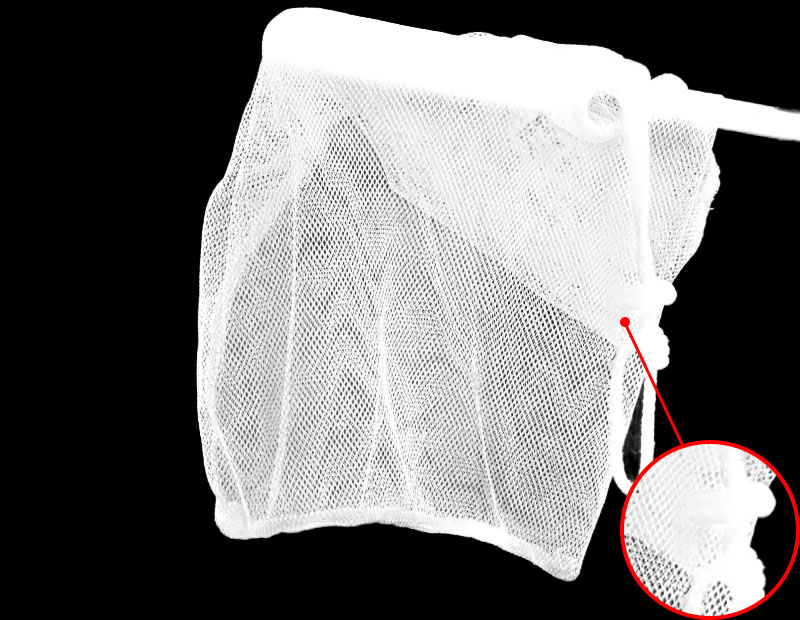} &			
			\includegraphics[scale=0.122]{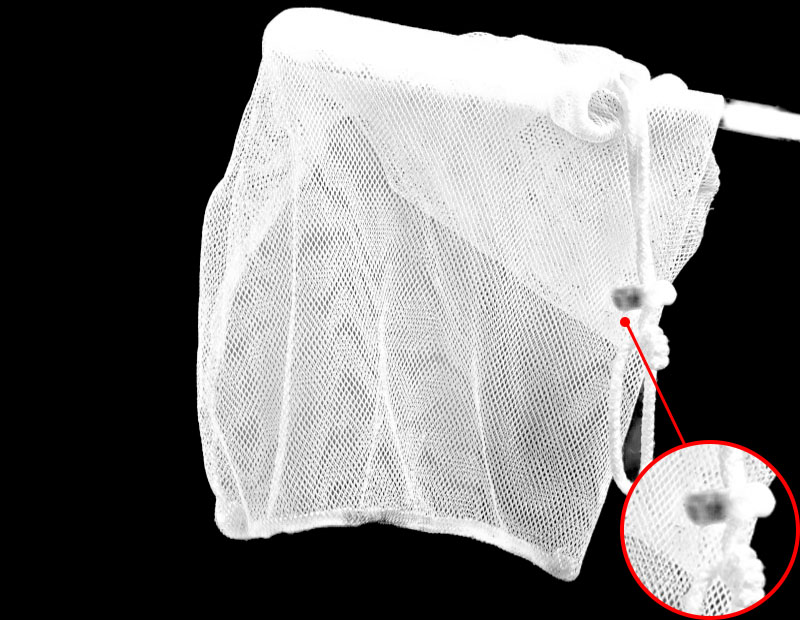} &
			\includegraphics[scale=0.122]{./figure/Fig.5/2/gca.jpg} &
			\includegraphics[scale=0.122]{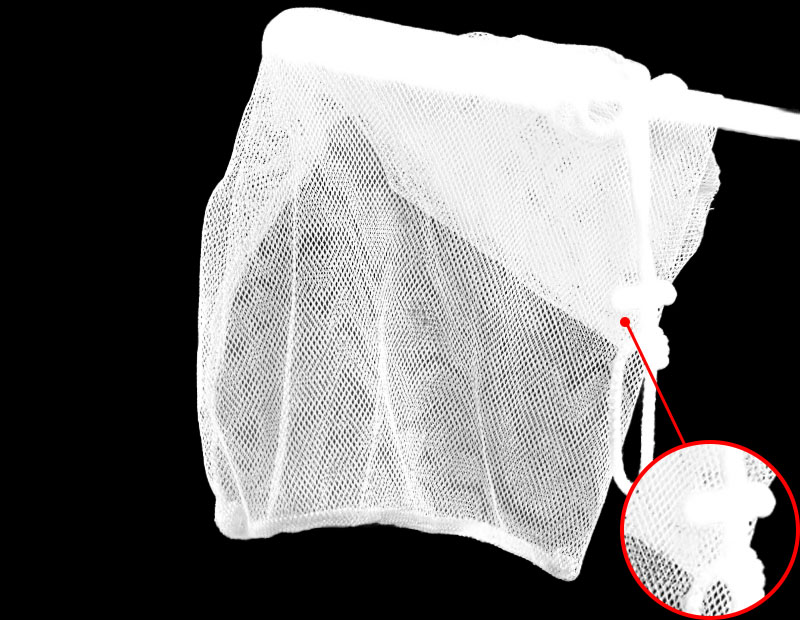} \\
			
			\includegraphics[scale=0.122]{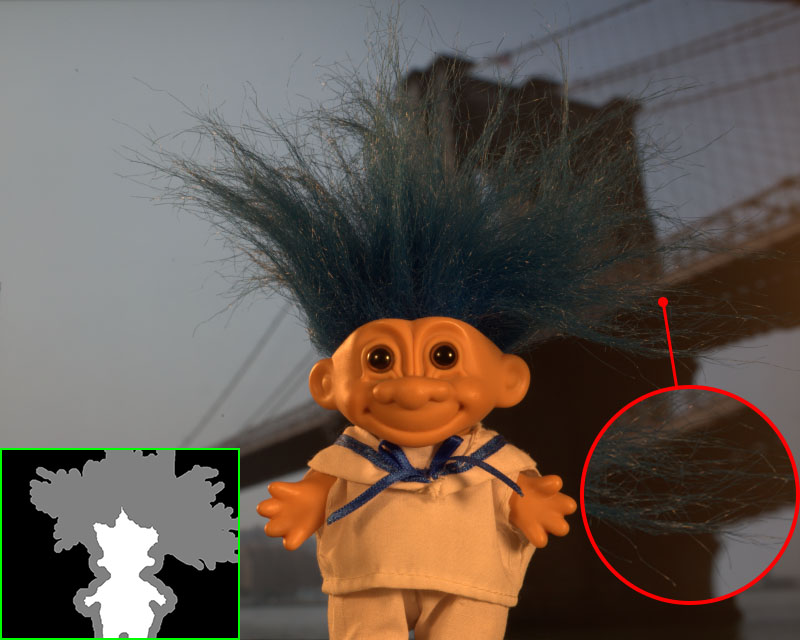} &
			\includegraphics[scale=0.122]{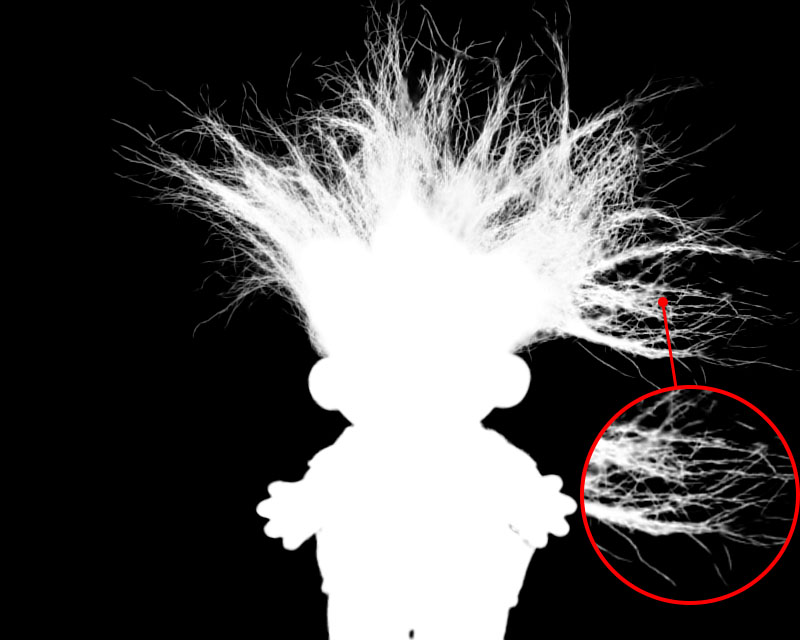} &			
			\includegraphics[scale=0.122]{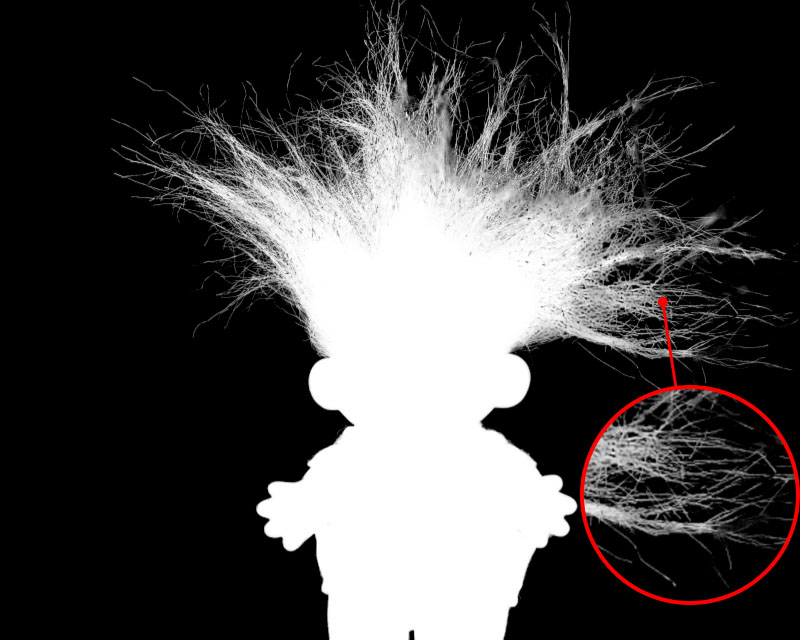} &
			\includegraphics[scale=0.122]{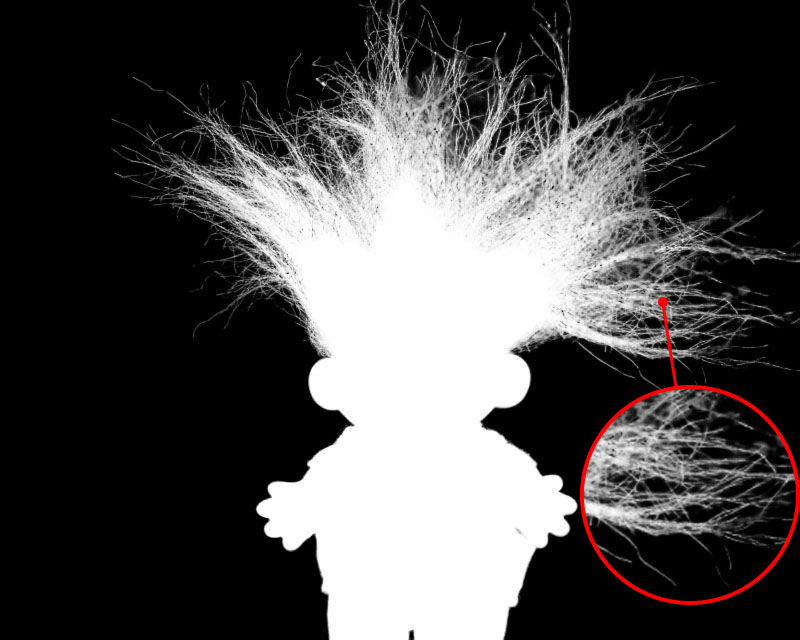} &
			\includegraphics[scale=0.2032]{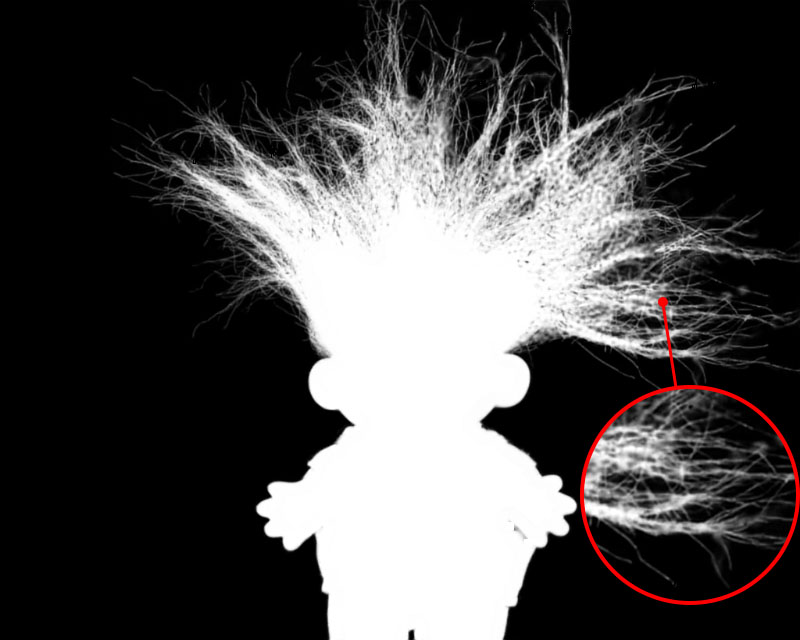} \\
			Inputs & DIM\cite{Xu2017Deep} & IndexNet\cite{lu2020index} & GCA\cite{li2020natural} & PIIAMatting (Ours)  \\

	\end{tabular}}
	\vspace{-1mm}
	\caption{Qualitative comparisons with other latest algorithms on the Alphamatting.com\cite{rhemann2009perceptually} test set. }
	\vspace{-3mm}
	\label{fig:visual_benchamrk}
\end{figure*}

\begin{figure*}[h]
	\setlength{\tabcolsep}{1pt}\small{
		\begin{tabular}{ccccc}			
			\includegraphics[scale=0.0762]{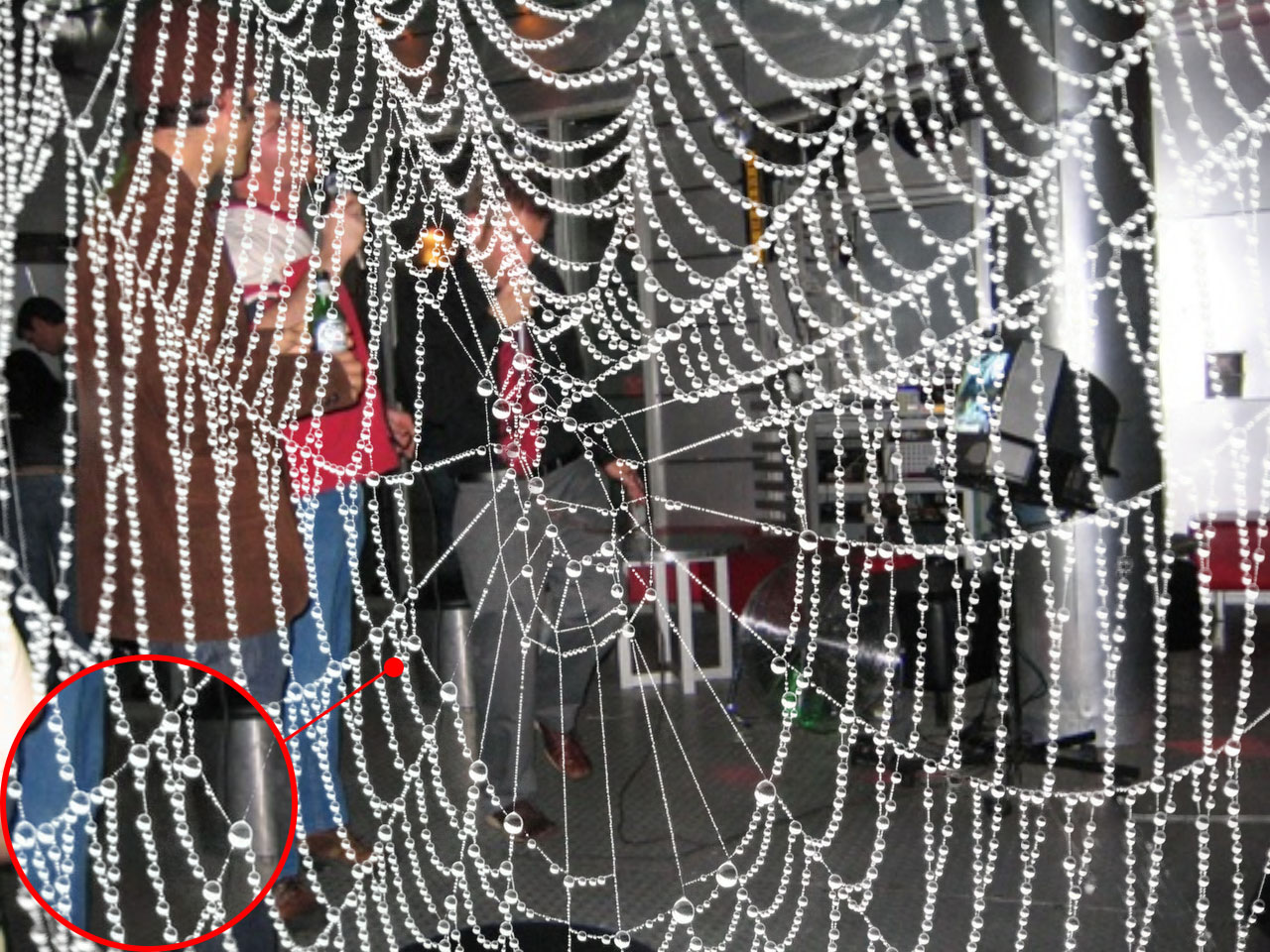} &
			\includegraphics[scale=0.0762]{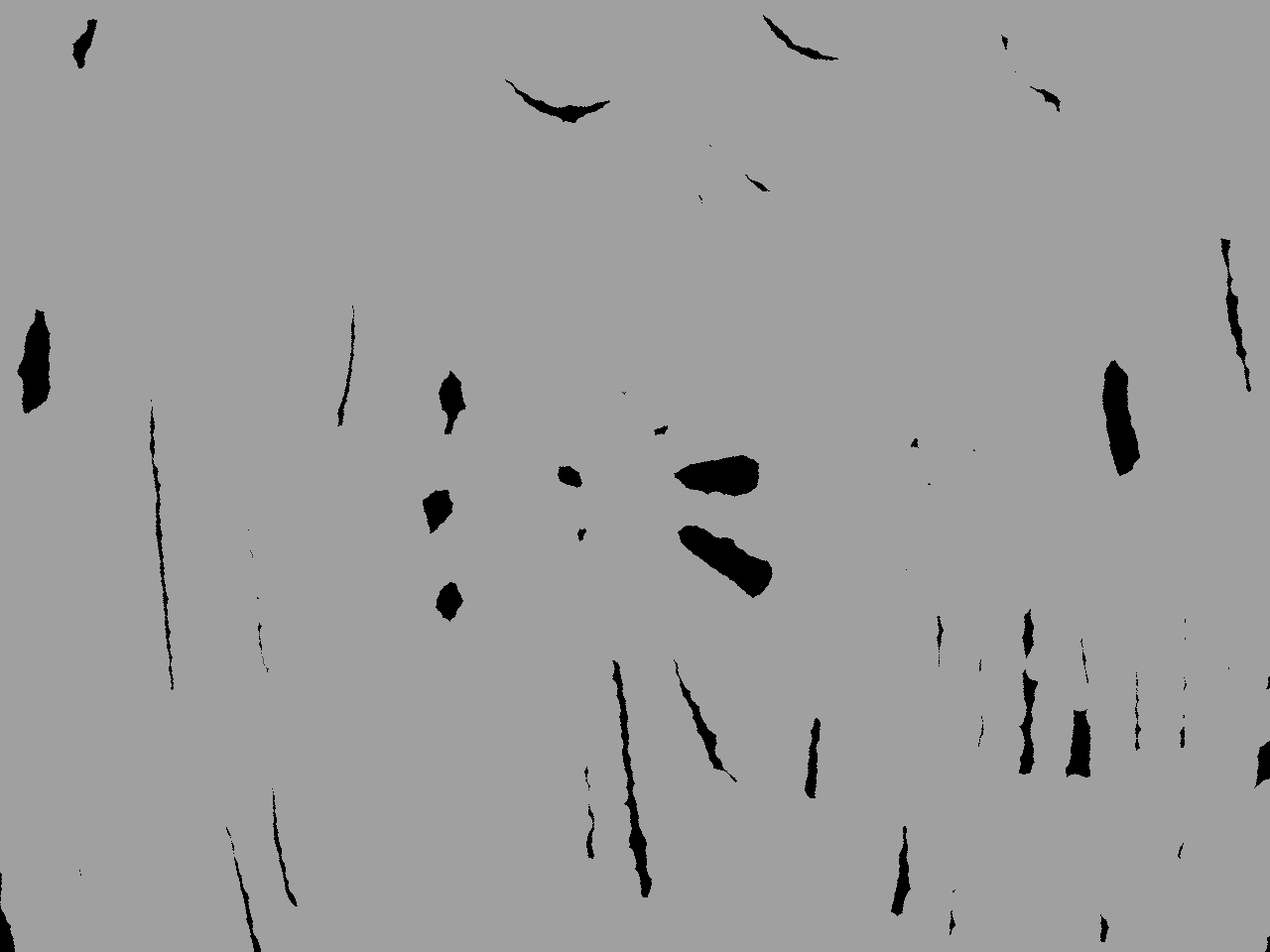} &			
			\includegraphics[scale=0.0762]{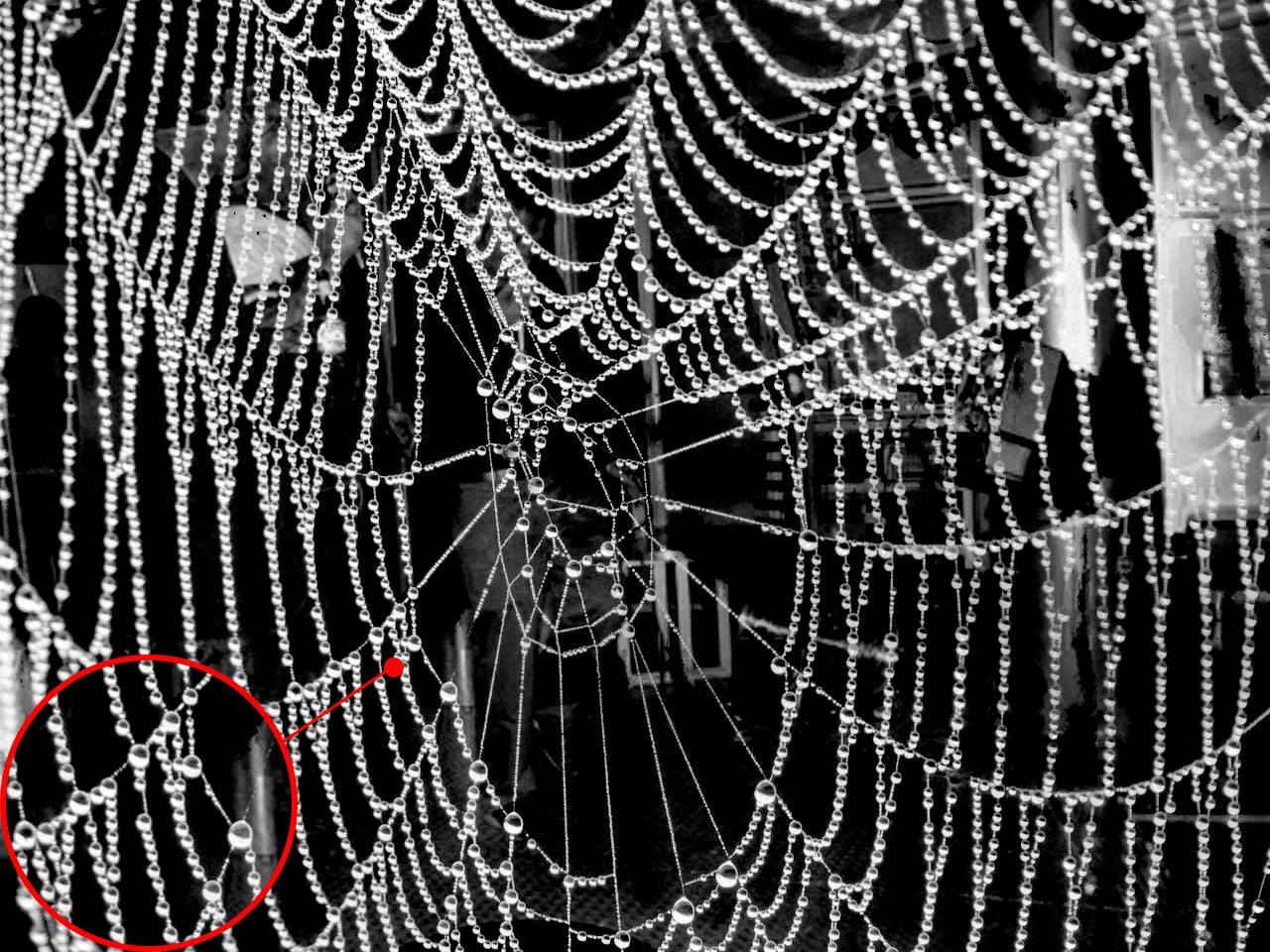} &
			\includegraphics[scale=0.0762]{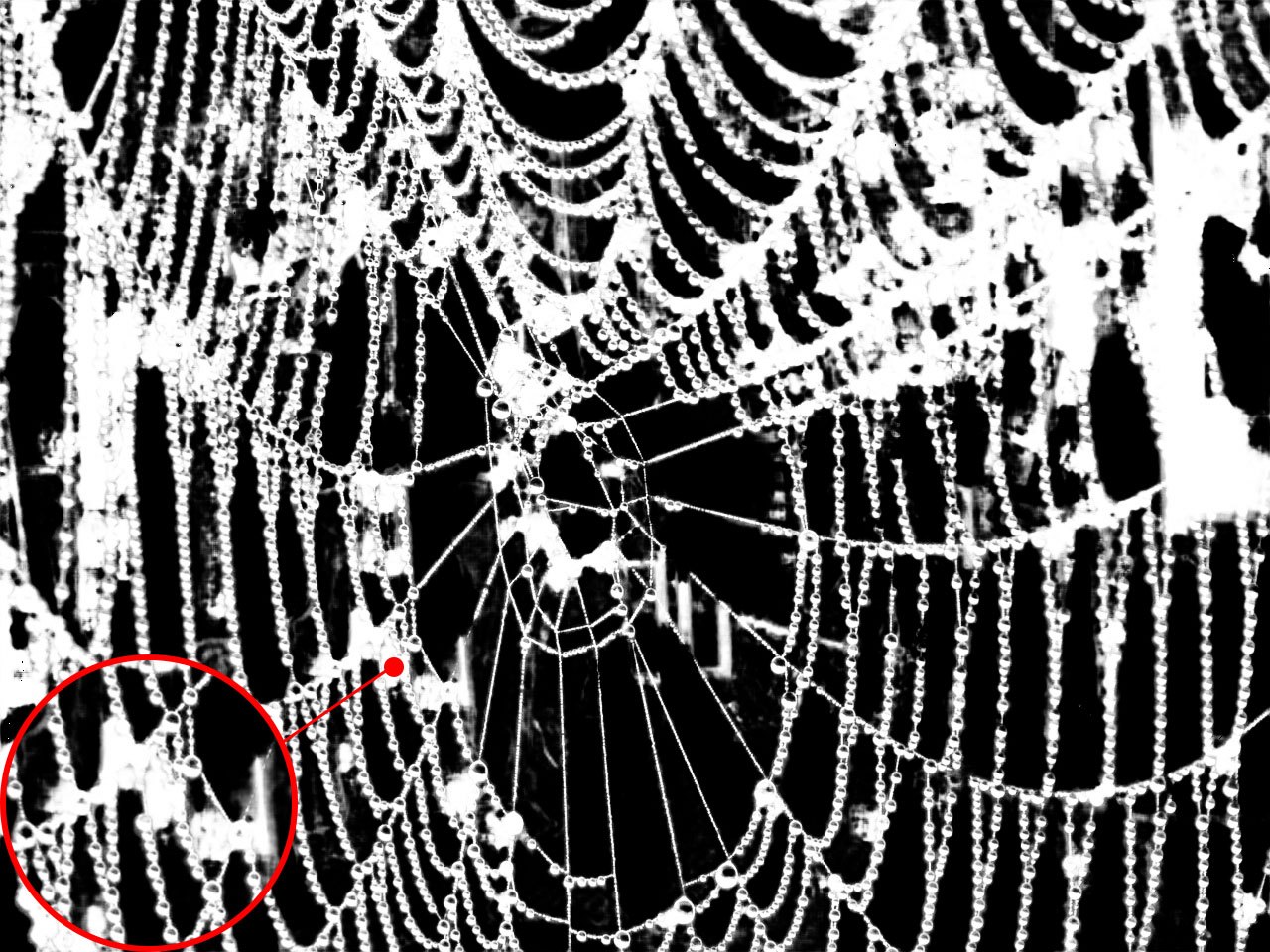} &
			\includegraphics[scale=0.0762]{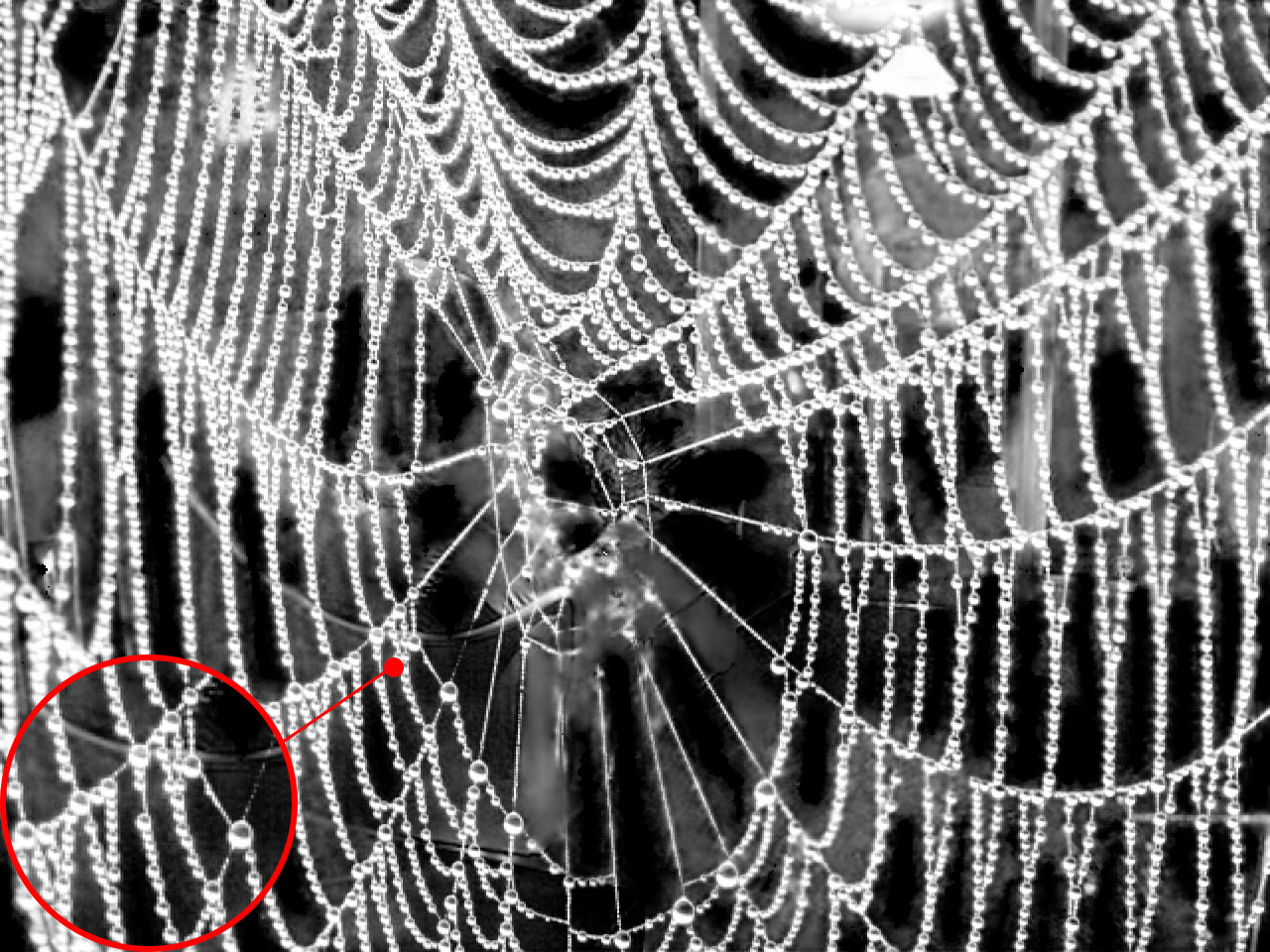} \\
			Image & Trimap & Information-Flow\cite{aksoy2017designing} & AlphaGAN\cite{lutz2018alphagan} & SampleNet\cite{tang2019learning}  \\
			\includegraphics[scale=0.0762]{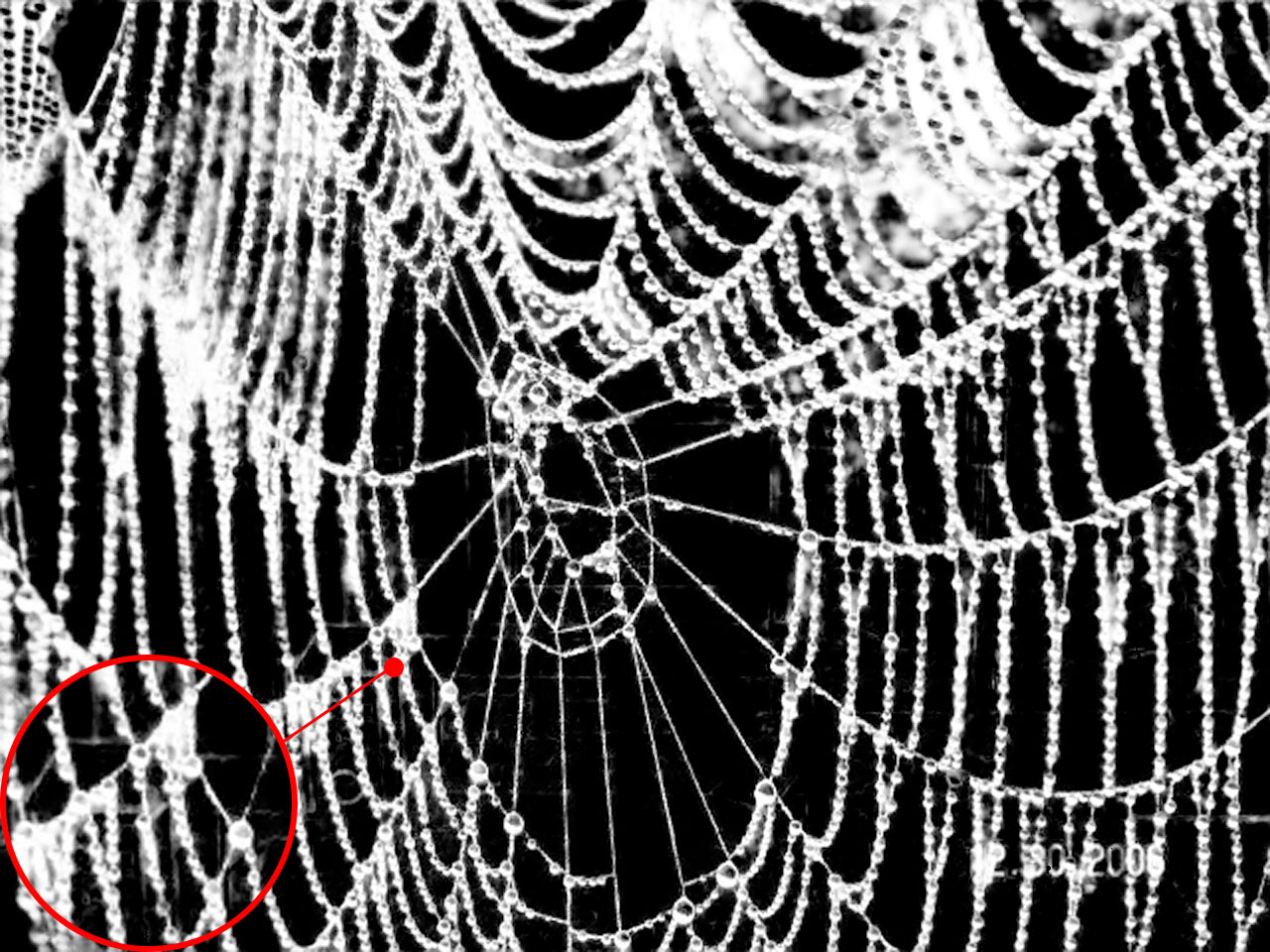} &
			\includegraphics[scale=0.0762]{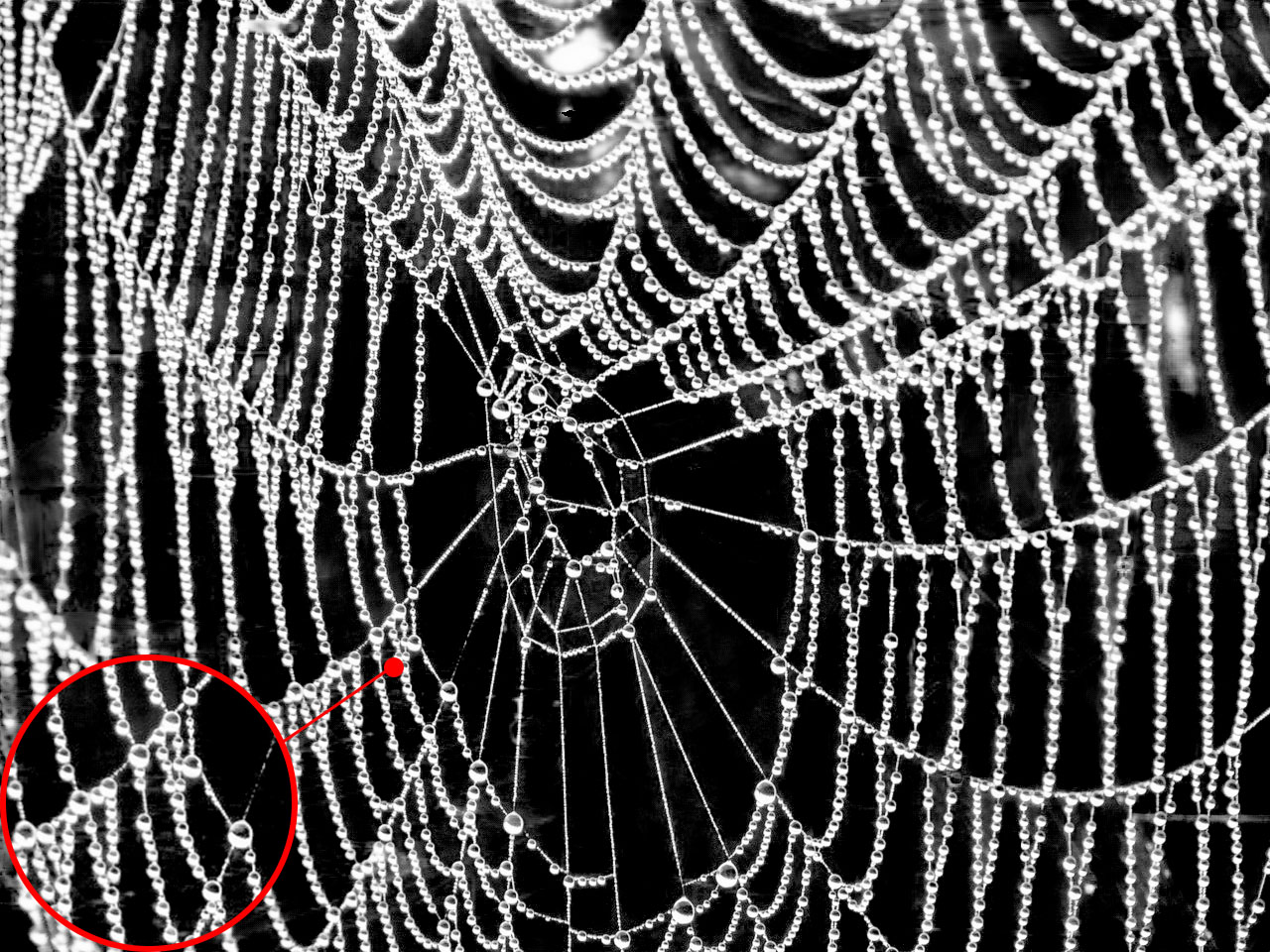} &			
			\includegraphics[scale=0.0762]{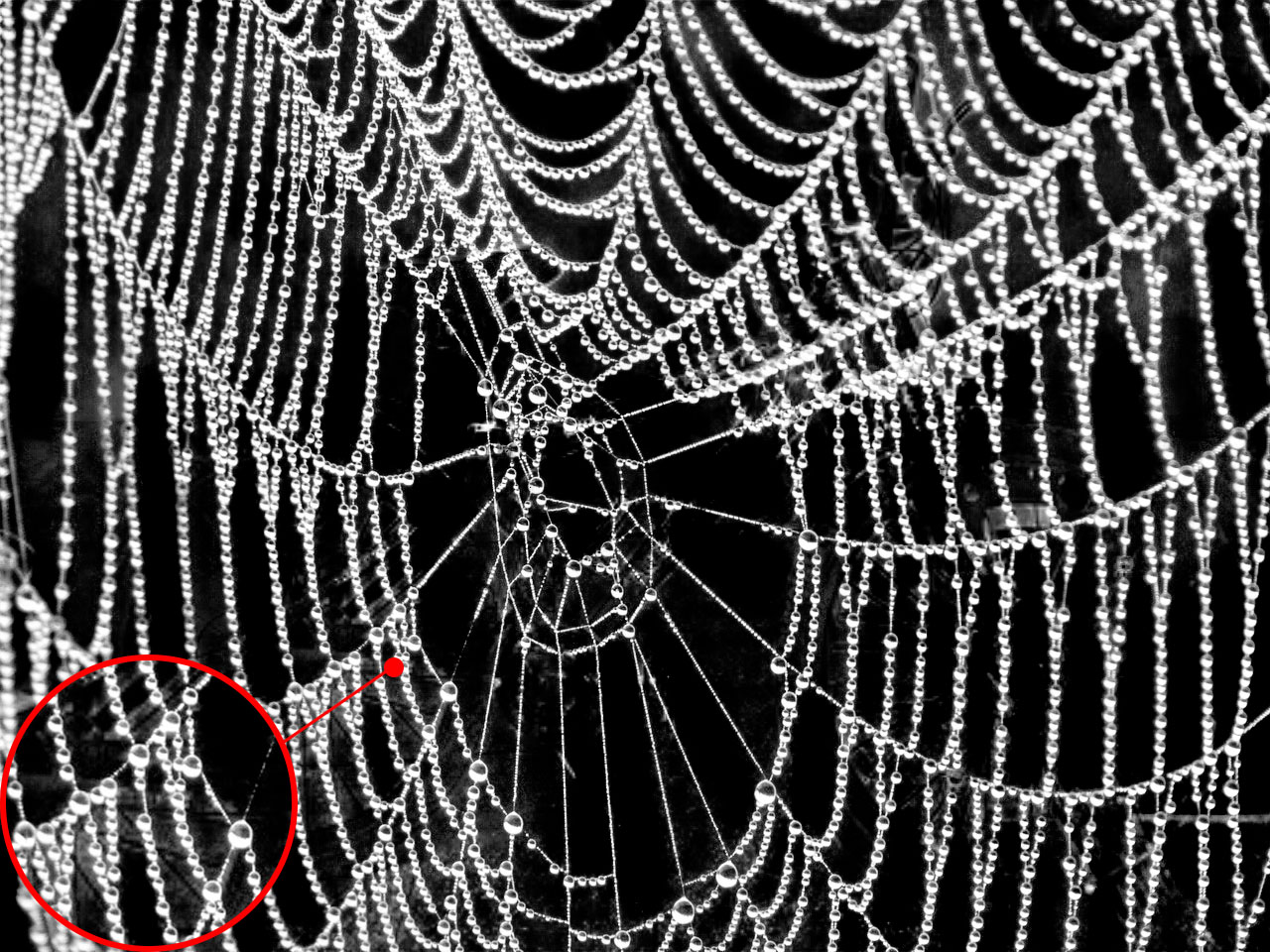} &
			\includegraphics[scale=0.0762]{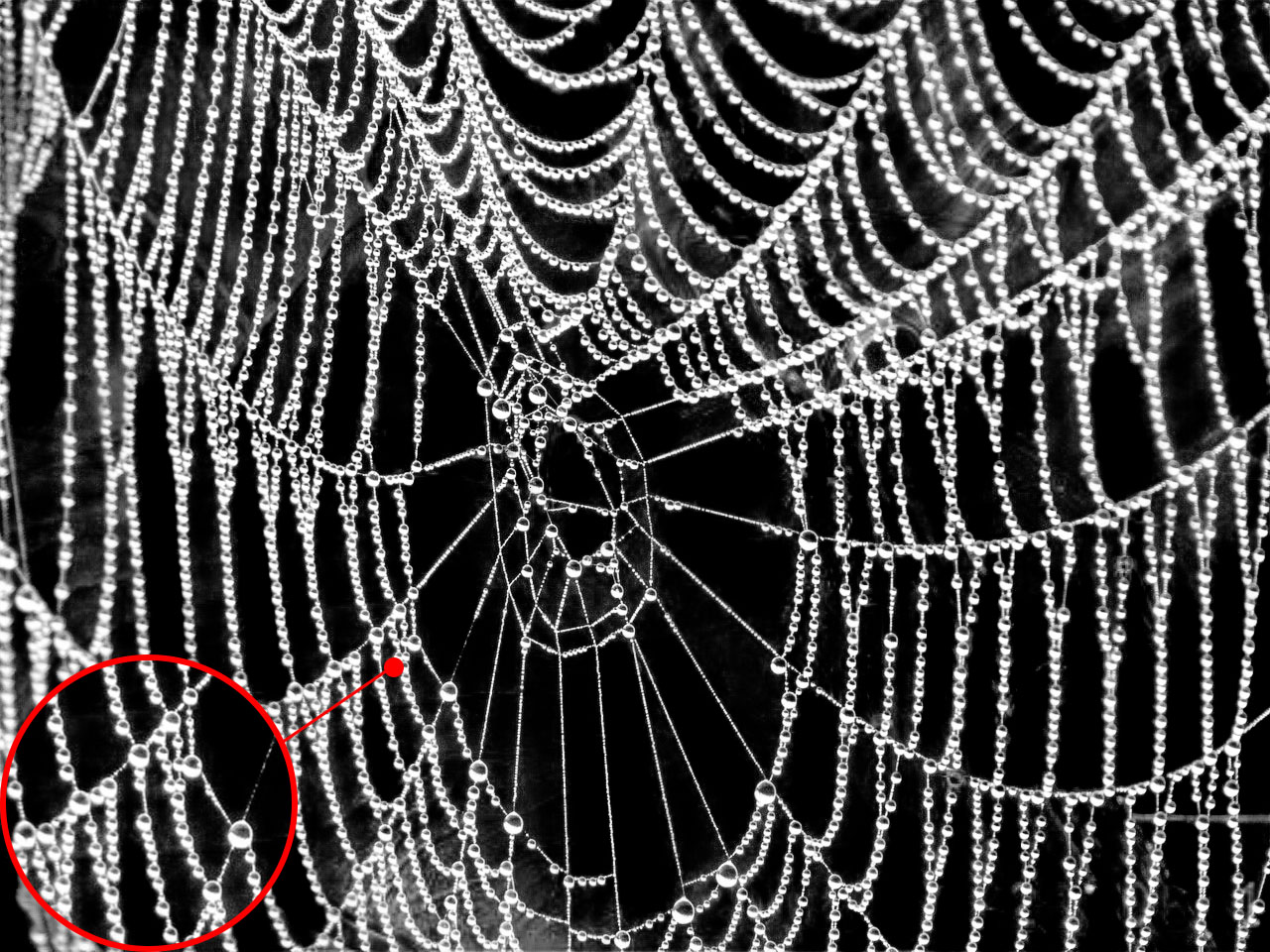} &
			\includegraphics[scale=0.0762]{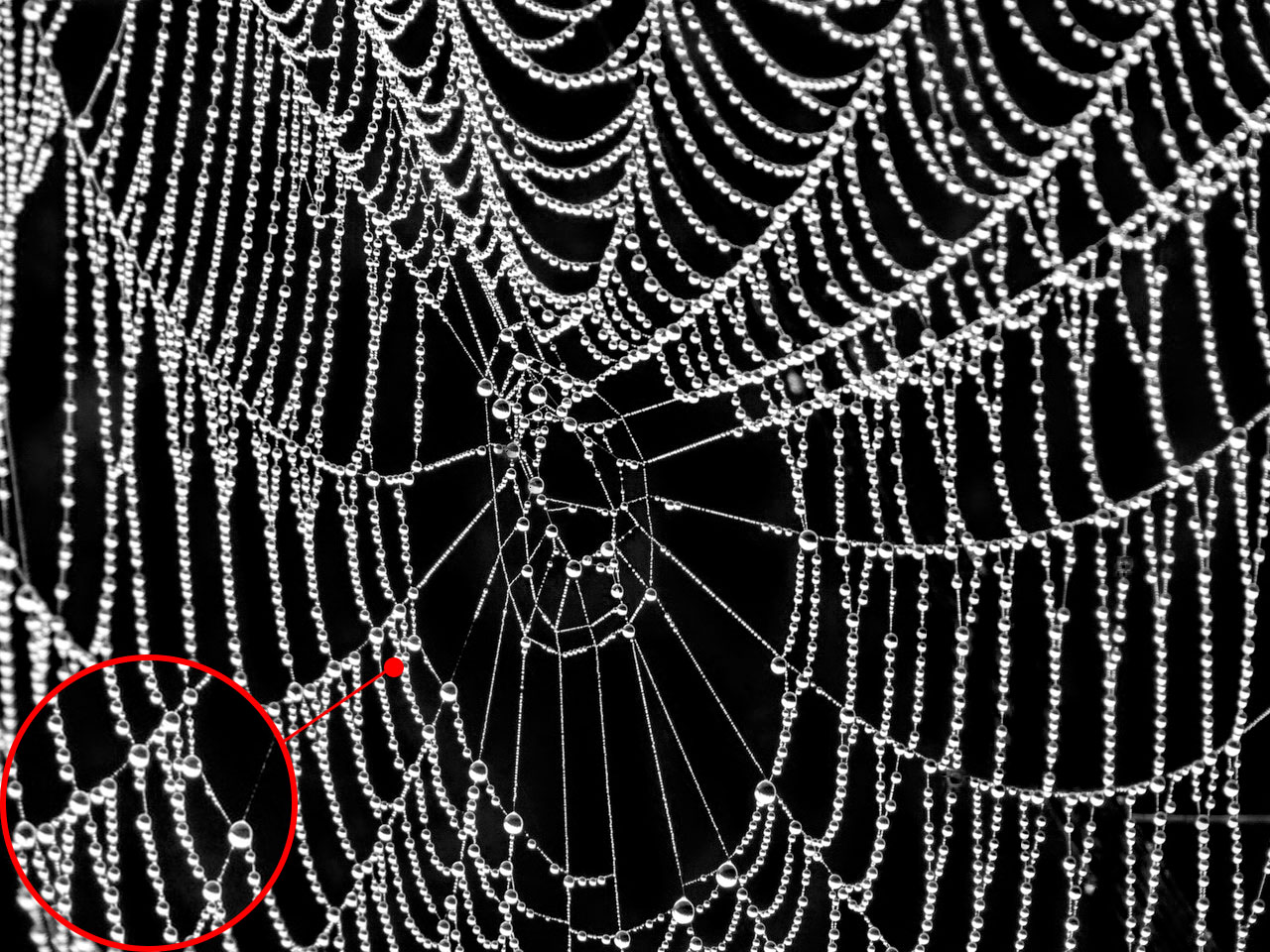} \\
			DIM\cite{Xu2017Deep} & IndexNet\cite{lu2020index} & GCA\cite{li2020natural} & PIIAMatting (Ours) & GT  \\
			
			\includegraphics[scale=0.0762]{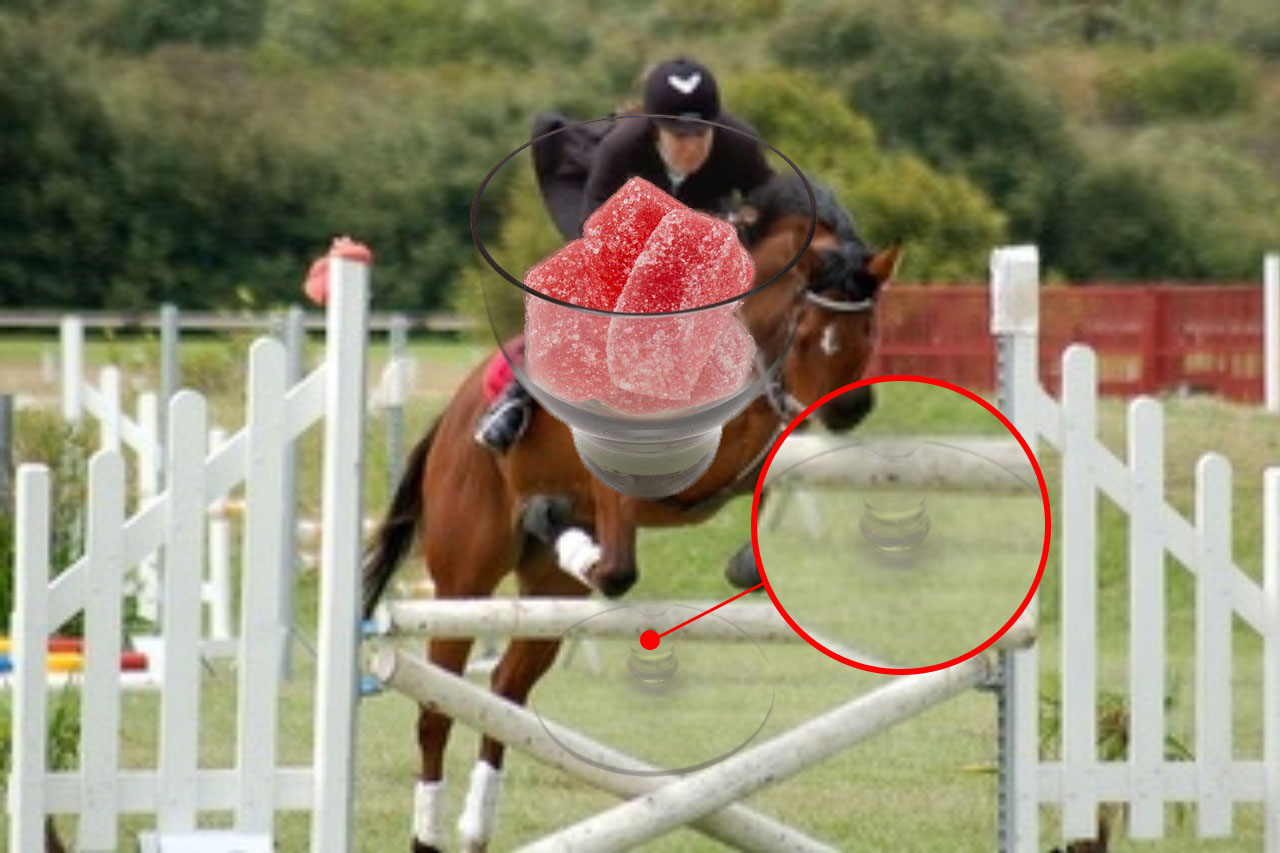} &
			\includegraphics[scale=0.0762]{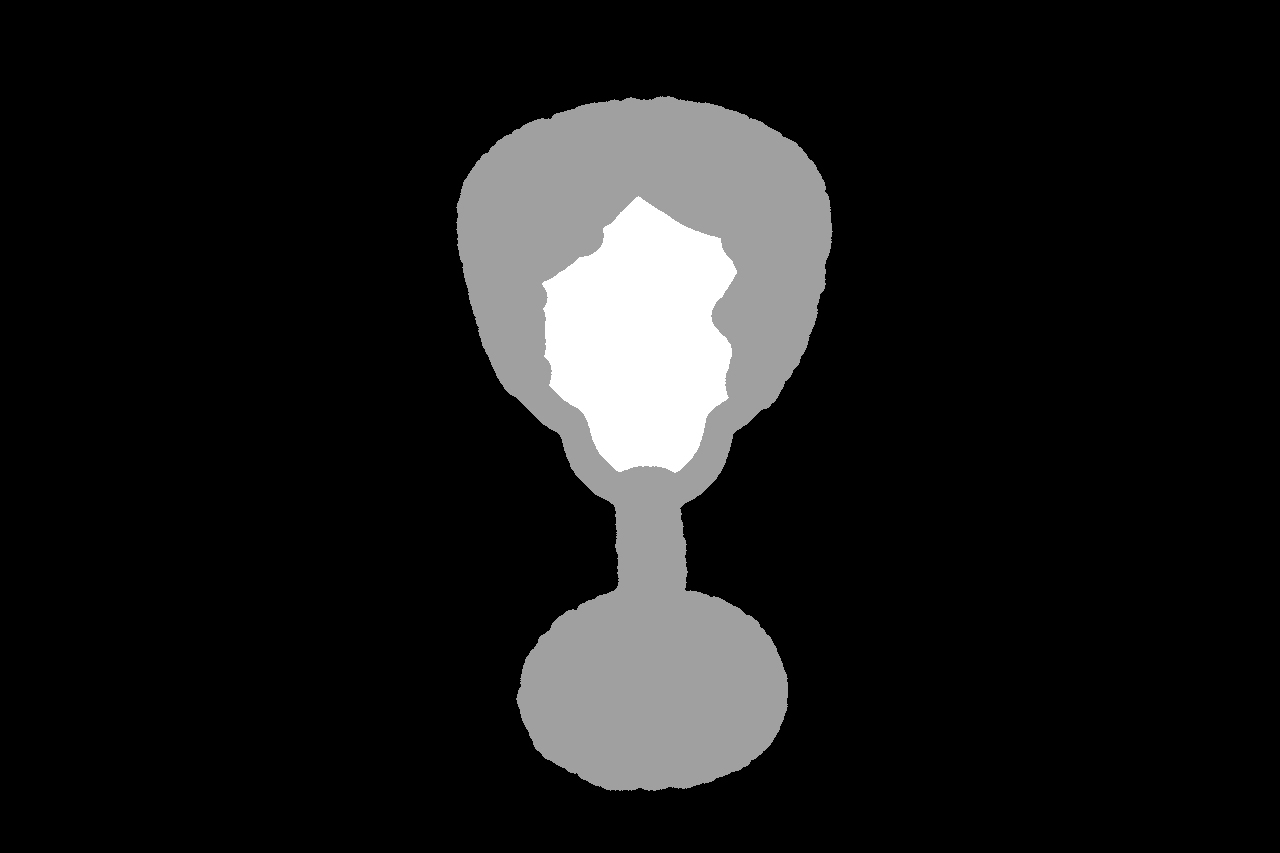} &			
			\includegraphics[scale=0.0762]{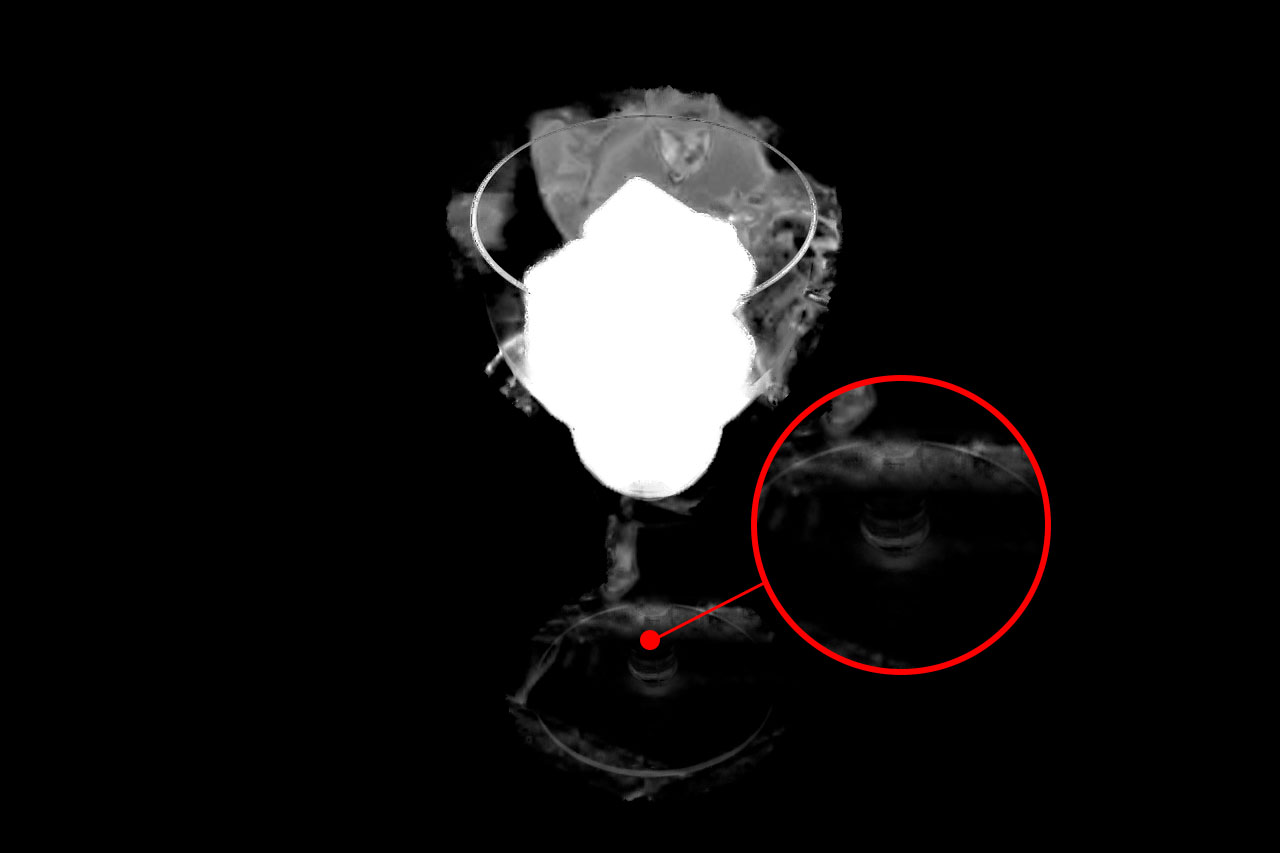} &
			\includegraphics[scale=0.0762]{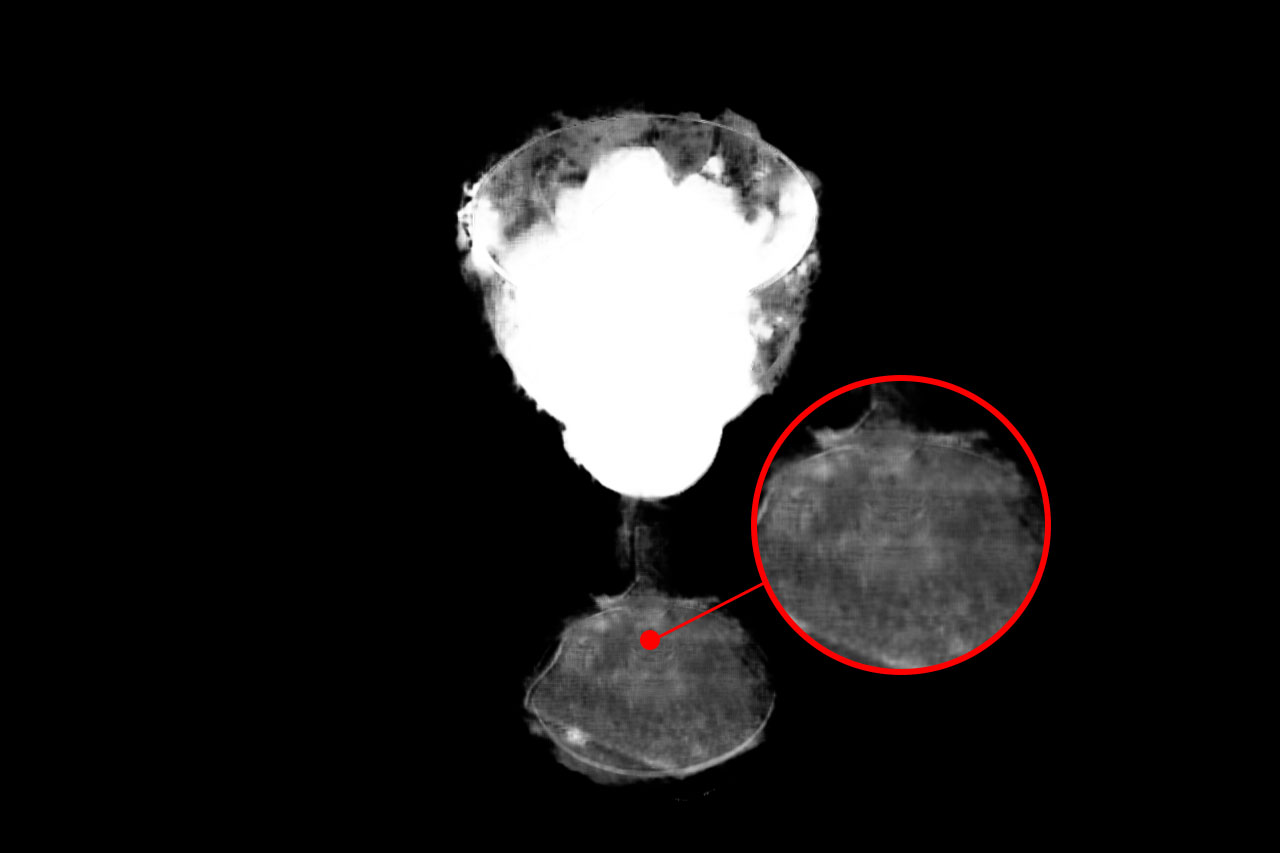} &
			\includegraphics[scale=0.0762]{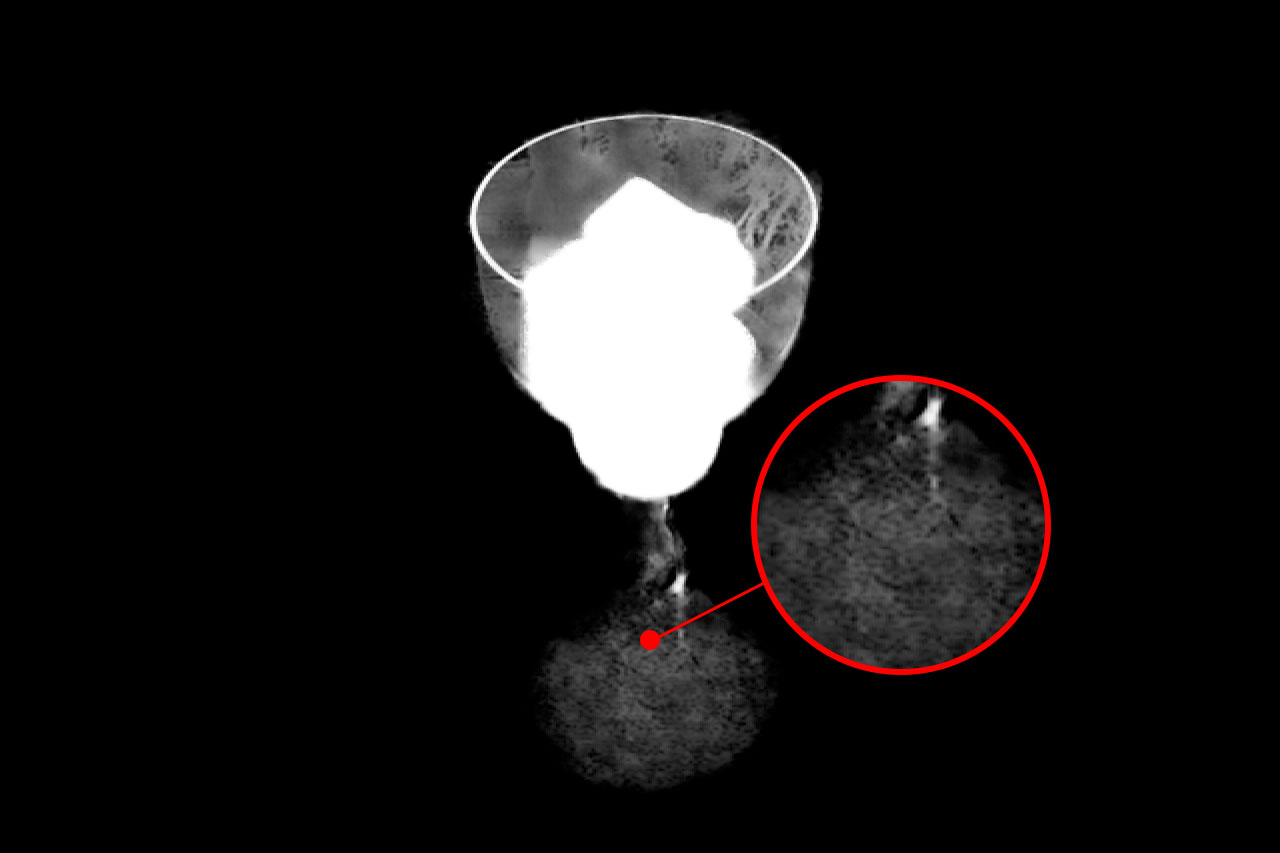} \\
			Image & Trimap & Information-Flow\cite{aksoy2017designing} & AlphaGAN\cite{lutz2018alphagan} & SampleNet\cite{tang2019learning}  \\
			\includegraphics[scale=0.0762]{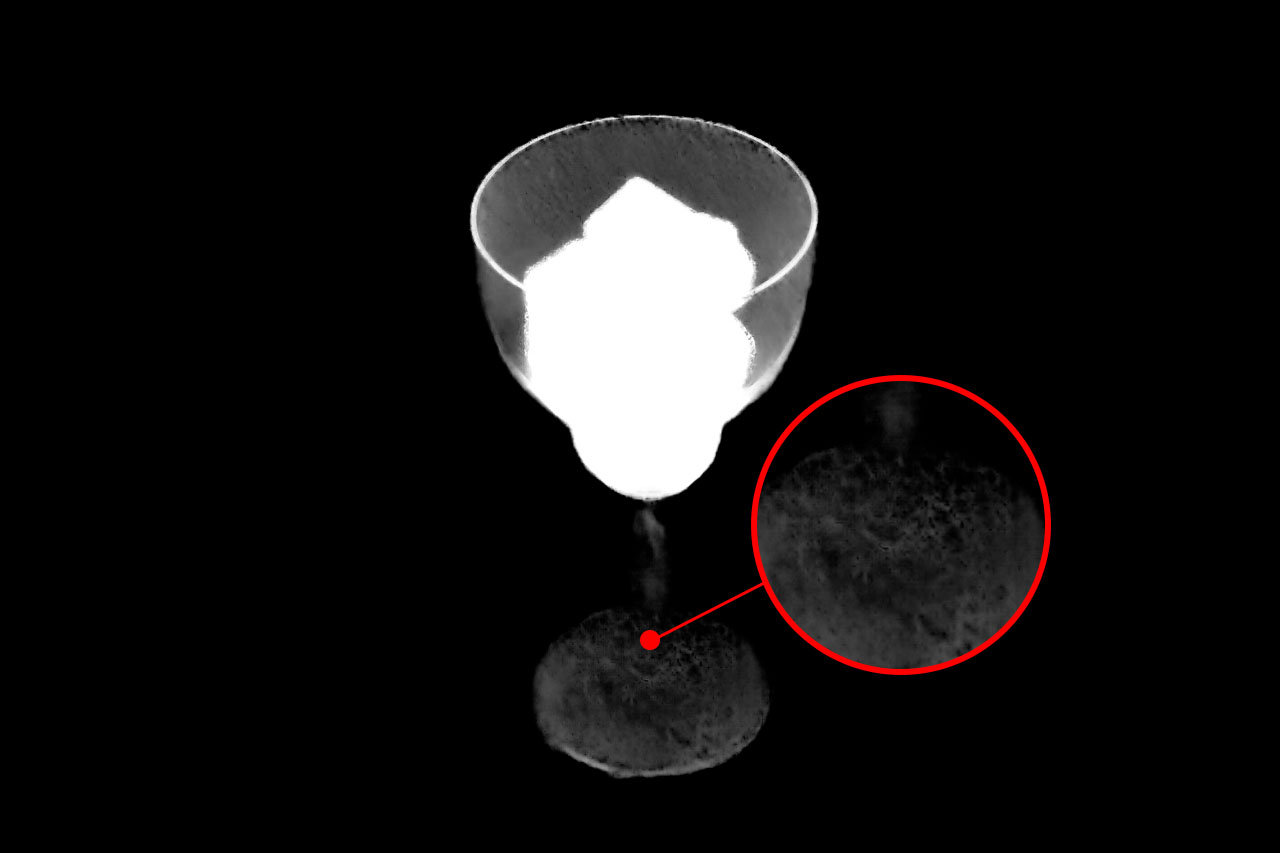} &
			\includegraphics[scale=0.0762]{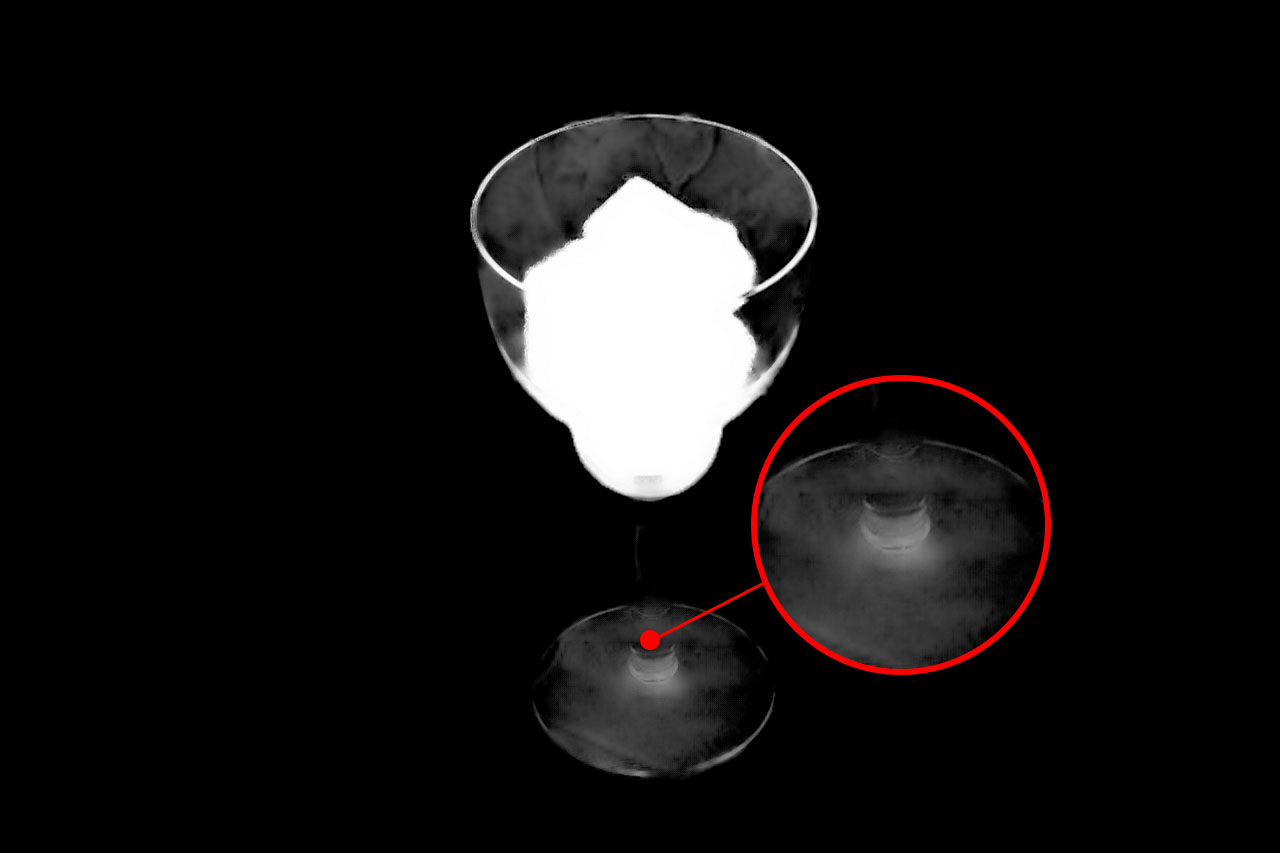} &			
			\includegraphics[scale=0.0762]{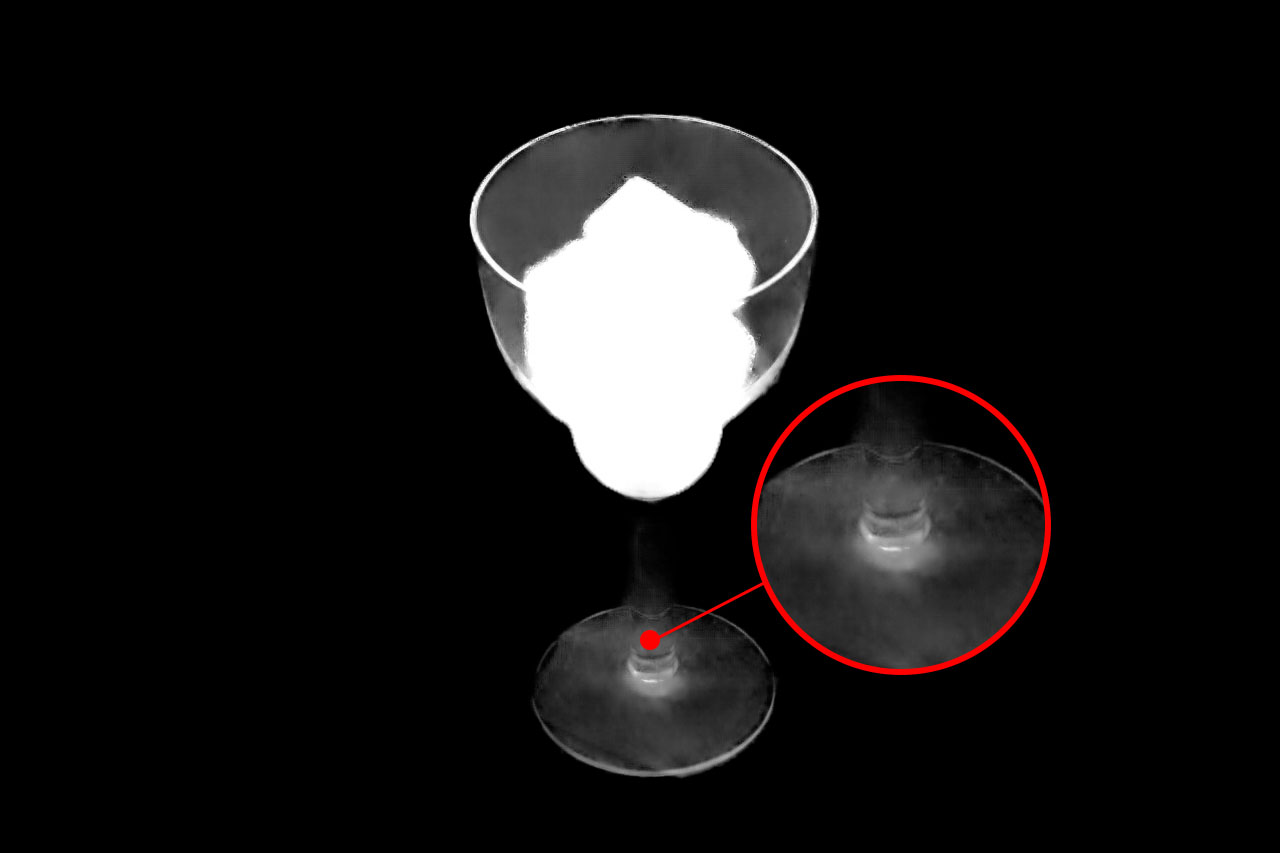} &
			\includegraphics[scale=0.0762]{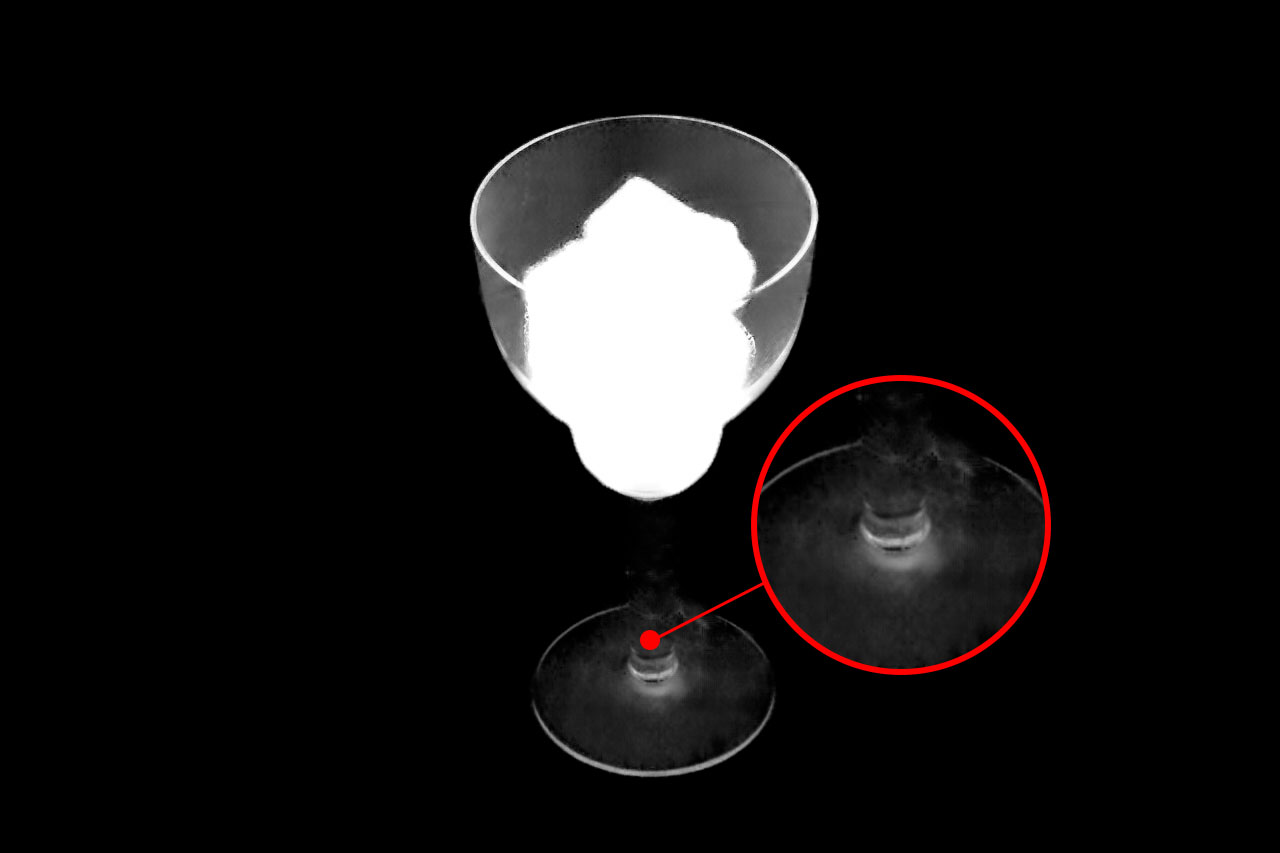} &
			\includegraphics[scale=0.0762]{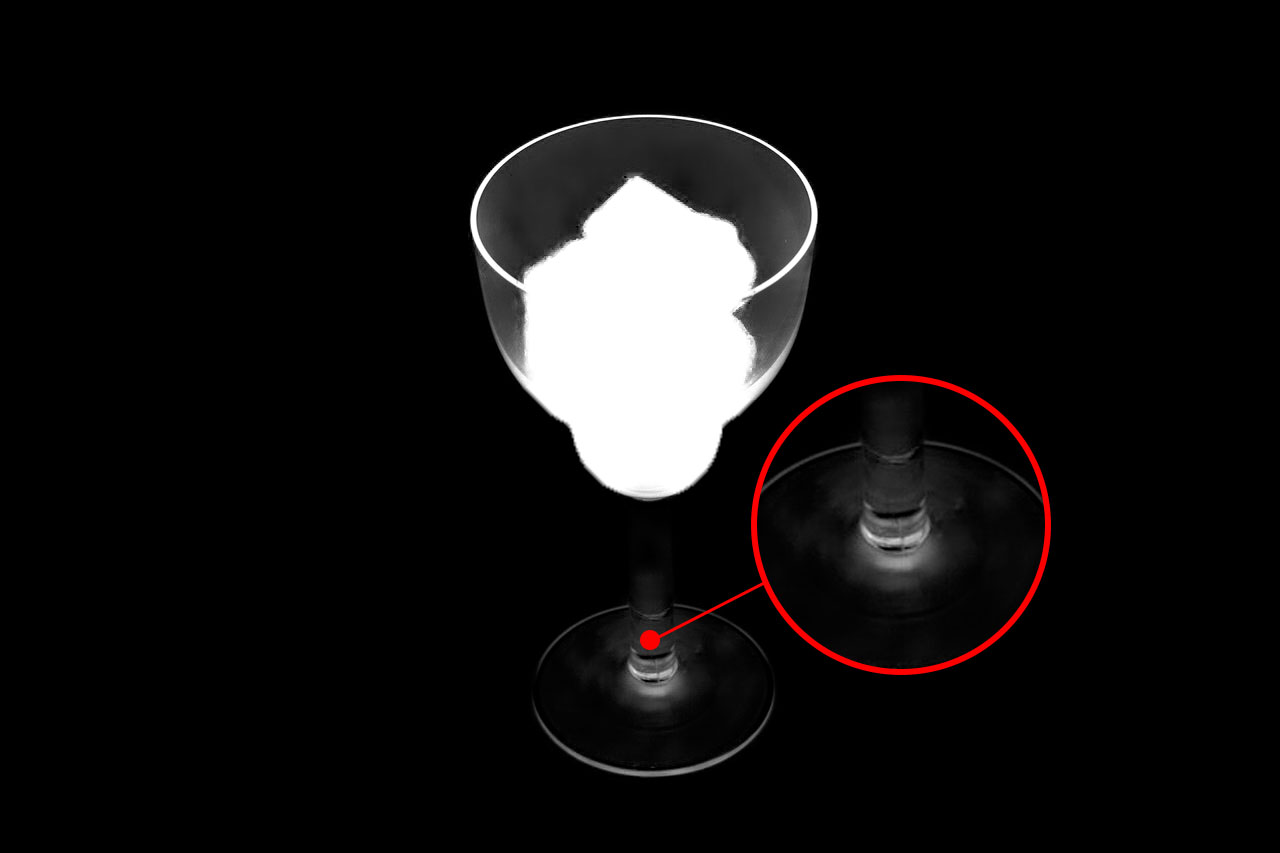} \\
			DIM\cite{Xu2017Deep} & IndexNet\cite{lu2020index} & GCA\cite{li2020natural} & PIIAMatting (Ours) & GT  \\
			 
			\includegraphics[scale=0.0506]{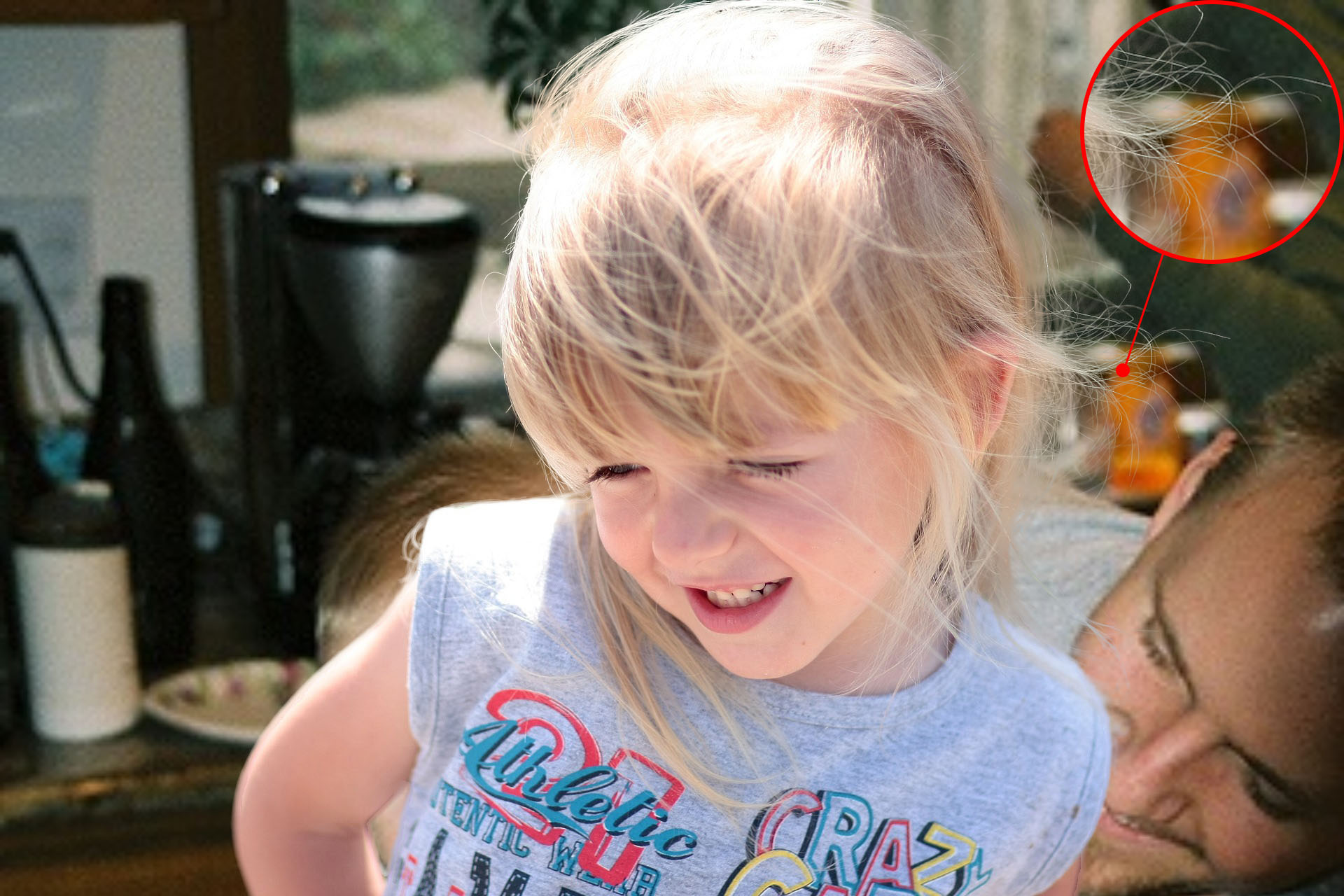} &
			\includegraphics[scale=0.0506]{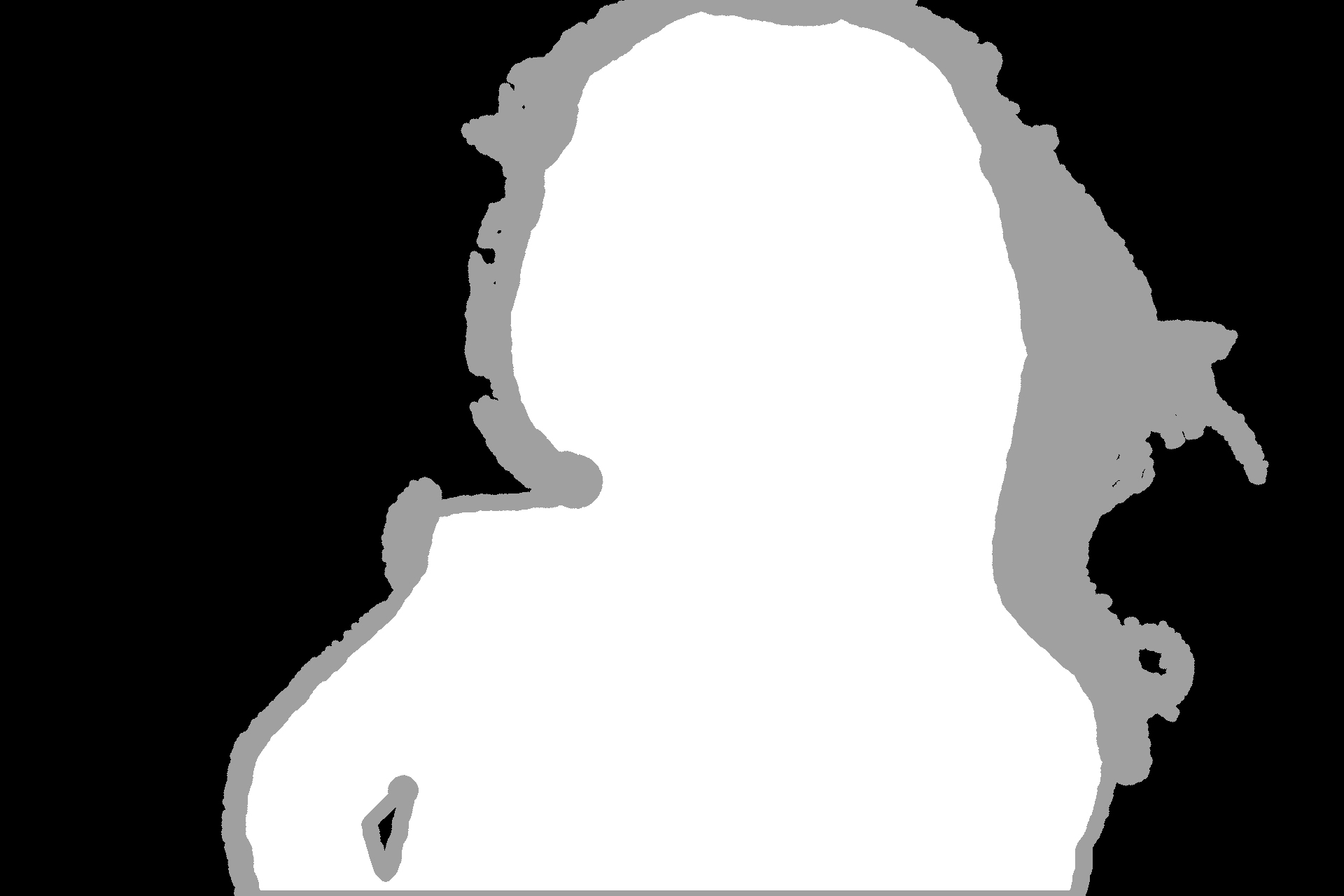} &			
			\includegraphics[scale=0.05065]{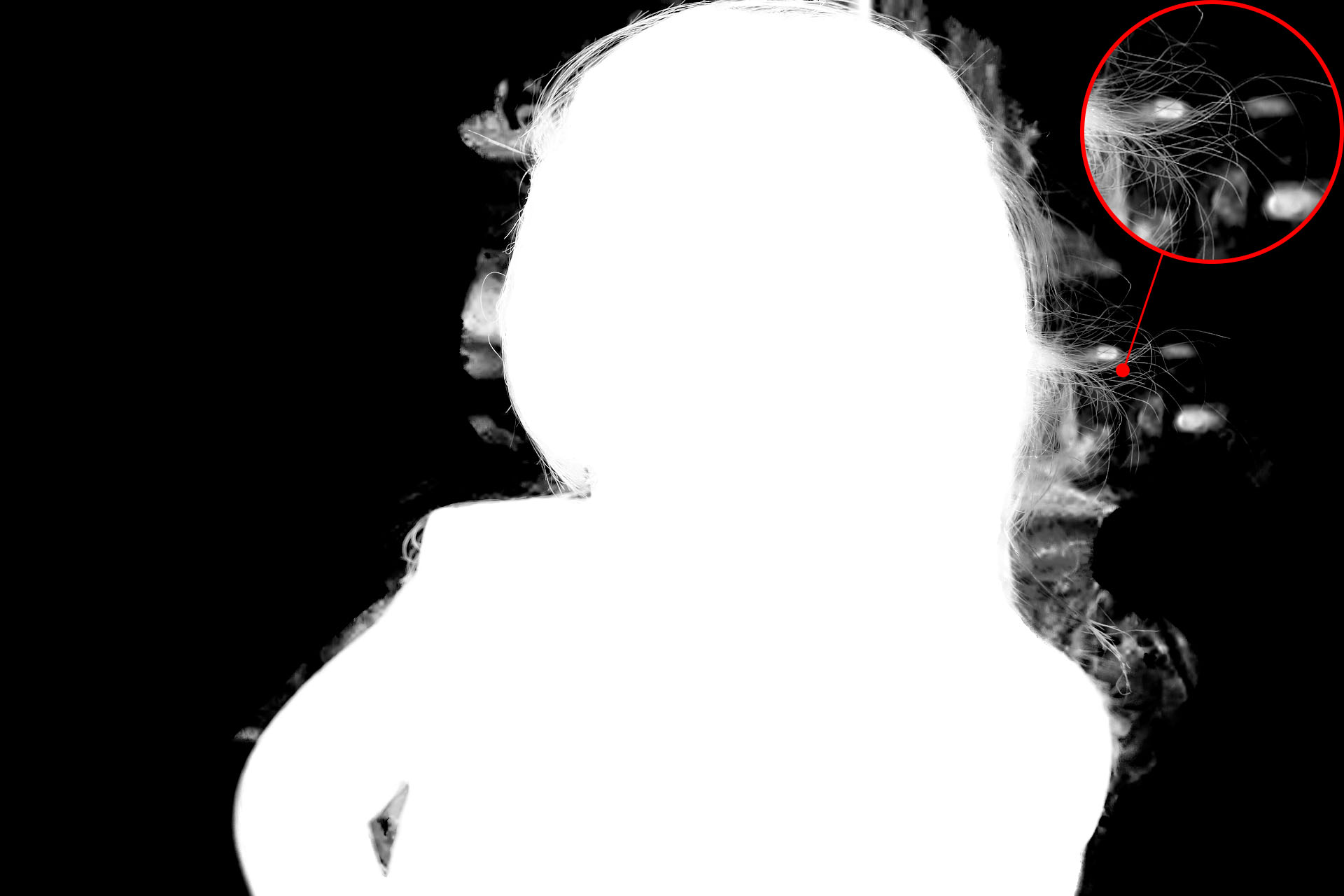} &
			\includegraphics[scale=0.05065]{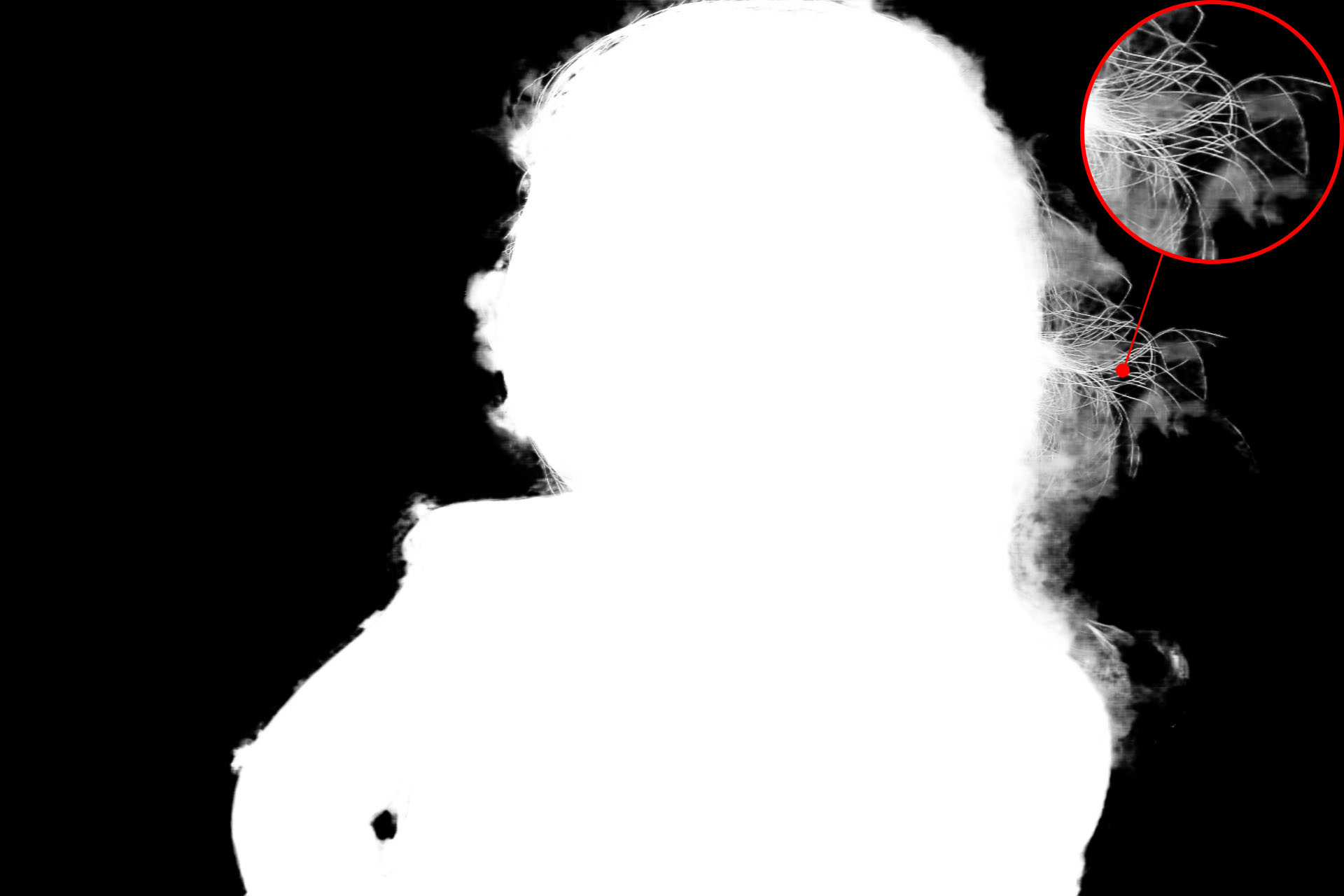} &
			\includegraphics[scale=0.05065]{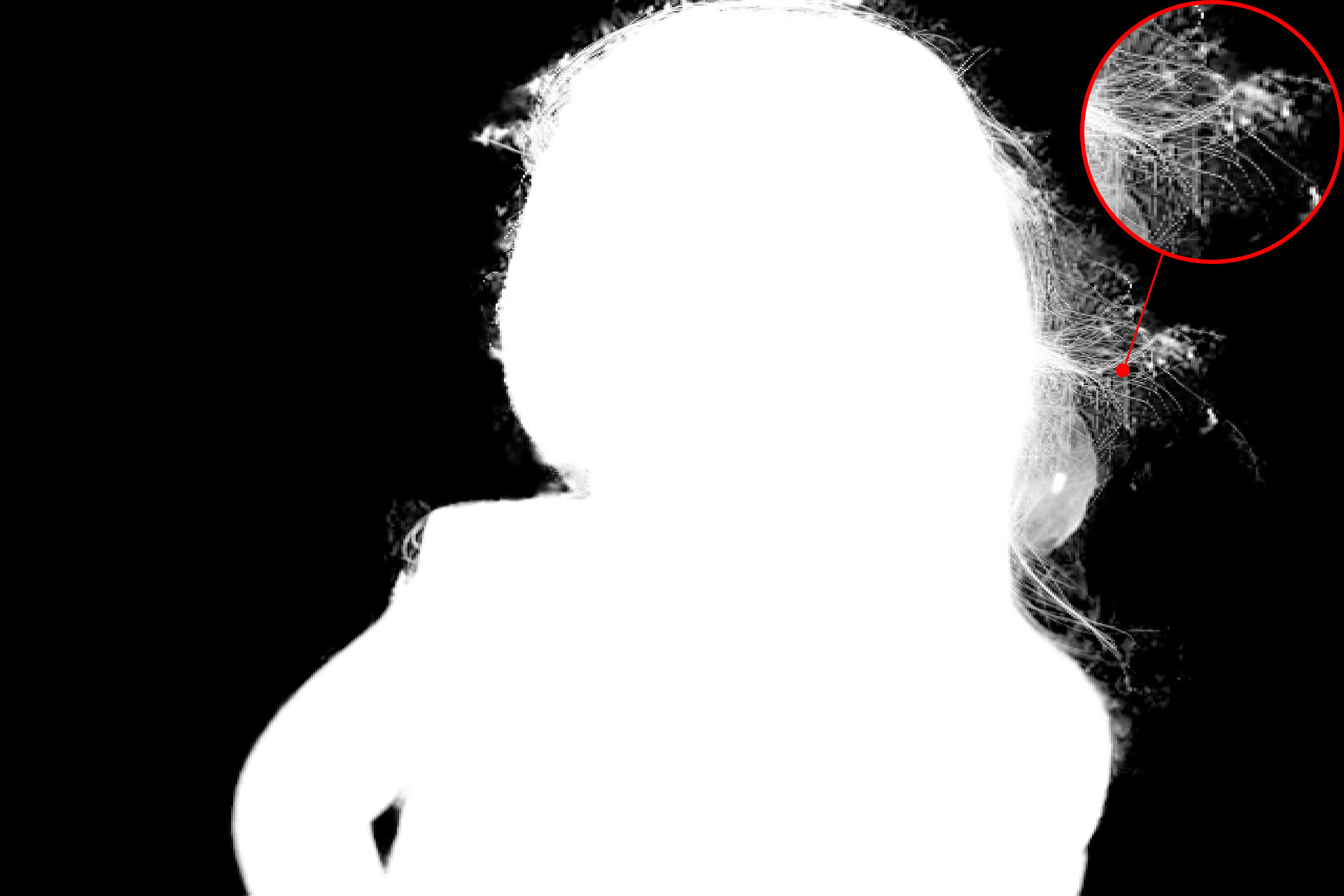} \\
			Image & Trimap & Information-Flow\cite{aksoy2017designing} & AlphaGAN\cite{lutz2018alphagan} & SampleNet\cite{tang2019learning}  \\
			\includegraphics[scale=0.05065]{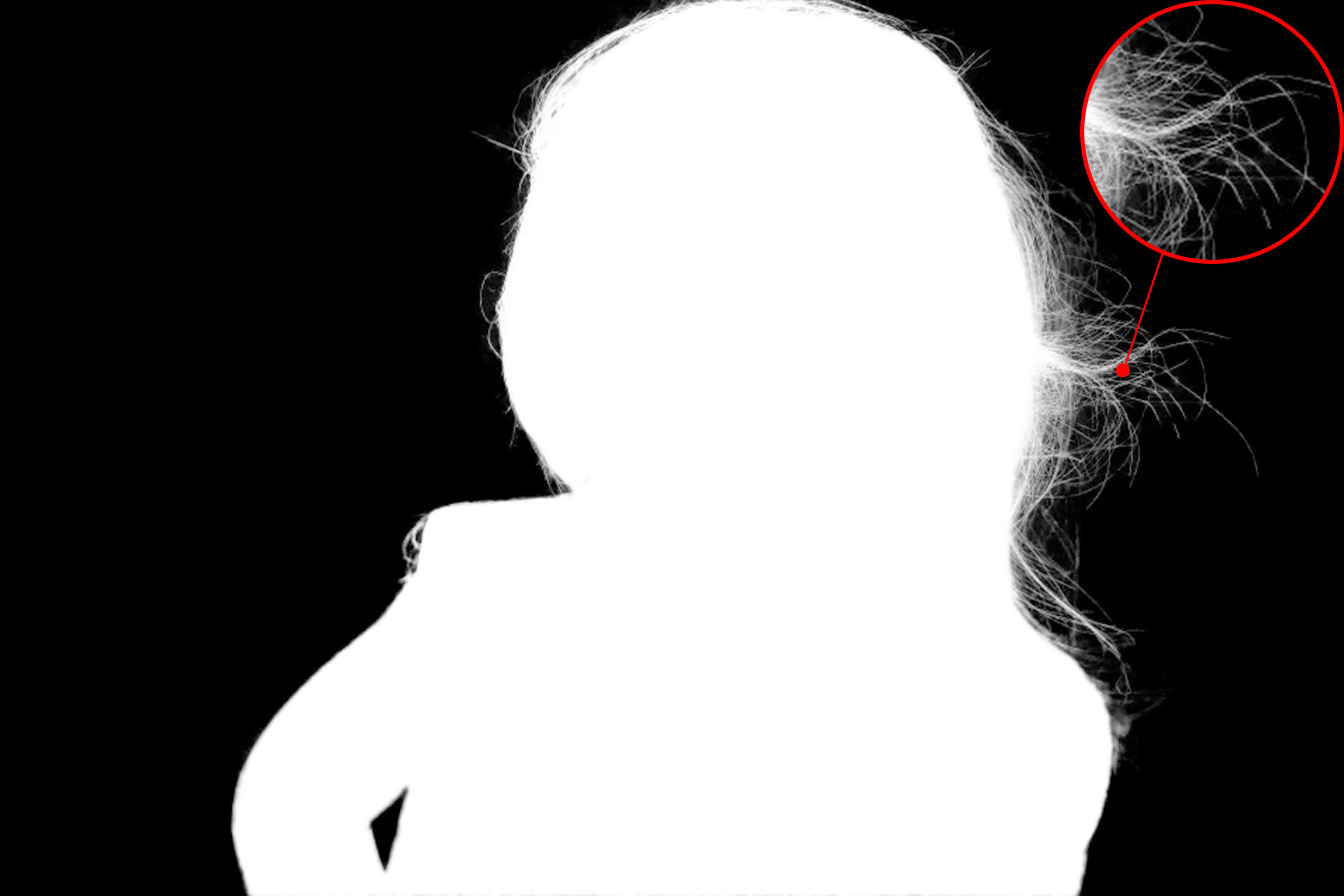} &
			\includegraphics[scale=0.05065]{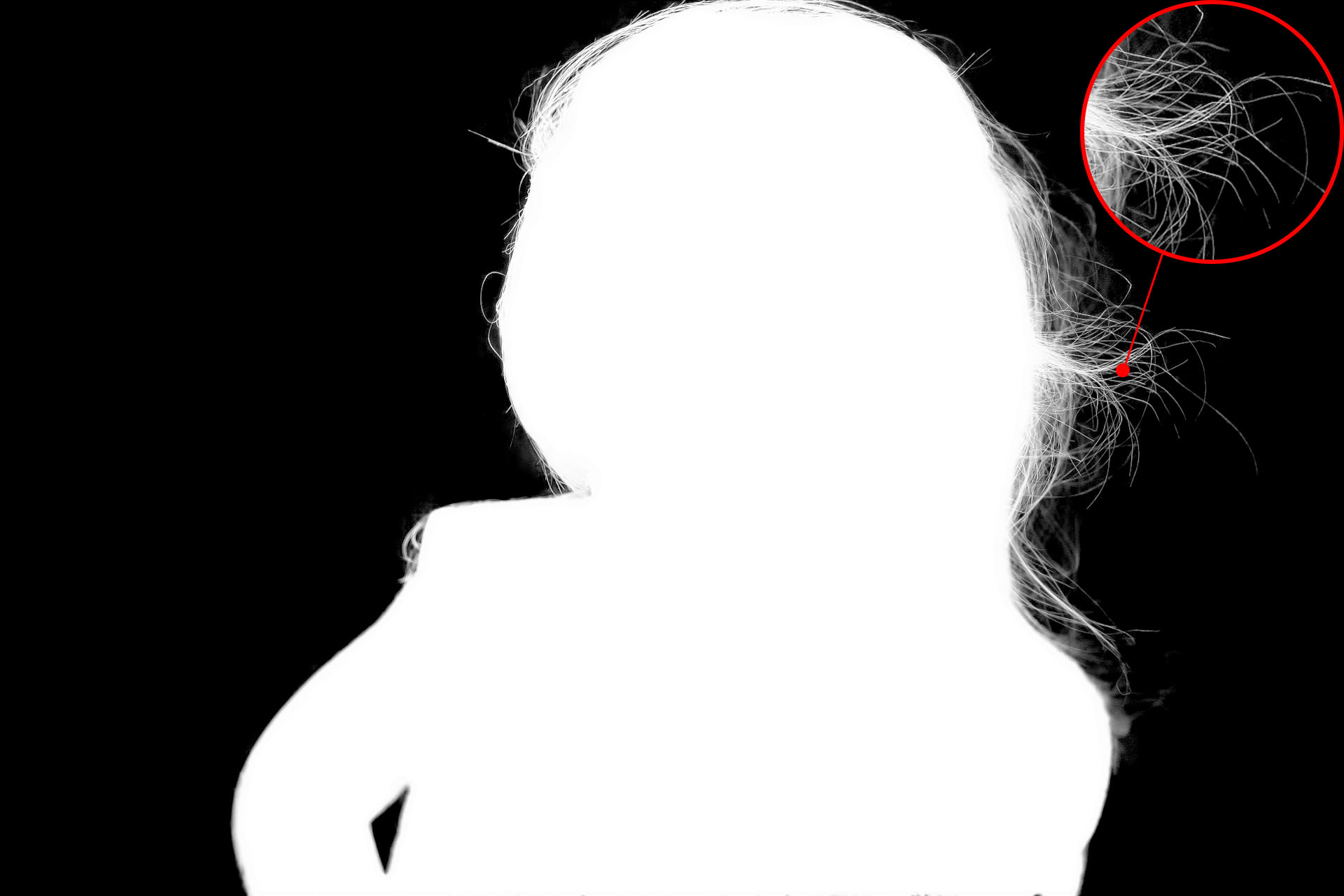} &			
			\includegraphics[scale=0.05065]{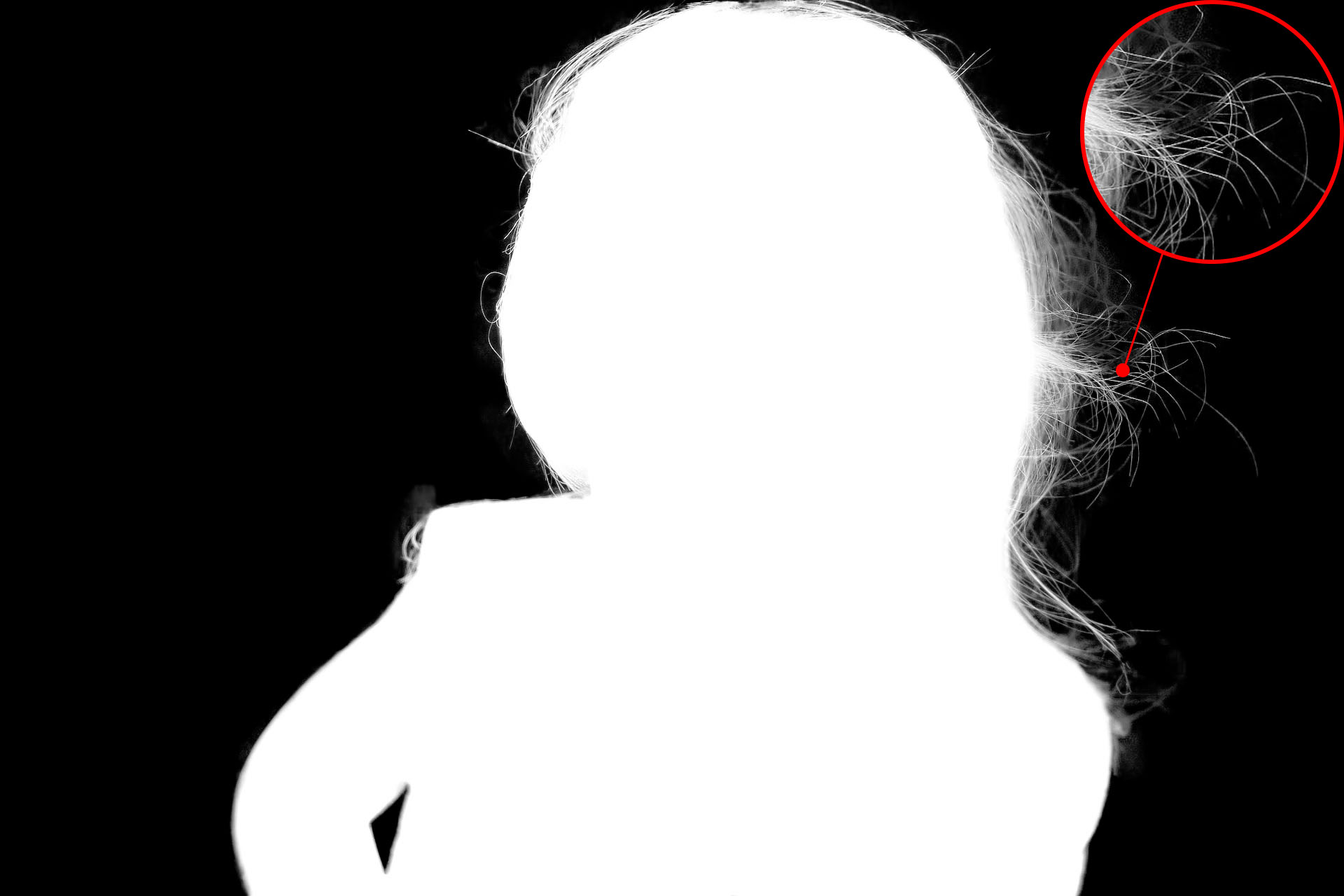} &
			\includegraphics[scale=0.05065]{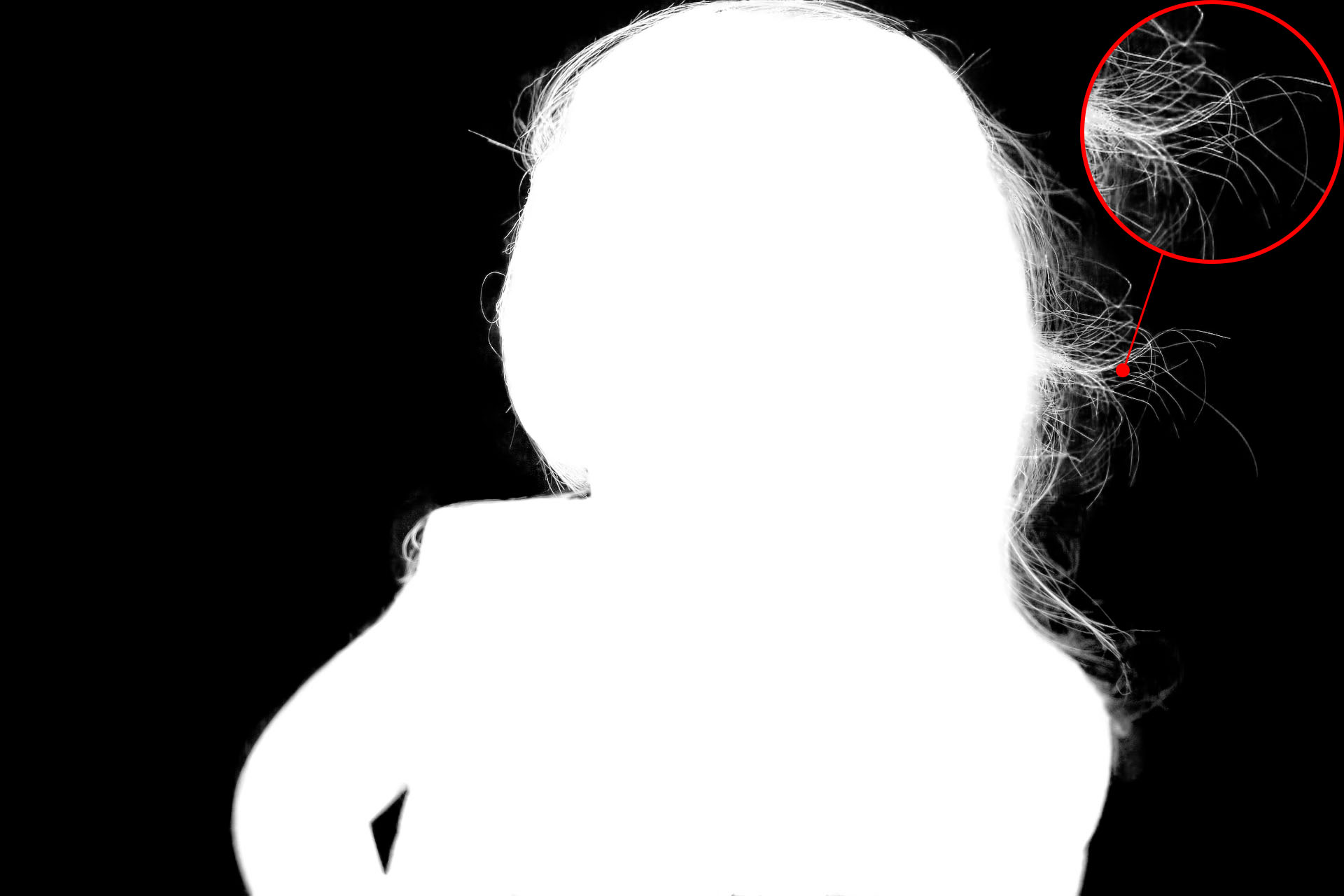} &
			\includegraphics[scale=0.05065]{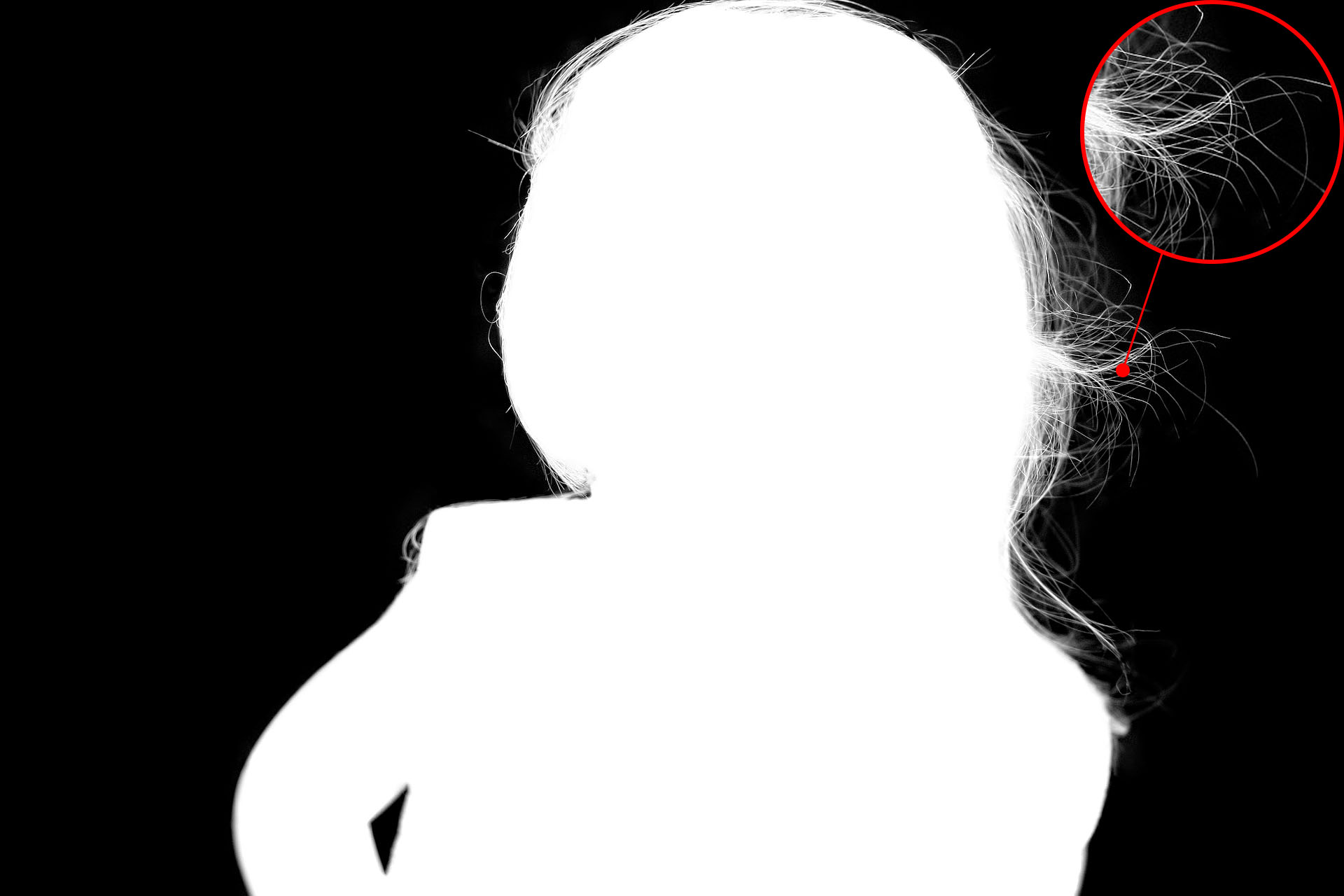} \\
			DIM\cite{Xu2017Deep} & IndexNet\cite{lu2020index} & GCA\cite{li2020natural} & PIIAMatting (Ours) & GT  \\

	\end{tabular}}
	\vspace{-1mm}
	\caption{Qualitative comparisons on the Adobe Composition-1k\cite{Xu2017Deep} test set.}
	\vspace{-3mm}
	\label{fig:visual_adobe}
\end{figure*}

\textit{Information Aggregation Module:~}
\label{ssec:iam}
The IAM  also takes two inputs, that is $F_{iam}^{i-1} \in R^{\frac{H}{2}*\frac{W}{2}*C}$  generated from the preceding IAM and $F_{imm}^{i} \in R^{H*W*C}$ 
obtained from IMM. 
Specifically, the IAM superimposes these two features by element-wise summation and then extracts the most critical features by element-wise product. 
The operations can be formally defined  as follows:

\begin{equation}
  \mathcal{F}^{i}_{iam}=\left\{
  \begin{aligned}
    (\mathcal{F}^{i}_{imm}+T(\mathcal{F}_{aspp}))*T(\mathcal{F}_{aspp}) ,\qquad  i=top.\\
    (\mathcal{F}^{i}_{imm}+T(\mathcal{F}^{i-1}_{iam}))*T(\mathcal{F}^{i-1}_{iam}),\qquad i=else . \\
  \end{aligned}
  \right.
\end{equation}
where \textit{$T(\cdot)$}, \textit{ i}, and \textit{$*$} hold the same meaning in IMM and IAM, and \textit{top} refers to the same depth as block4 in ResNet-50. As shown in the blue box in Fig.~\ref{fig:pipeline}, we adopt the output features of ASPP\cite{chen2018encoder} as the initialization for the whole decoder process, and it is sufficient to directly aggregate the initialization with features from the IMM since  features from the encoder have been matched through our IMM. 

In this way, via the utilization of relatively high feature information by our IA strategy, valuable information can be effectively matched and retained between adjacent cascading features, resulting in a more accurate aggregation of information at the sampling stage.

\textbf{Muti-Scale Refinement Module:}
\label{ssec:msr}
In the encoder-decoder structure, the incorporation of trimap helps to guide the model to learn significant information. However, the unknown regions only account for a small portion of the overall image, leading to the delicate details under natural scenes that are still difficult to be learned. 
Therefore, some information may be potentially lost after the process of the first stage.
 Inspired by \cite{Xu2017Deep, cai2019disentangled}, our model is subsequently extended with a Multi-Scale Refinement module further to  improve the quality of the estimated alpha mattes.
As depicted in the orange box in Fig.~\ref{fig:pipeline}, we apply Atrous Convolutions\cite{chen2018encoder} with different kernels to the preliminary alpha mattes obtained from the encoder-decoder stage  to explore multi-scale features. The original image was also introduced at this stage to extract rich location and colour information to guide the learning process. 
The accuracy increase improved by our MSR can also be seen in Tab.~\ref{tab:ablation_networks}.
\subsection{Loss Function}
\label{ssec:loss}
Considering the Dynamic Gaussian Modulation mechanism, we abandon the Composition-L1 loss in DIM\cite{Xu2017Deep} and only resort to the Alpha prediction loss to achieve effective alpha matte optimization in the PIIAMatting. Our loss function is defined as follows:
\begin{align}
  \mathcal{L}^{i}_{alpha} = \mathcal{R}_{g}^{i}*\left | \alpha_{p}^{i}-\alpha_{g}^{i}  \right |.
  \end{align}
 where $i$ denotes the pixel position,  $*$ means the element-wise product, and
 the ${R}_{g}^{i}$ represents the response coefficient at which pixel i obtained according to the prior information in the Ground Truth. The $\alpha_{p}^{i}$ and $\alpha_{g}^{i}$ are the predicted alpha matte and ground truth alpha matte at pixel i, separately.
  
\section{Experimental Results}
	\vspace{-1mm}
In this section, we conducted extensive experiments and evaluated our PIIAMatting on three challenging datasets to prove the effectiveness of it: (1) Alphamatting.com dataset\cite{rhemann2009perceptually}, (2) Adobe Composition-1K dataset\cite{Xu2017Deep} and (3) Distinctions-646 dataset\cite{qiao2020attention}. Firstly, we compare our PIIAMatting quantitatively and qualitatively with the current state-of-the-art methods. Then, we take Composition-1K\cite{Xu2017Deep} as an example and carry out the experimental analysis of the efficacy of each component in our PIIAMatting. Finally, we apply our PIIAMatting to some Real-World images to further validate our method and predict their alpha mattes.
\subsection{Datasets}
\label{ssec:datasets}
The first dataset is the Alphamatting.com dataset\cite{rhemann2009perceptually}, which is the existing benchmark for image matting. It consists of 27 images with user-defined trimaps and alpha mattes, and 8 test with 
    three different kinds of trimaps, namely, "small", "large", and "user".
The second dataset is  Composition-1K\cite{Xu2017Deep}, which is the first public large-scale image matting dataset. The training set is composed of 431 foreground objects with the corresponding alpha mattes. Each foreground image is combined with 100 distinct backgrounds from MS COCO\cite{lin2014microsoft} to form the new synthesis. The test set consists of 50 pair of foreground objects and the corresponding alpha matte, and each foreground is composited with 20 different backgrounds from PASCAL VOC\cite{everingham2010pascal}. The third dataset is the Distinctions-646, which is the  current biggest dataset for image matting. It provides 646 distinct foreground images with the corresponding alpha mattes and takes the same rule as DIM\cite{Xu2017Deep} to generate new images. We completely follow the composition rule provided by \cite{Xu2017Deep} and \cite{qiao2020attention} when using the datasets.
\subsection{Evaluation metrics}
\label{ssec:metrics}
We evaluate the alpha mattes following four common quantitative metrics: the sum of absolute differences (SAD), mean square error (MSE), gradient (Grad), and connectivity (Conn) error proposed by \cite{rhemann2009perceptually}. 
 The first two metrics  measure the difference in absolute pixel space, while the last two metrics mean to weigh the visual quality of generated alpha mattes. Anyway, the lower the values of all metrics, the better the predicted alpha mattes.
     
\subsection{Implementation Details}
\label{ssec:implementation}
Our method is implemented with PyTorch\cite{pytorch} Toolbox. For training, we first randomly crop the images with a crop size of 320$\times$320, 512$\times$512, and 640$\times$640, and then resize the images to 320$\times$320. Trimaps are generated using either dilation/corrosion\cite{Xu2017Deep} or Euclidean distance\cite{lu2020index} based on random selection, and the kernel sizes are randomly selected in [5, 25]. We also perform the data augmentation, including random scaling, flipping, and rotation between -60 and 60 degrees. For all experiments, we train for 50 epochs with a batch-size of 64. The Adam\cite{adam} optimizer was adopted with the initial learning rate of 0.001, and it was adjusted to 0.0001 and 0.00001 at the 30th and 40th epochs, respectively. 
The $\mu$ and  $\sigma^2$ were initially set to 0.5 and 0.25.
 With the increase of iterations, the deviation $\sigma^2$ is adapted to make the Gaussian distribution flatter so that the model gradually treats all samples more fairly.
 The training time of our network on a GPU server (configuration: XEON Gold-6240 CPU, 32G RAM, and 3 Tesla V100 graphics card) is 3 days for the DIM\cite{Xu2017Deep} dataset and 3.5 days for the Distinctions-646\cite{qiao2020attention} dataset.
 During inference, the full-resolution input images with trimaps are concatenated as a 4-channel input  fed into the network.

\begin{figure*}[h]
	\setlength{\tabcolsep}{1pt}\small{
		\begin{tabular}{ccccc}			
			\includegraphics[scale=0.152]{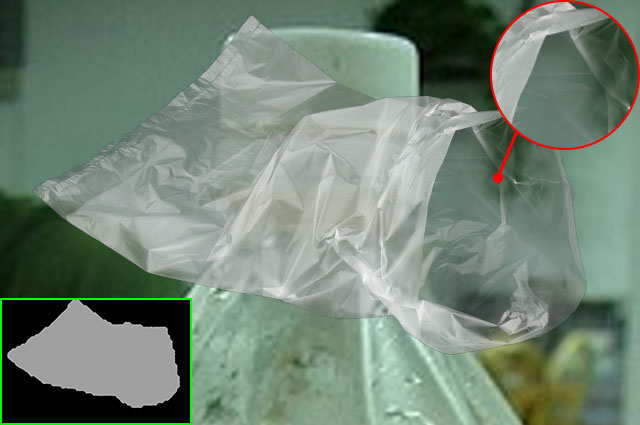} &
			\includegraphics[scale=0.152]{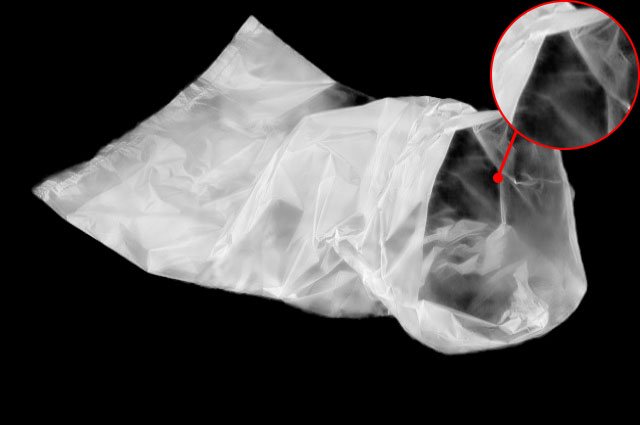} &	
			\includegraphics[scale=0.152]{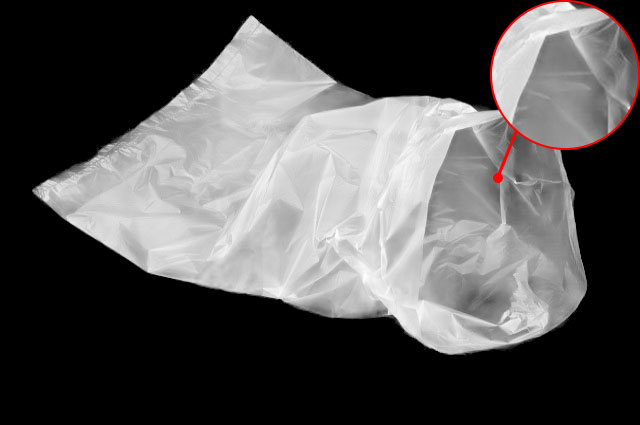} &
			\includegraphics[scale=0.152]{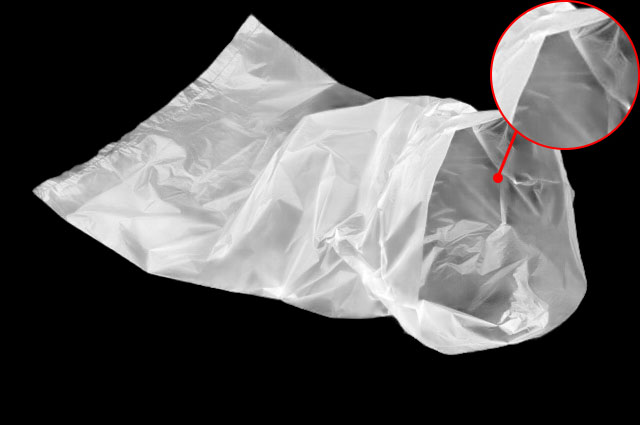} &
			\includegraphics[scale=0.152]{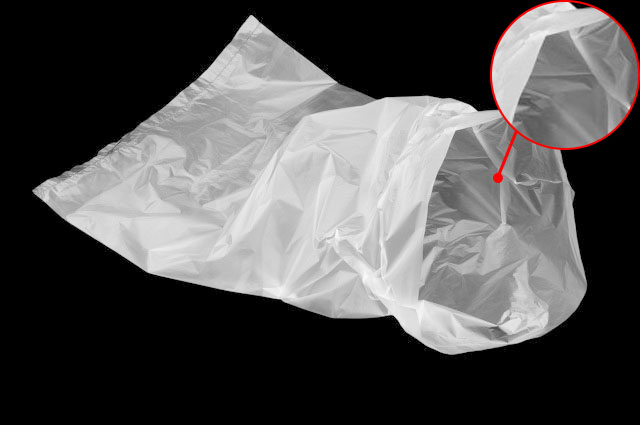} \\

			\includegraphics[scale=0.0763]{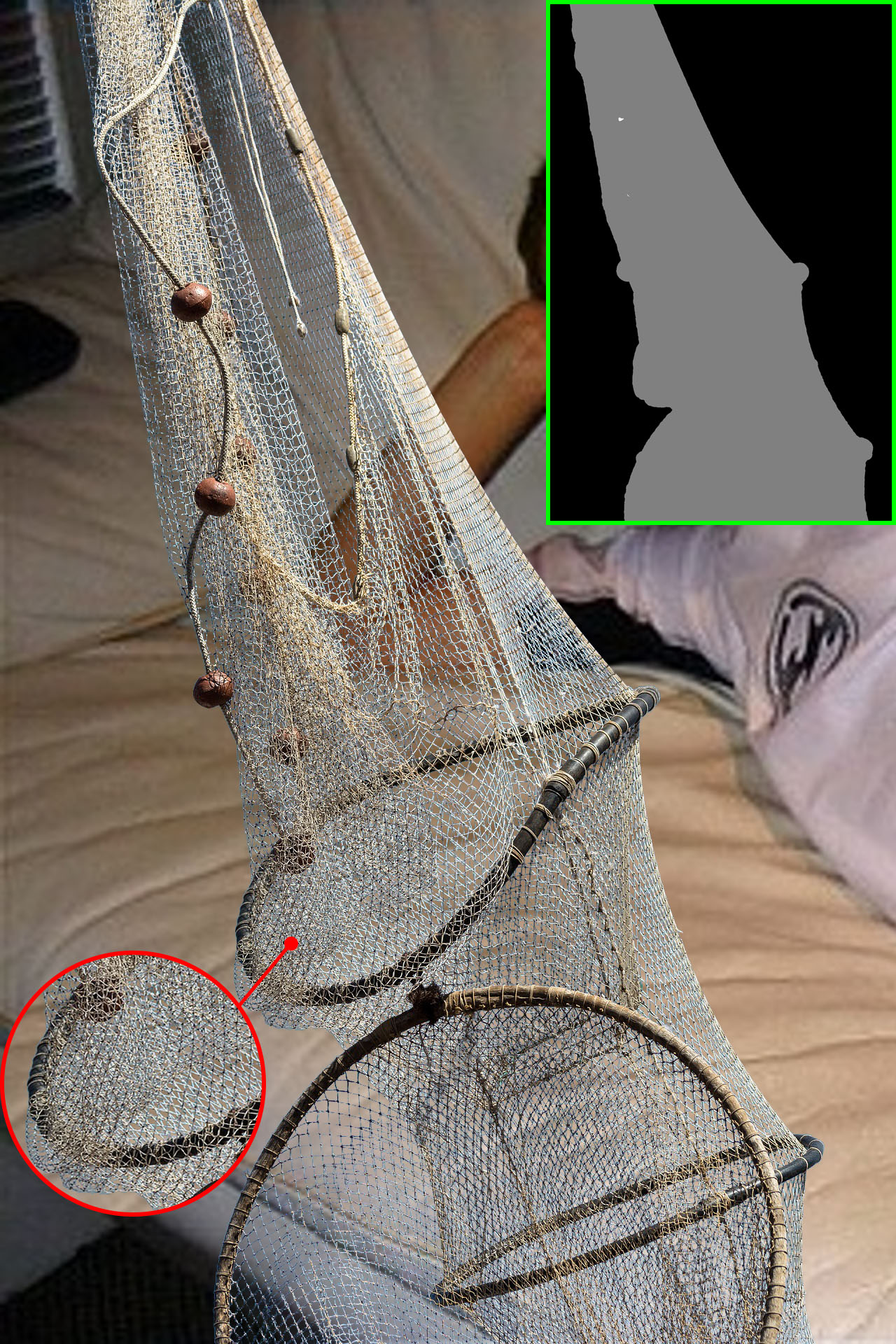} &
			\includegraphics[scale=0.0763]{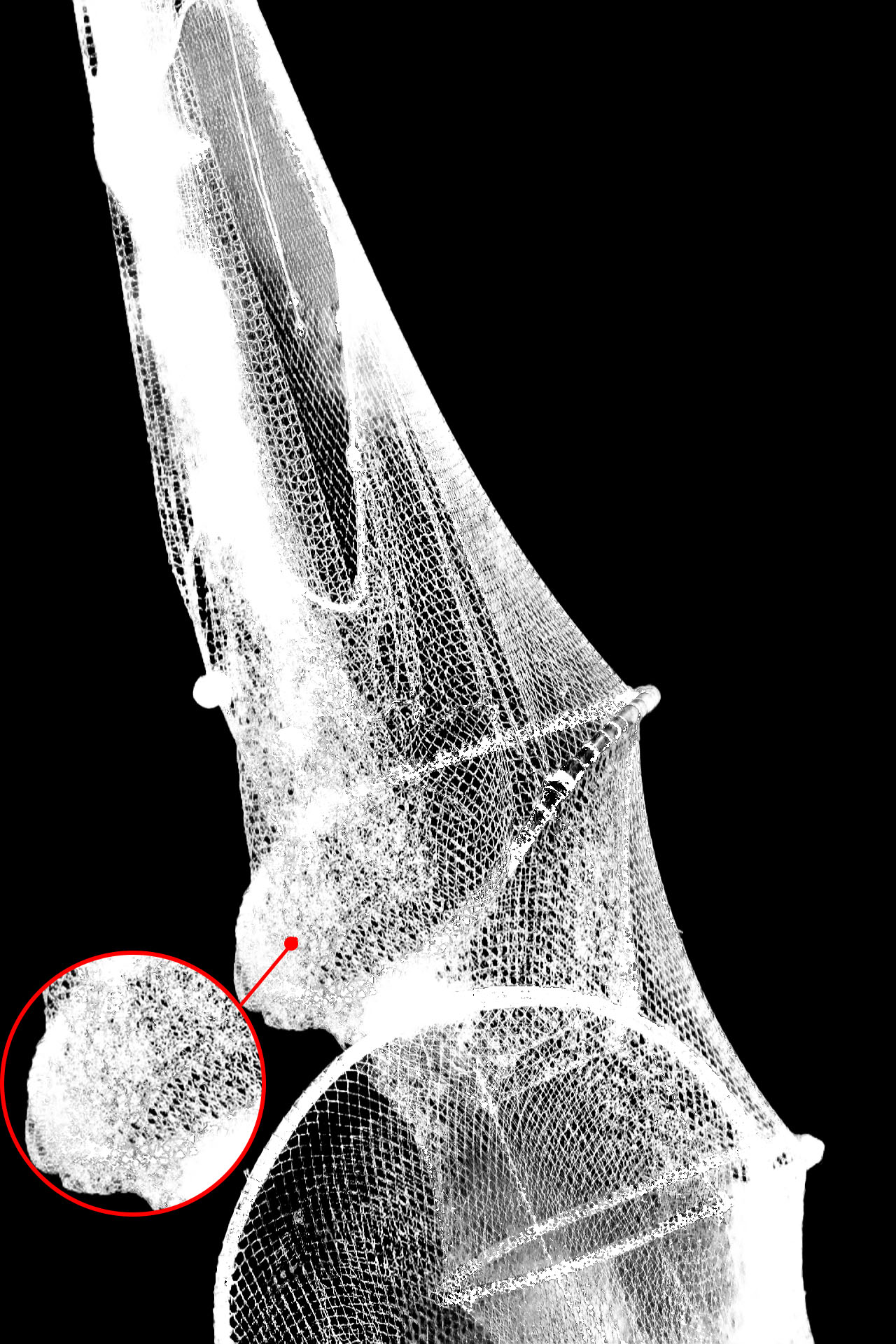} &	
			\includegraphics[scale=0.0763]{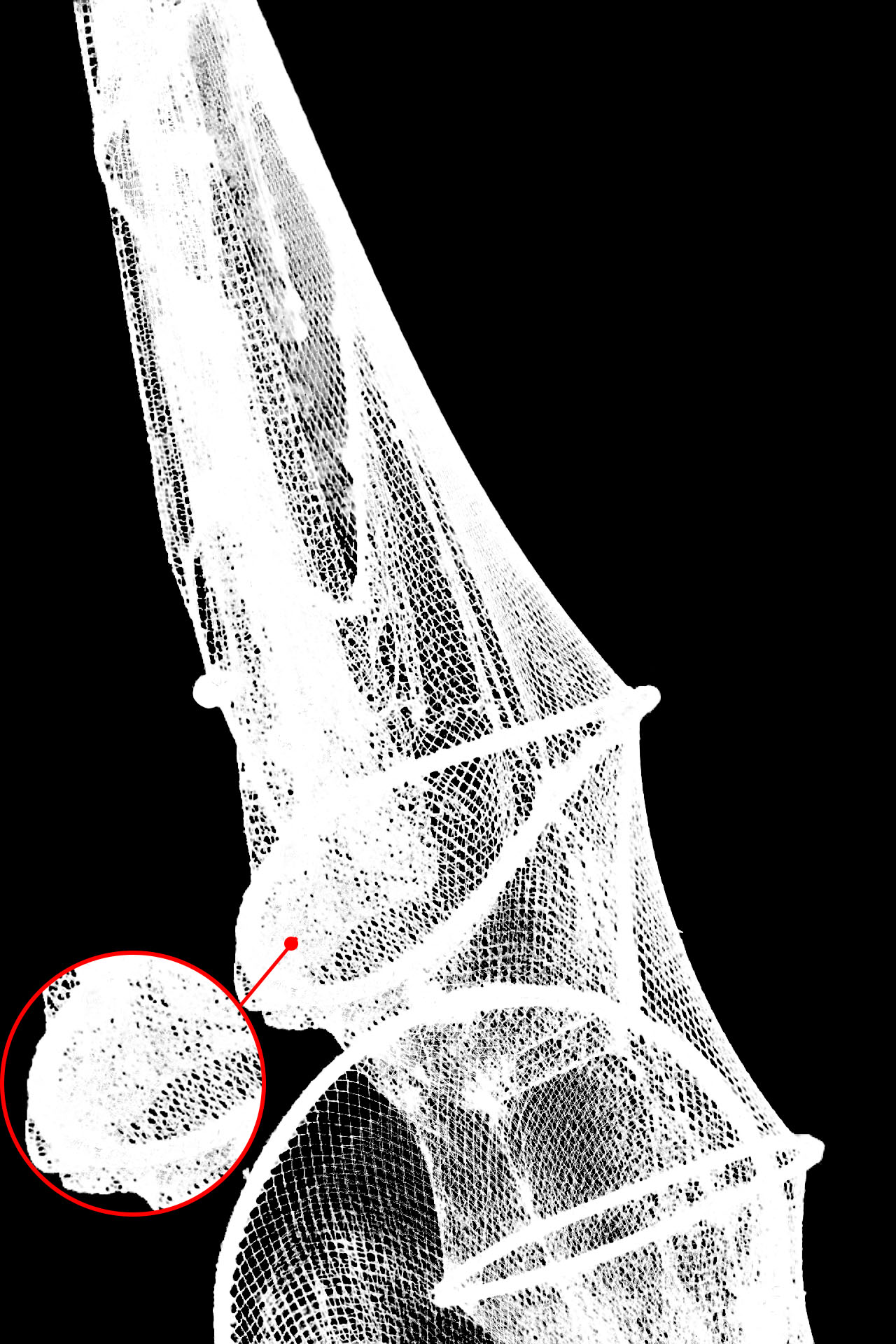} &
			\includegraphics[scale=0.0763]{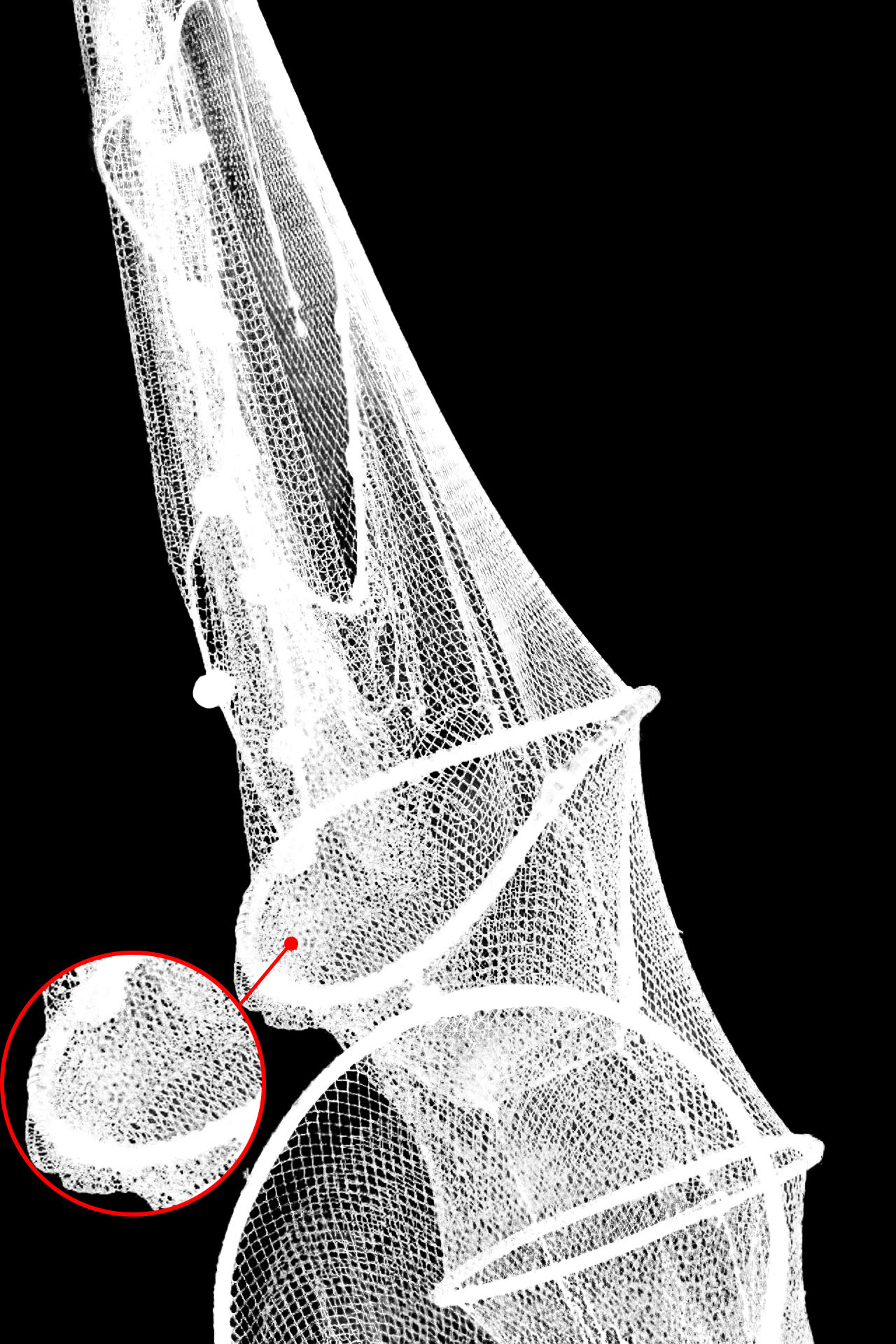} &
			\includegraphics[scale=0.0763]{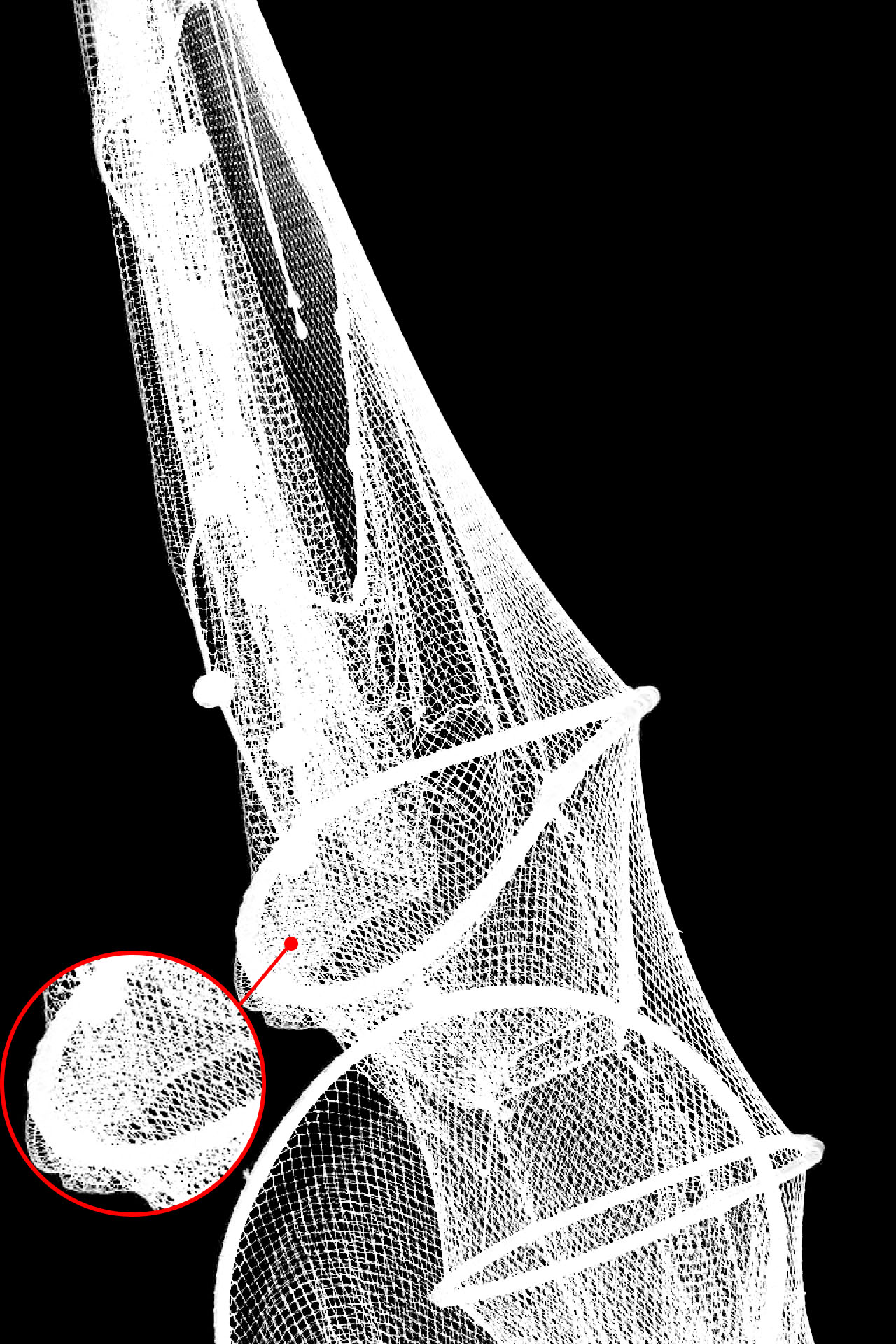} \\

			\includegraphics[scale=0.0424]{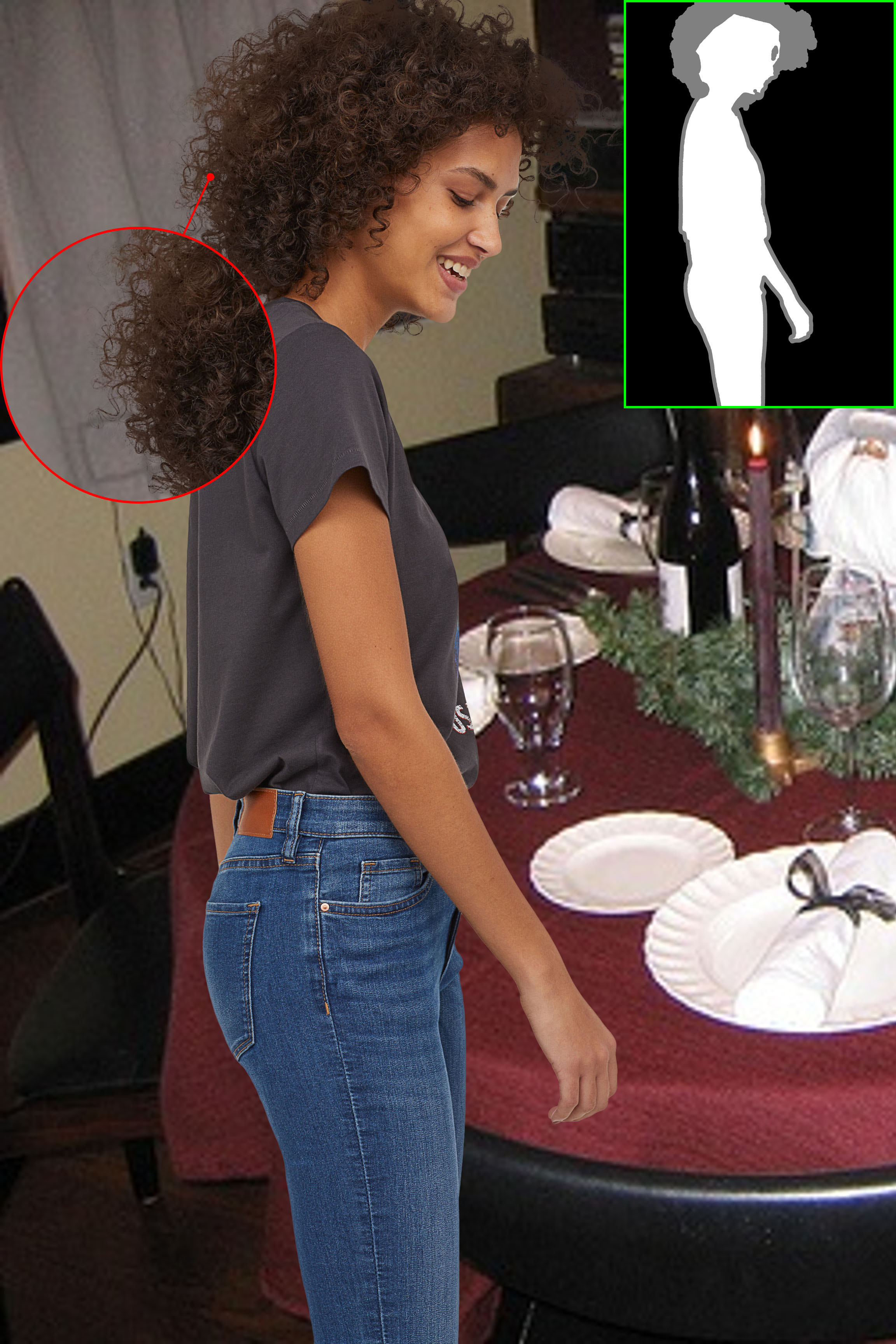} &
			\includegraphics[scale=0.0424]{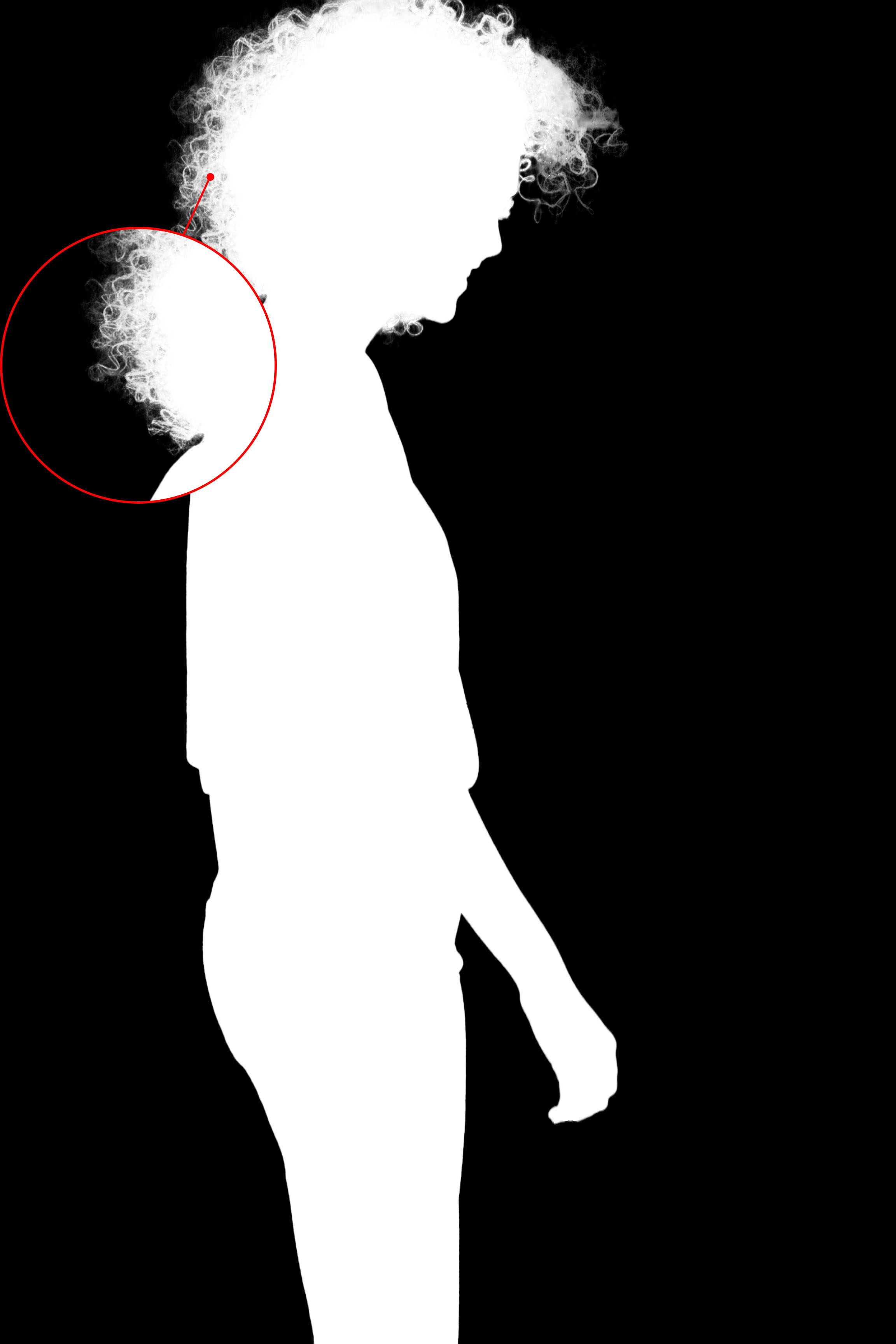} &	
			\includegraphics[scale=0.0424]{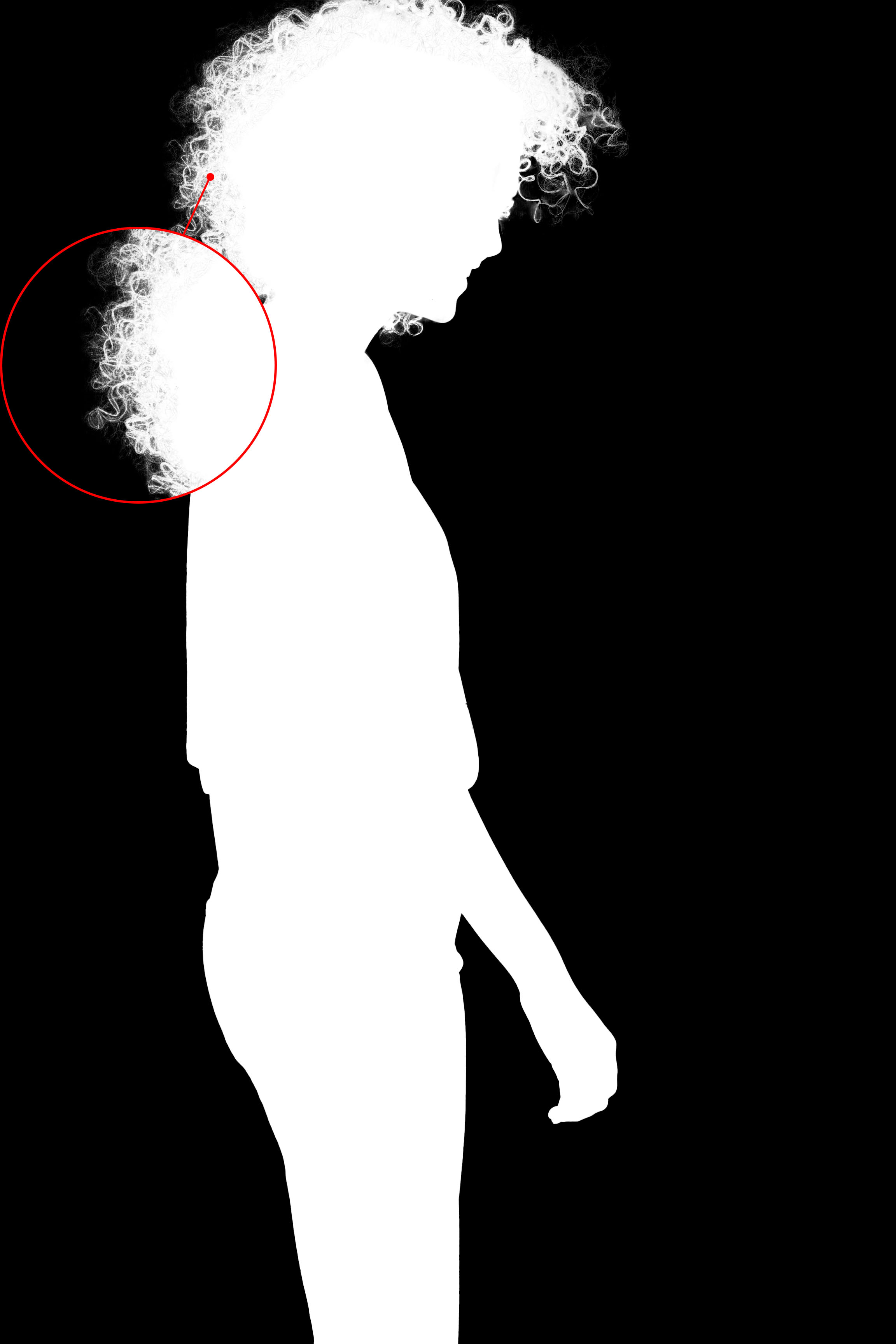} &
			\includegraphics[scale=0.0424]{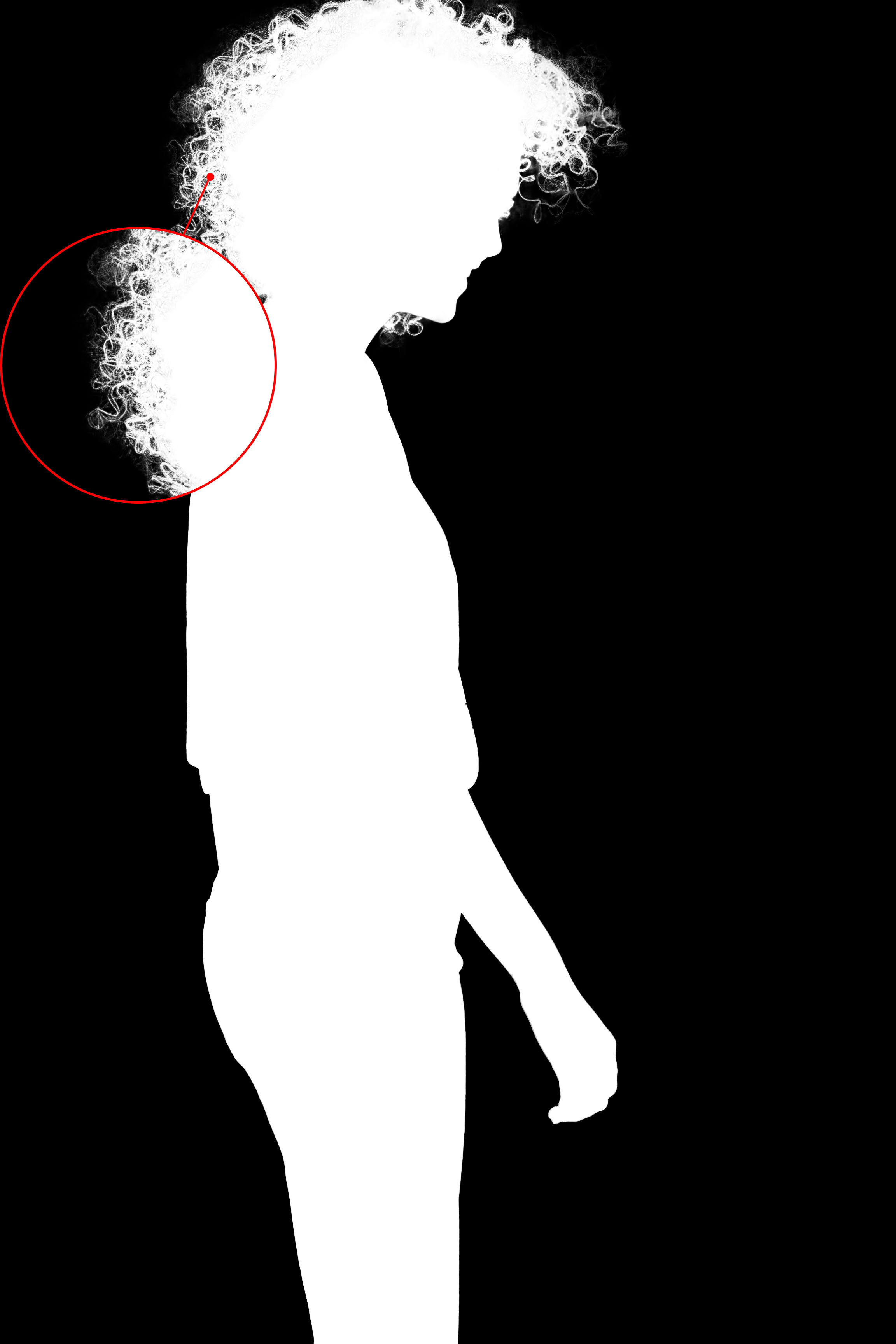} &
			\includegraphics[scale=0.0424]{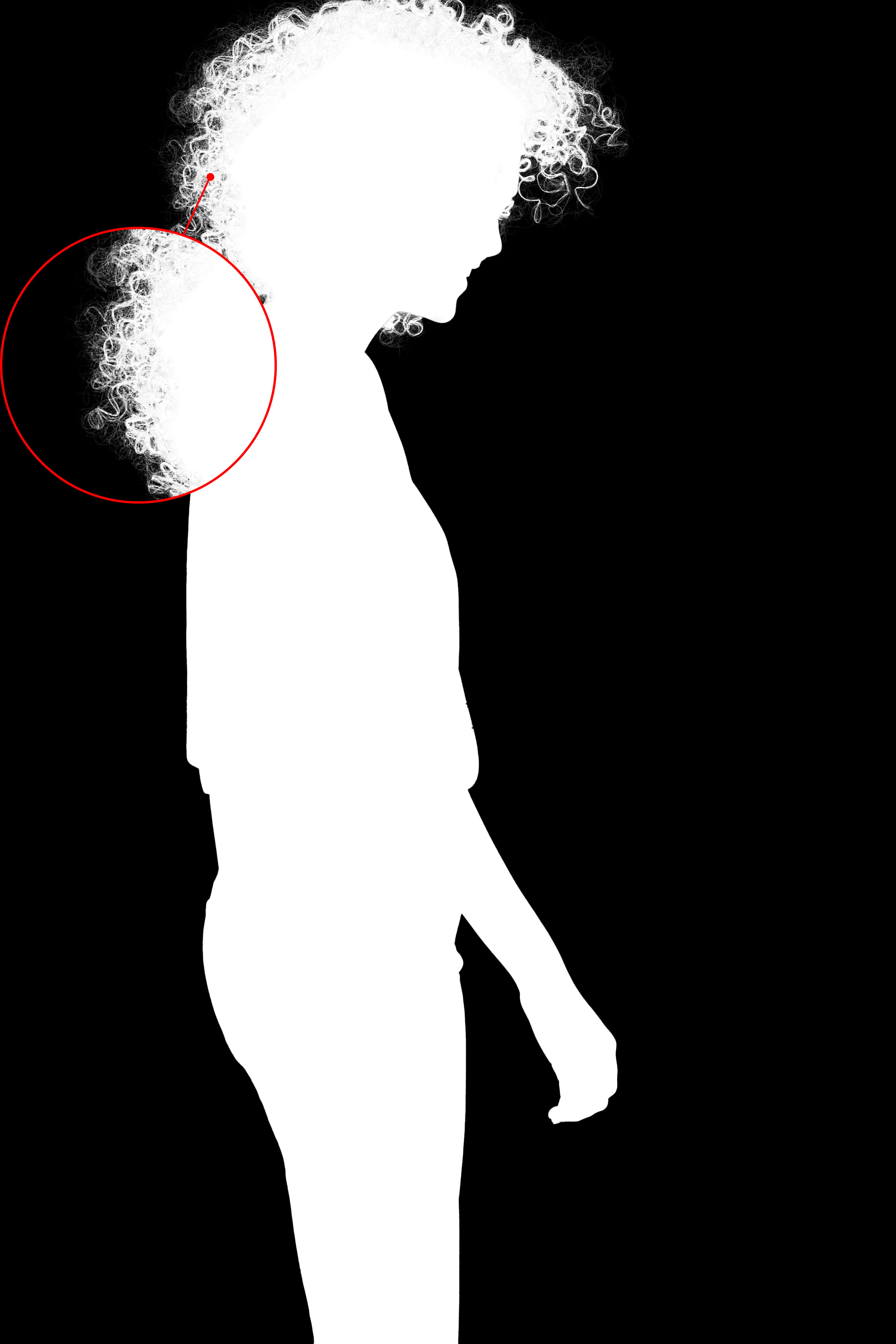} \\
			Inputs & DIM\cite{Xu2017Deep} & IndexNet\cite{lu2020index} & PIIAMatting (Ours) & GT  \\

	\end{tabular}}
	\vspace{-1mm}
	\caption{Qualitative comparisons on the Distinctions-646\cite{qiao2020attention} test set.}
	\vspace{-3mm}
	\label{fig:visual_646}
\end{figure*}

\begin{figure}[t]
\setlength{\tabcolsep}{1pt}\small{
	\begin{tabular}{ccc}
		\includegraphics[scale=0.0985]{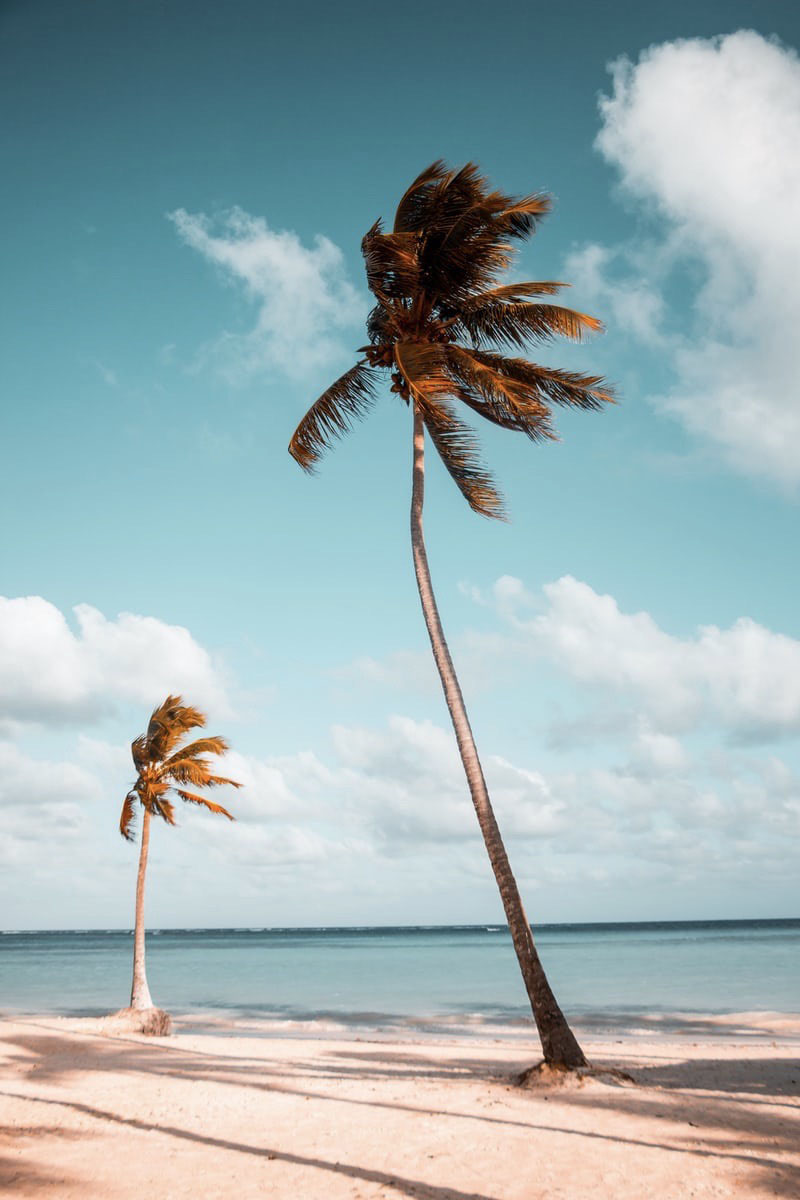} &
		\includegraphics[scale=0.0985]{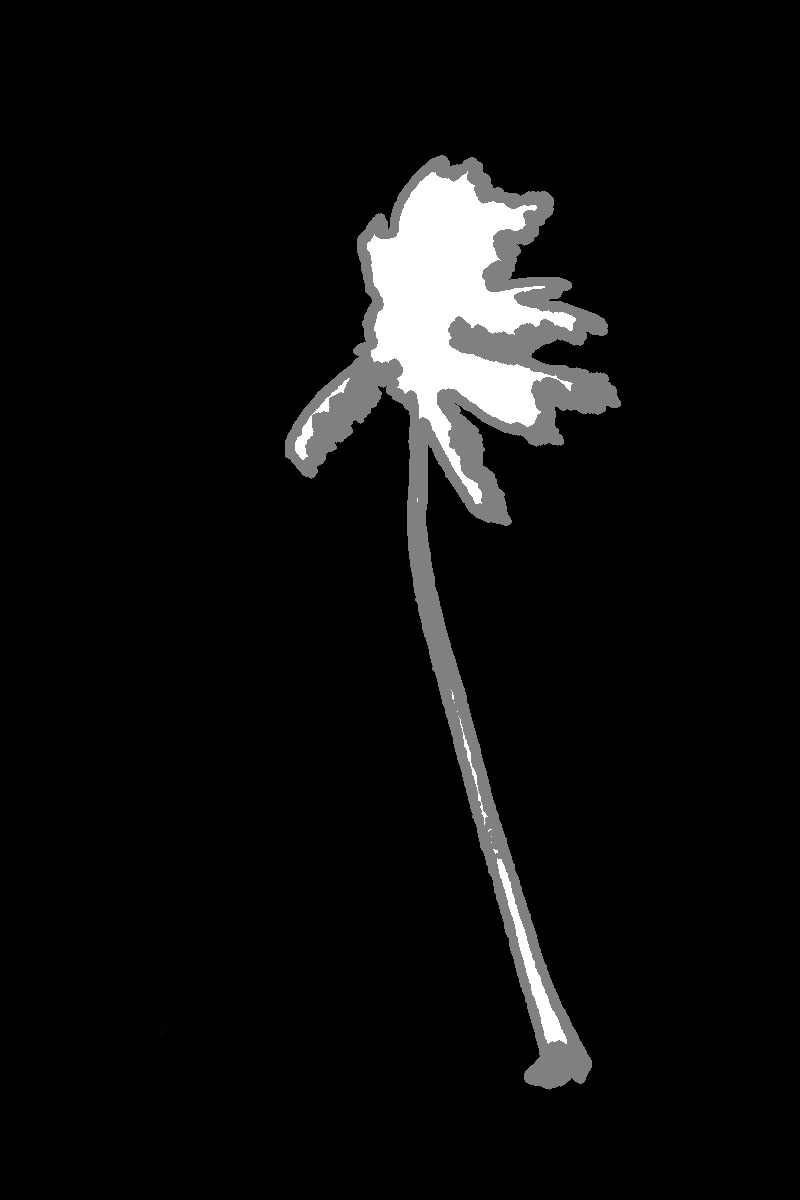}& 
		\includegraphics[scale=0.0985]{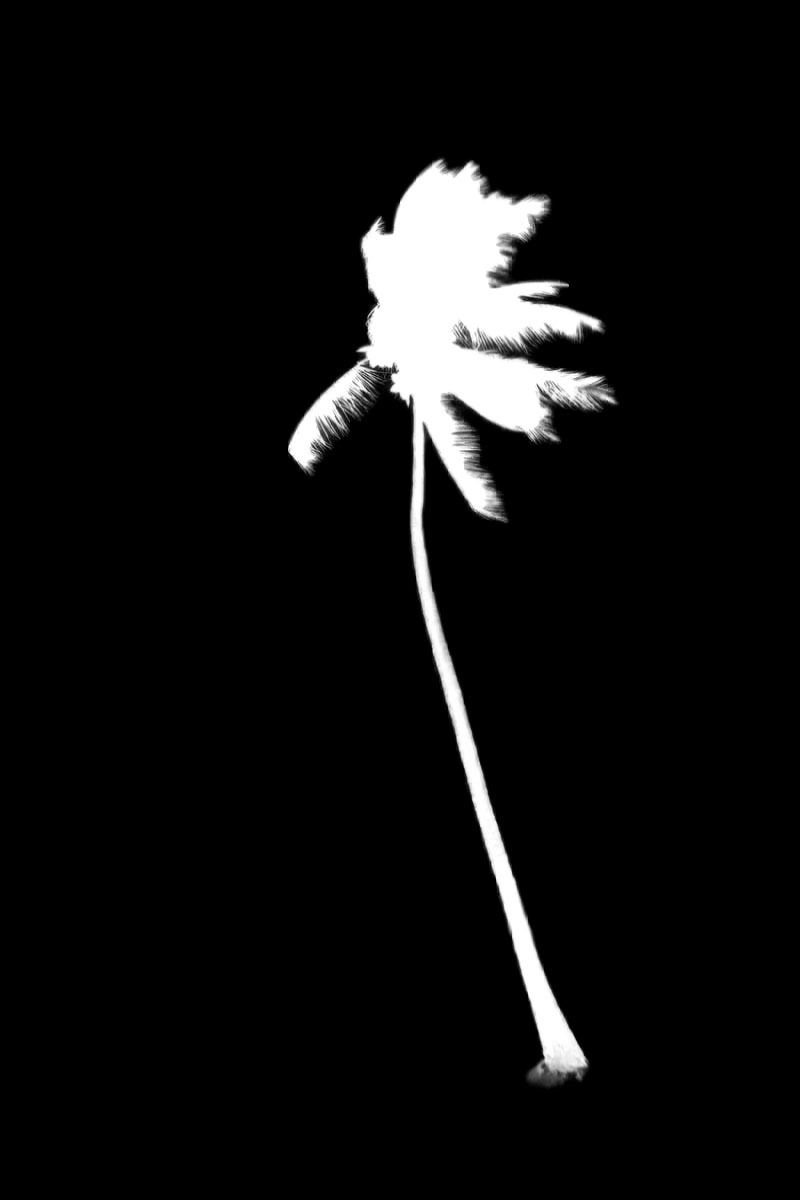}\\
		
		\includegraphics[scale=0.0617]{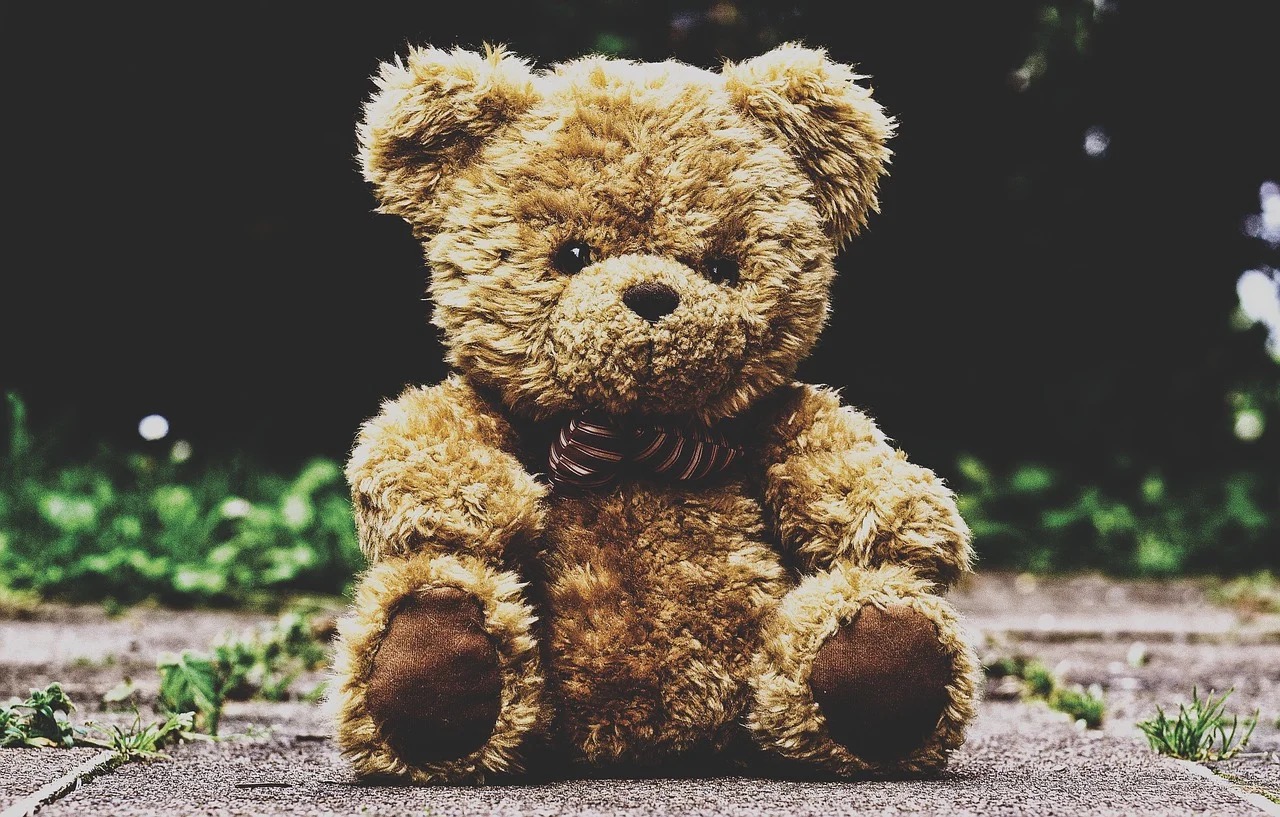} &
		\includegraphics[scale=0.0617]{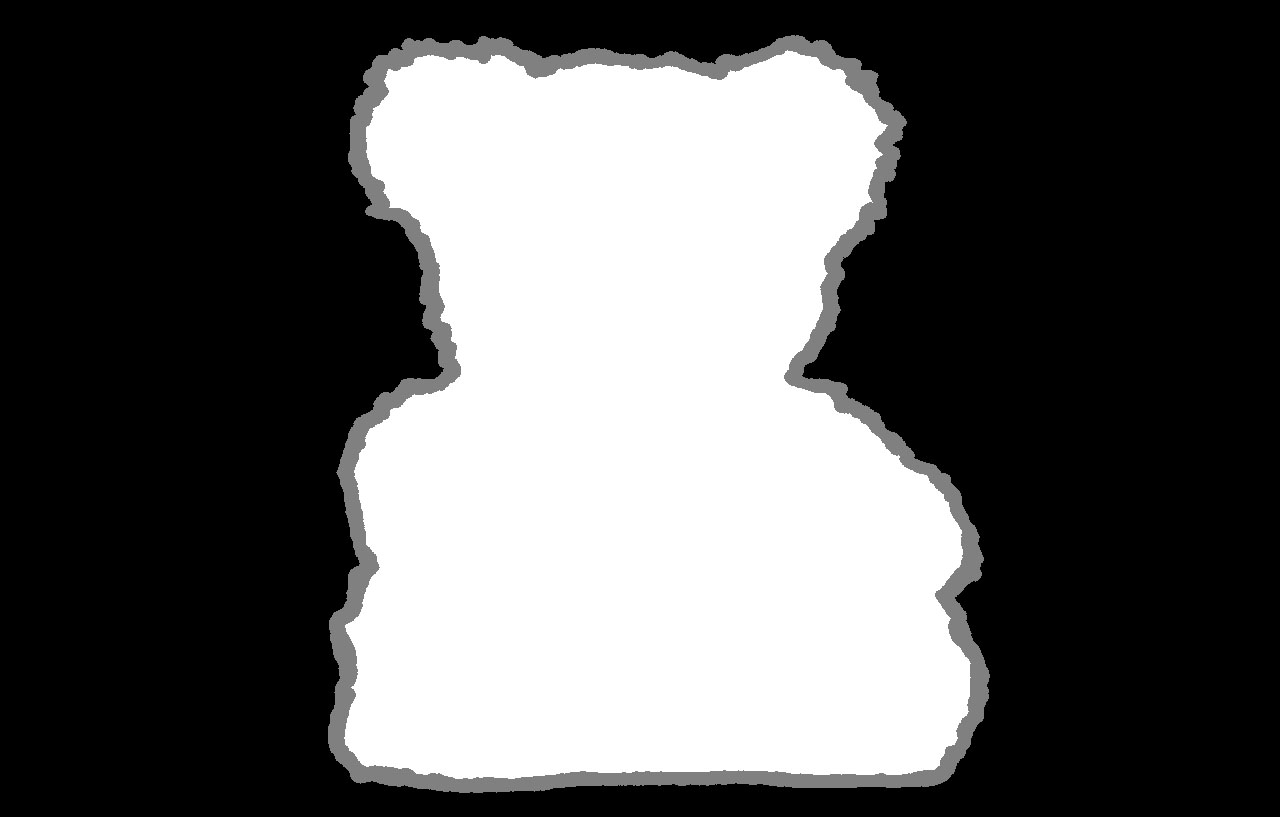} &
		\includegraphics[scale=0.0617]{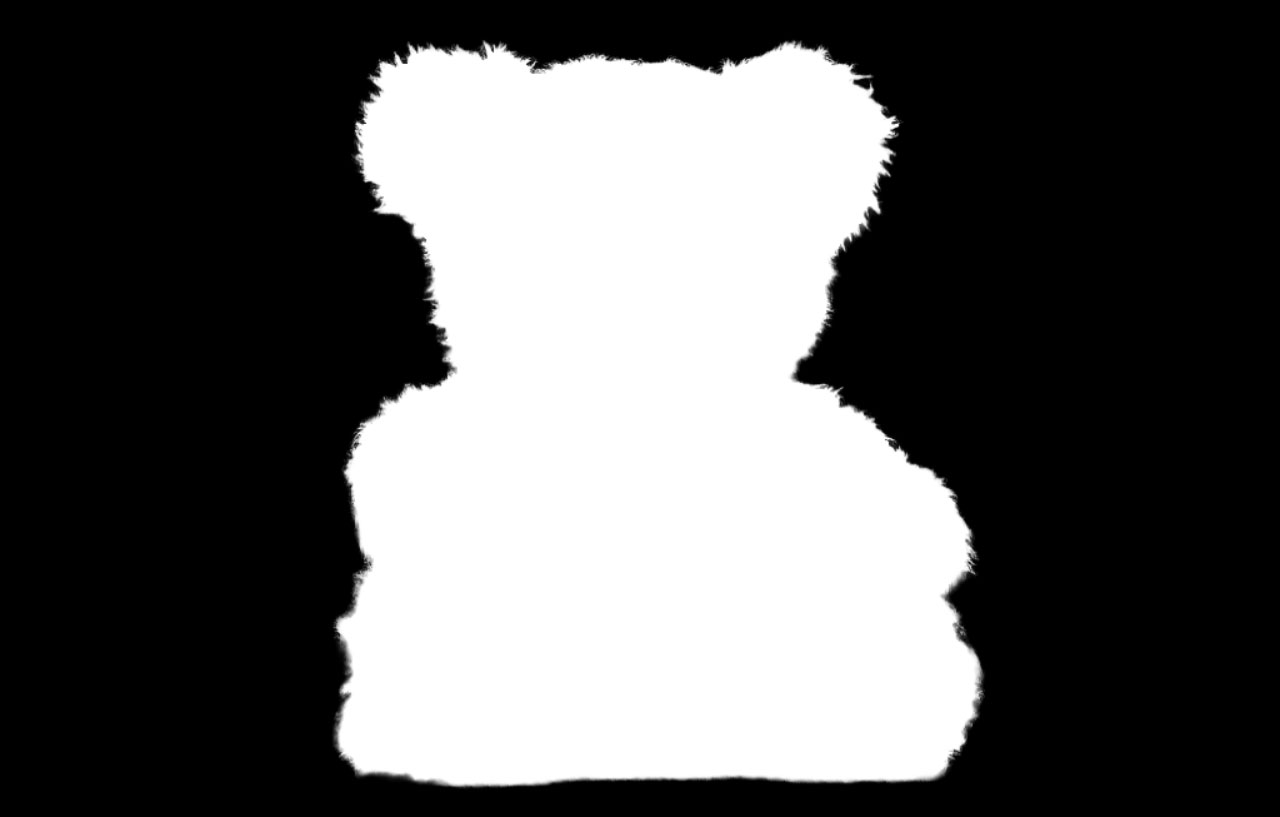} \\
		
		\includegraphics[scale=0.0616]{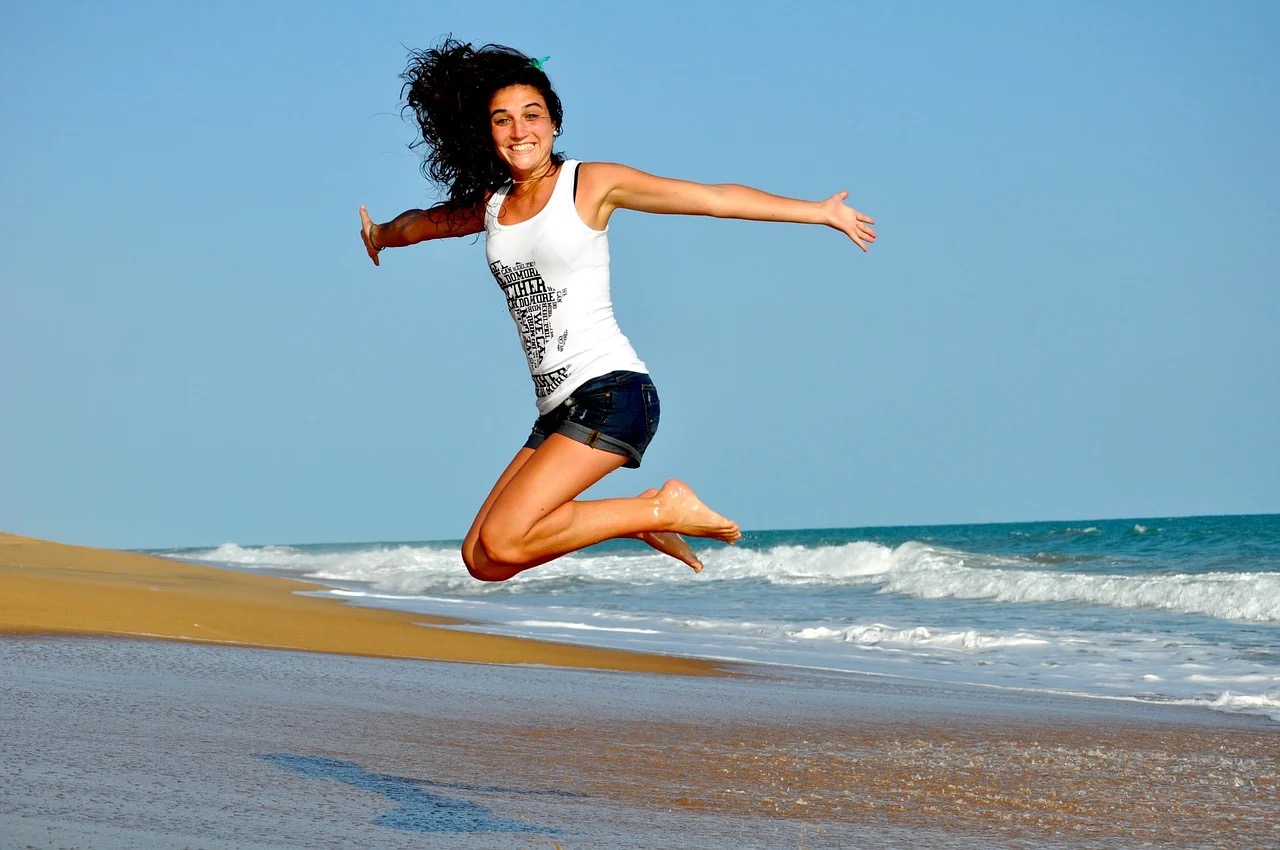} &
		\includegraphics[scale=0.0616]{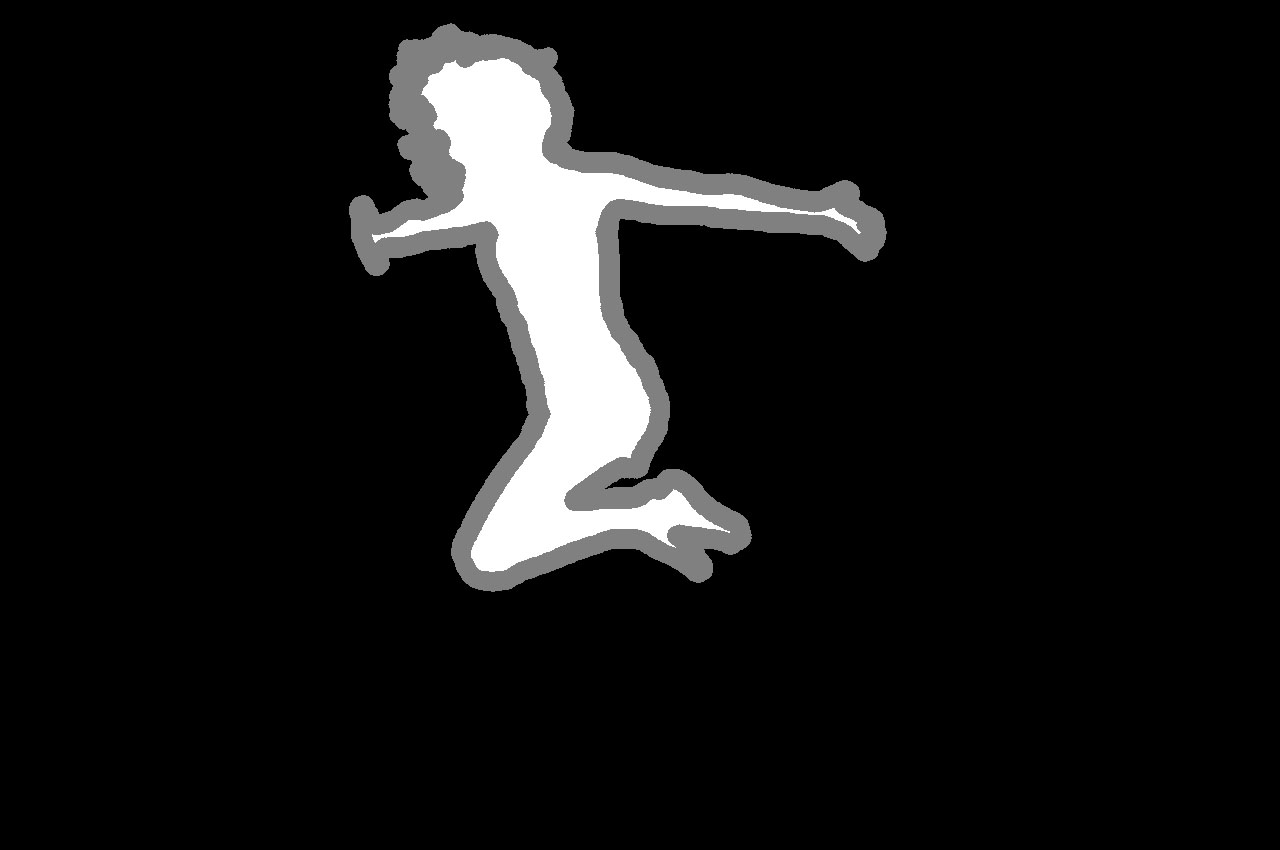} &
		\includegraphics[scale=0.0616]{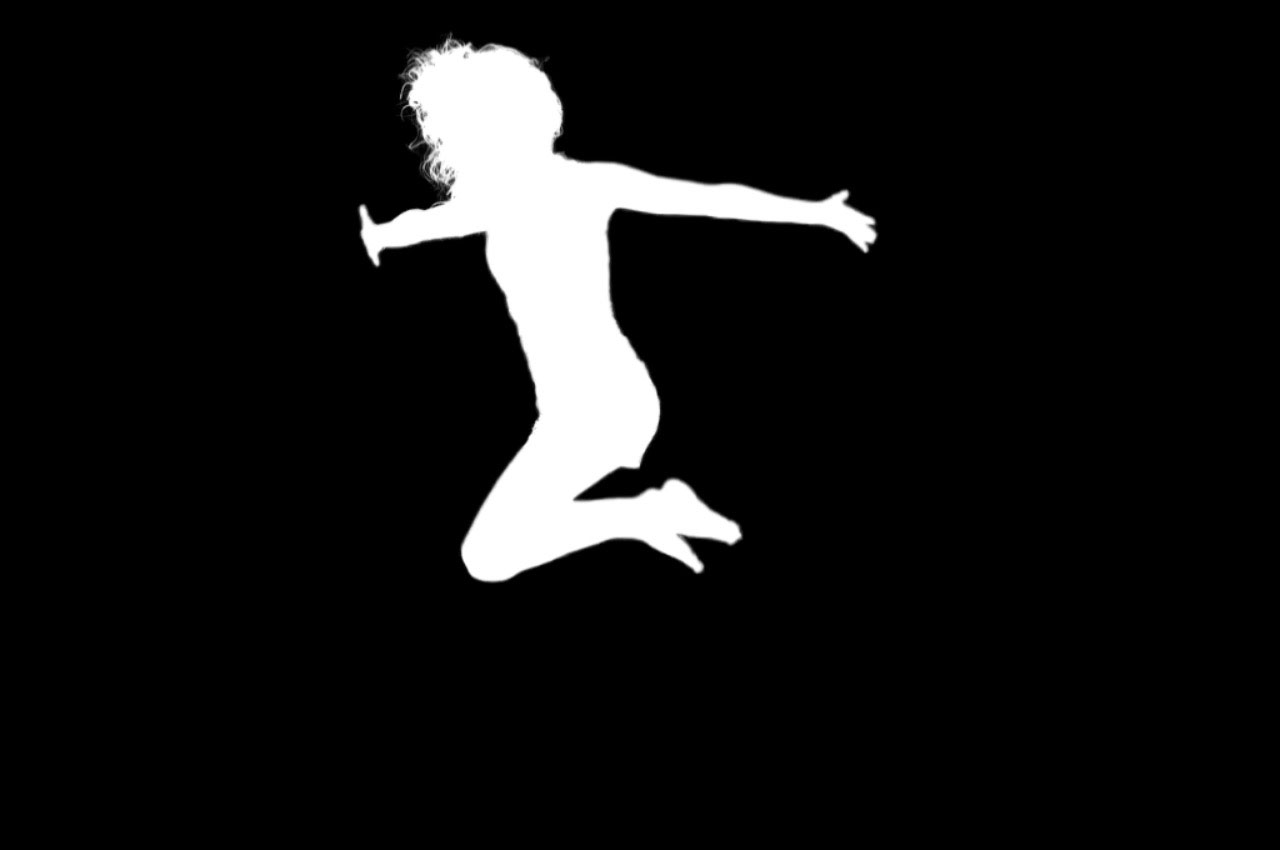} \\
		
		\includegraphics[scale=0.0615]{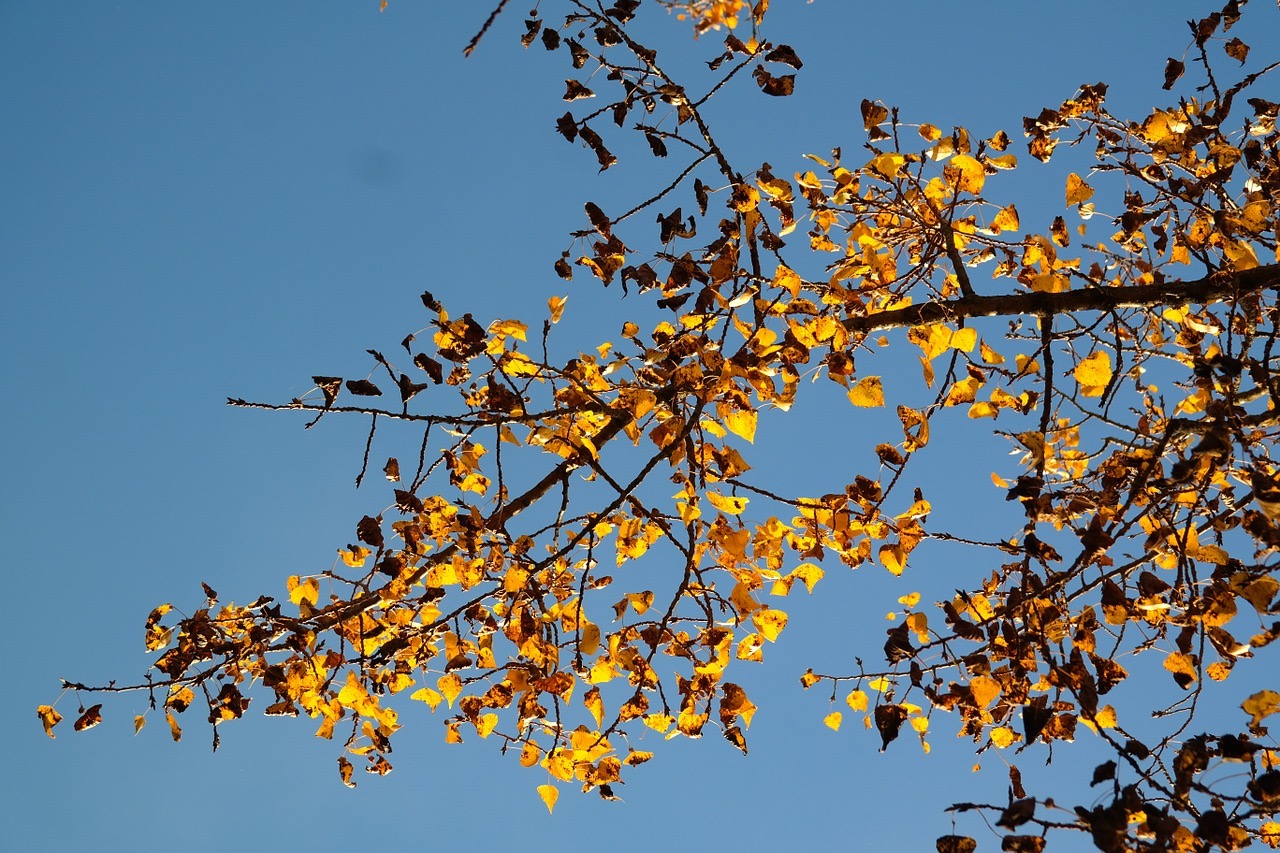} &
		\includegraphics[scale=0.0615]{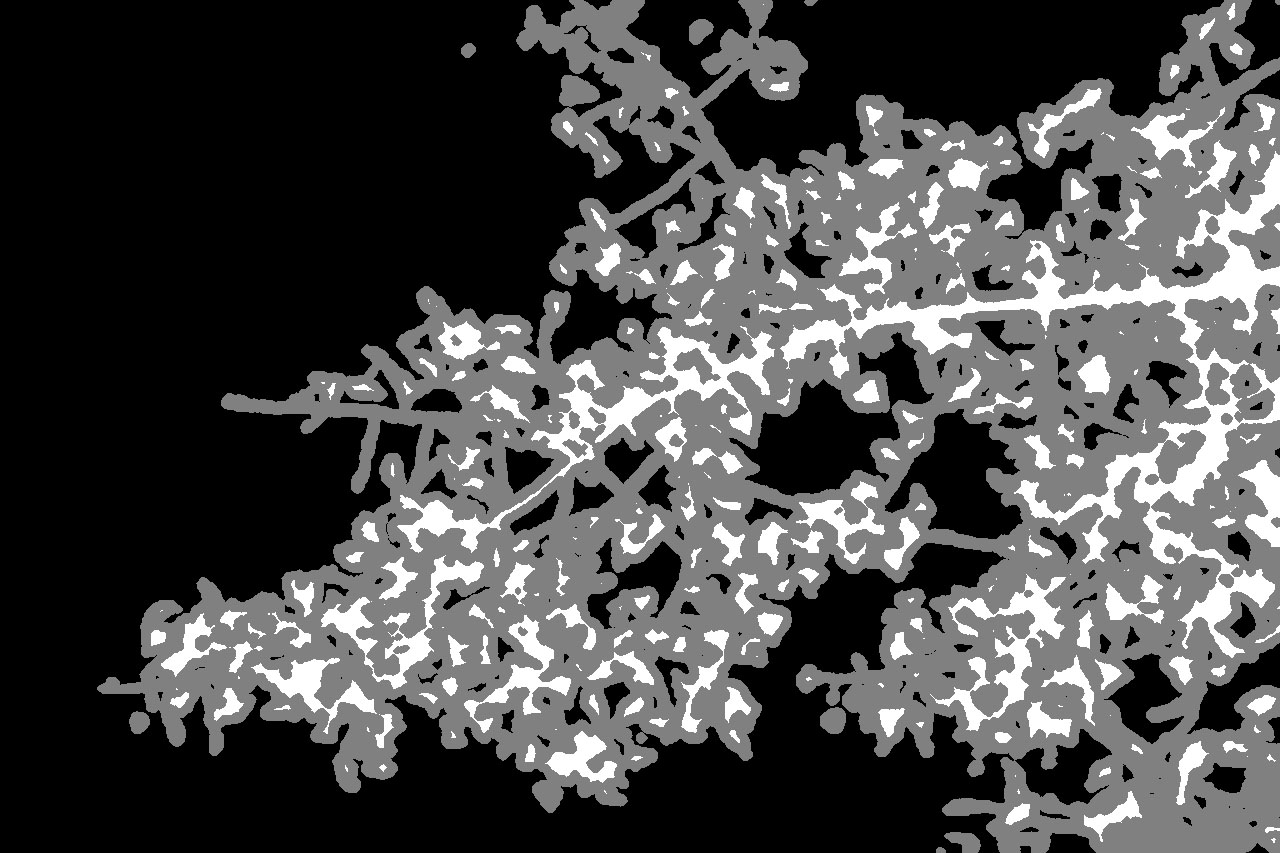} &
		\includegraphics[scale=0.0615]{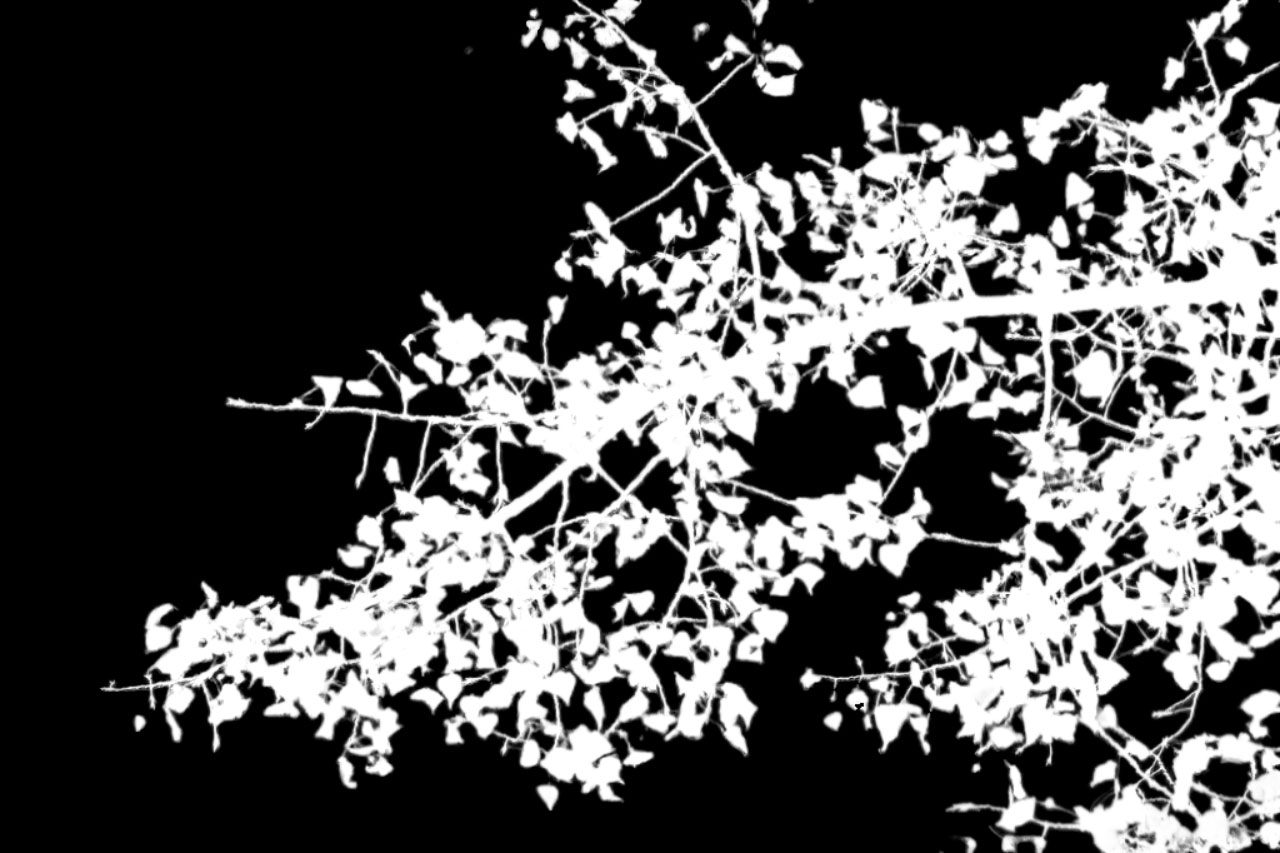} \\
		
		Input Image & Input Trimap & Alpha Matte \\
\end{tabular}}
\vspace{-2mm}
\caption{The results of our method on Real-World images.}
\vspace{-3mm}
\label{fig:visual_real}
\end{figure}

\subsection{Comparison to  Prior Work}
\label{ssec:comaprison_to_sota}
\subsubsection{Results on the Alphamatting.com dataset}
\label{sssec:alphamatting}
We evaluate our method on Alphamatting.com\cite{rhemann2009perceptually}, and the rank is shown in Table.~\ref{tab:performance_comparison}.  
There are some SOTA methods listed for comparison. We can see that our method (PIIAMatting) achieves  state-of-the-art performance, ranking in the ﬁrst place for the average performance on all three metrics at the time of this submission. 

Some  qualitative comparisons are presented in Figure.~\ref{fig:visual_benchamrk}. Generally, our method is more robust to unknown regions of different sizes and can retain more detailed information. 
Compared with  IndexNet\cite{lu2020index} and GCA \cite{li2020natural}, our method has better performance on the "Plastic bag", "Net" and "Troll" image cases, as shown in  Fig.~\ref{fig:visual_benchamrk}.

\subsubsection{Results on the Composition-1K dataset}
\label{sssec:adobe}
We compare our method with 16 state-of-the-art methods, including 6   traditional colour-based  methods:  Shared Matting\cite{gastal2010shared}, Learning-based\cite{zheng2009learning}, Global Matting\cite{he2011global}, Information-Flow \cite{aksoy2017designing}, ClosedForm\cite{levin2007closed}, KNN Matting\cite{chen2013knn},
and 10 latest deep learning-based approaches:
DCNN\cite{cho2018deep}, DIM\cite{Xu2017Deep}, AlphaGAN\cite{lutz2018alphagan}, SampleNet\cite{tang2019learning}, IndexNet\cite{lu2020index}, AdaMatting\cite{cai2019disentangled}, Context-Aware\cite{hou2019context} and GCA\cite{li2020natural}, LFM\cite{zhang2019late}, HAttMatting\cite{qiao2020attention}.
 For fair comparisons, we use the released codes and their default parameters to implement these methods. In terms of methods without released source code, we use their published results for comparisons.
\begin{table}[t]
\caption{Quantitative comparisons with other latest algorithms using the Composition-1k\cite{Xu2017Deep} test set. The gradient loss is scaled by 1000 and “-“means  the value that was not given in the original paper. Ours-Big and Ours-Data represent extra models that our PIIAMatting furnished with the ResNet-101\cite{he2016deep} and data augmentation\cite{tang2019learning}, respectively.}
	\label{tab:quantitative_adobe}
	\small
	\centering
	\setlength{\tabcolsep}{1.8mm}{
		\begin{tabular}{l|ccccc}
			\hline
			Methods & SAD$\downarrow$ & MSE$\downarrow$  & Grad$\downarrow$  & 
			Conn$\downarrow$  \\
			\hline
			Shared Matting~\cite{gastal2010shared} & 128.9 & 0.091 & 126.5 & 
			135.3 \\
			Learning Based~\cite{zheng2009learning} & 113.9 & 0.048 & 91.6 & 122.2 \\
			Global Matting~\cite{he2011global} & 133.6 & 0.068 & 97.6 & 133.3 \\
			Information-Flow~\cite{aksoy2017designing} & 75.4 & 0.066 & 63.0 & 
			- \\
			
			DCNN~\cite{cho2018deep} & 161.4 & 0.087 & 115.1 & 161.9 \\
			KNN Matting~\cite{chen2013knn} & 175.4 & 0.103 & 124.1 & 176.4 \\
			ClosedForm~\cite{levin2007closed} & 168.1 & 0.091 & 126.9 & 167.9 \\
			AlphaGAN~\cite{lutz2018alphagan} & - & 0.031 & - & - \\
			DIM~\cite{Xu2017Deep} & 54.6 & 0.017 & 36.7 & 55.3 \\
		
			LFM \cite{zhang2019late} & 49.0 & 0.020 & 34.3 & 50.6\\
			HAttMatting \cite{qiao2020attention} & 48.8 & 0.016 & 25.3 & 48.6\\
			SampleNet\cite{tang2019learning} & 40.4 & 0.010 & - & -\\
			IndexNet~\cite{lu2020index} & 45.8 & 0.013 & 25.9 & 43.3 \\
			AdaMatting~\cite{cai2019disentangled} & 41.7 & 0.010 & 16.8 & - \\

			Context-Aware~\cite{hou2019context} & 35.8 & \textbf{0.008} & 17.3 & 33.2\\
			GCA\cite{li2020natural}&35.3 & 0.009 & 16.9 & 32.6\\
			\hline \hline
			\rowcolor{mygray}
			PIIAMatting (\textbf{Ours}) & 36.4 & 0.009 & 16.9 & \textbf{31.5}\\
			PIIAMatting (Ours-Big) & 35.8 & 0.009 & \textbf{16.2} & 34.7\\	
			PIIAMatting (Ours-Data) & \textbf{35.2} & 0.009 & 16.9 & 33.1\\	
			\hline
		
	\end{tabular}}
	
\end{table}

Quantitatively, as shown in Tab.~\ref{tab:quantitative_adobe}, our approach achieves the SOTA results  with respect to four metrics, compared with other counterparts. It demonstrates the superior performance of the proposed PIIAMatting. 
Specifically, Our model without using additional data augmentation (\textbf{Ours}) has surpassed the DIM, LFM, SampleNet, IndexNet, AdaMatting. 
Although the result of our model (\textbf{Ours}) is slightly inferior to the Context-Aware and GCA, the former utilizes the ResNet-101\cite{he2016deep} as its two-branch backbone that learns stronger representations, and the latter resorts to the data augmentation for the improvement. 
To enable a fair competition, we also equipped our model with data augmentation and ResNet-101\cite{he2016deep} denoted as \textbf{Ours-Data}, \textbf{Ours-Big} respectively. As expected, both \textbf{Ours-Data} and \textbf{Ours-Big} outperform the GCA and Context-Aware.

Qualitatively, Fig.~\ref{fig:visual_adobe} shows a visual comparison of our method against  six top-performing approaches. For better visualization, we highlight the salient differences in the red circle on each image.
It is worth noting that even the  IndexNet \cite{lu2020index} and GCA \cite{li2020natural} can get smooth and performing results, but the effect is not ideal for the highly translucent pixels, and some artefacts can still be observed on their alpha mattes.
On the contrary, the result of our model is more delicate on the undetermined  domains, mainly due to the fact that our method takes the pixel domains disproportion and information alignment into account during the training.  The DGM mechanism can alleviate the disproportion of discrepant pixel domains, while the IA strategy can lighten the information discrepancy. Both of them contribute together to achieve such good results.

\subsubsection{Results on the Dinstinction-646 dataset}
\label{sssec:distinction}
For the Distinctions-646 dataset, we take the same standards as above to make the comparisons, and here we mainly compare the following most representative methods:
Learning Based\cite{zheng2009learning}, Shared Matting \cite{gastal2010shared}, Global Matting\cite{he2011global}, DIM\cite{Xu2017Deep},  IndexNet\cite{lu2020index} and HAttMatting\cite{qiao2020attention}.
We first visually compare the results of the proposed method to the state-of-the-art image matting methods. Fig.~\ref{fig:visual_646} shows the results of DIM\cite{Xu2017Deep} and IndexNet\cite{lu2020index} on the Distinctions-646 test set. We can see that DIM\cite{Xu2017Deep} and IndexNet\cite{lu2020index} are incapable of handling the situation very well when the undetermined domain accounts for most of the unknown regions (see the first and second row). The DIM\cite{Xu2017Deep} and IndexNet\cite{lu2020index} treat the distinct domains equally and thus neglect the opacity variation. While the IndexNet\cite{lu2020index} can somewhat enhance the boundary details, it did not fully consider the relationship between semantic context and texture information on the adjacent layer-wise features. In contrast, our results show that the proposed method can notably enhance the adaptability of the model to opacity variation while retaining details that are easy to lose.
We have also quantitatively compared our method to these methods. As shown in Tab.~\ref{tab:quantitative_646}, the PIIAMatting consistently outperforms other methods on the Distinctions-646 test set with respect to four metrics, demonstrating the superior performance of our method on image matting. 
\begin{table}[t]
\caption{Quantitative comparisons on the Distinctions-646\cite{qiao2020attention} dataset.}
	\label{tab:quantitative_646}
	\small
	\centering
	\setlength{\tabcolsep}{1.8mm}{
		\begin{tabular}{l|ccccc}
			\hline
			Methods & SAD$\downarrow$ & MSE$\downarrow$  & Grad$\downarrow$  & 
			Conn$\downarrow$  \\
			\hline
			Learning Based~\cite{zheng2009learning} & 93.3 & 0.096 & 129.2 & 94.9 \\
			Shared Matting~\cite{gastal2010shared} & 71.5 & 0.064 & 83.2 & 70.9 \\
			Global Matting~\cite{he2011global} & 63.6 & 0.052 & 76.2 & 64.5 \\
			DIM~\cite{Xu2017Deep} & 44.2 & 0.029 & 39.1 & 44.6 \\
			HAttMatting~\cite{qiao2020attention} & 42.6 & 0.027 & 47.0 & 42.9 \\
			IndexNet~\cite{lu2020index} & 35.5 & 0.019 & 25.1 & 35.6 \\
			\hline
			\rowcolor{mygray}
			PIIAMatting (Ours) & 27.7 & 0.014 & 18.2 & 20.6\\
			\hline
	\end{tabular}}
	
\end{table}

\subsection{Internal Analysis}
\label{ssec:ablation}
In this section, with the Composition-1K dataset\cite{Xu2017Deep} as the training set, we detailedly explore the validity and generalization of different components in the proposed PIIAMatting. Analyses are conducted on the following three aspects: (i) Dynamic Gaussian Modulation mechanism (DGM), (ii) Information Alignment strategy (IM)  and (iii) Multi-Scale Refinement module (MSR).

\begin{table}
	\begin{center}
		\caption{Analytical experiments of our model with different loss functions on the Composition-1K dataset~\cite{Xu2017Deep}. Comp$+$ Alpha means the compositional loss$+$alpha loss. L1$+$L2 refers to the  L2 loss is calculated when alpha is equal to 0 and 1, and L1 loss is calculated on the rest. Gaussian L1 represents the L1 loss combined with the static  Gaussian mechanism.  Gaussian L1-D means that the Dynamic Gaussian Modulation mechanism is attached to Gaussian-L1. All experiments here abandoned the use of the MSR.}
		\label{tab:gaussian_self}
		\begin{tabular}{l|ll}
			\hline
			Distinct Loss Function  & MSE$\downarrow$ & SAD$\downarrow$ \\
			\hline
			Comp $+$ Alpha Loss~\cite{Xu2017Deep} & 0.0118 & 43.2 \\
			L1 $+$ L2 Loss~\cite{zhang2019late} & 0.0107 & 41.5 \\
			Gaussian L1~(Ours) & 0.0105 & 40.6 \\
			Gaussian L1-D~(Ours) & 0.0102 & 40.3 \\
			\hline
		\end{tabular}
	\end{center}
\end{table}

\textbf{Analysis on DGM.~}
In this subsection, we investigate the role of DGM, which is also our core component. 

1). \textit{The validity of DGM}: In the following, we illustrate this efficacy by comparing different loss functions on our model. For a fair comparison and to prevent the disturbance from other components, we removed the MSR and only utilized the rest of  components. 
 
 As shown in Tab.~\ref{tab:gaussian_self},  Alpha$+$Comp refers to the loss function applied in DIM~\cite{Xu2017Deep}, L1$+$L2 indicates the loss function in LFM~\cite{zhang2019late}, the Gaussian L1 is the static Gaussian  Mechanism that $\mu = 0.5$ and $\sigma = 0.25$ in this submission. Gaussian L1-D denotes that the Dynamic Modulation mechanism is attached to Gaussian-L1.
Our model combined with Gaussian L1 is capable of getting obvious improvements compared with the Comp$+$Alpha, especially the decrement of SAD from 43.2 to 40.6, which validates the ability of DGM to adapt to the opacity variation.

With the incorporation of the  Gaussian-L1 dynamic mechanism, the degree of exploration of our model in the undetermined domain decreased progressively with the training progresses. We can observe that our model can obtain further improvement (\eg~the SAD decreased from 40.6 to 40.3).
Meanwhile, the SAD of our PIIAMatting decreased by 1.2 compared to the L1 $+$ L2 that used in LFM. This is mainly due to the L1 $+$ L2 can only be regarded as a pre-defined static weighting loss function based on the different partition. Hence it can not handle properly with the variation of opacity at distinct domains.

2). \textit{The generalization of DGM}: In what follows, we embed DGM in other SOTA methods (\eg~DIM\cite{Xu2017Deep}, IndexNet\cite{lu2020index}, and GCA\cite{li2020natural}) to prove its effectiveness and generalization.

\begin{table}
  \begin{center}
  \caption{Analysis of the DGM on the Composition-1K\cite{Xu2017Deep} dataset. Ori loss represents the primal loss function used in the method. DGM indicates our Dynamic Gaussian Modulation mechanism, and 'wo' means disable the data augmentation. }
  \label{tab:gaussian_other}
  \begin{tabular}{lll | ll}
  \hline\noalign{\smallskip}
  Baseline Model & Ori Loss  & DGM & MSE & SAD \\
  \noalign{\smallskip}
  \hline
  \noalign{\smallskip}
  DIM \cite{Xu2017Deep}& \makecell[c]{\checkmark} &&\makecell[c]{0.0171}&\makecell[c]{54.6}\\
  DIM \cite{Xu2017Deep}& \makecell[c]{\checkmark} &\makecell[c]{\checkmark} &\makecell[c]{0.0152}&\makecell[c]{52.4} \\
  IndexNet \cite{lu2020index}& \makecell[c]{\checkmark} &&\makecell[c]{0.0132}&\makecell[c]{45.8}\\
  IndexNet\cite{lu2020index}&\makecell[c]{\checkmark}  &\makecell[c]{\checkmark} &\makecell[c]{0.0128}&\makecell[c]{45.2} \\
  GCA (wo) \cite{li2020natural}& \makecell[c]{\checkmark} &&\makecell[c]{0.0126}&\makecell[c]{47.1}\\
  GCA (wo)\cite{li2020natural}& \makecell[c]{\checkmark} &\makecell[c]{\checkmark} &\makecell[c]{0.0117}&\makecell[c]{44.7}\\
  \hline
  \end{tabular}
  \end{center}
  \end{table}
  
As shown in Tab.~\ref{tab:gaussian_other}, after combination with the DGM, all three methods achieved a certain degree of improvement.  For DIM, the metric of MSE and SAD decreased by 0.0019 and 2.2, respectively. While for GCA, MSE and SAD decreased by 0.0009 and 2.3.
Due to IndexNet's low-capacity nature of its model demonstrated by the authors, the two evaluation metrics were just reduced by 0.0004 and 0.6 on IndexNet  $+$ DGM, respectively.
In a word, we can see that the DGM can greatly increase the adaptability of the model by modulating the disproportionality between different domains, which also confirms the effectiveness.

\textbf{Analysis on IA.~}To demonstrate the proposed the Information Alignment strategy efficiently retained the valuable details, we validate it from the following two sides: 1). Self-comparison with and without IA. 2). The generality of IA to other methods. We disable the MSR and employ only the Composition$+$Alpha loss on our model for a fair comparison.
Since the IMM and IAM are the concrete implementations of IA, we remove the IMM and IAM and take the decoder in U-net\cite{ronneberger2015u} which is denoted Ours-wo/IA. 
As shown in Tab1,  the four metrics of Ours-w/IA  decreased a lot compared to Ours-wo/IA after removing the IA, particularly the SAD dropped from 53.8 to 43.2 and the MSE decreased from 0.015 to 0.012. 
Such results confirm that our Information Alignment strategy is beneficial for image matting due to the ability to match and aggregate useful information during the sampling process. Furthermore, we  apply our IA to DIM\cite{Xu2017Deep} to prove its generality. Specifically, we switch the original decoder in DIM with our IA and take the same training parameters and policy. We can see that the result of DIM-IA is far superior to the result of the fundamental DIM. It is worth noting that the improvements of 0.002 and 14.7 with respect to the metrics of MSE and SAD, which further demonstrate that the ability of IA to excavate valuable information is effective and efficient.

\begin{table}[t]
\caption{Analysis of the IA and MSR on the Composition-1K\cite{Xu2017Deep} dataset. 'w' and 'wo' mean that our model combines and disables the corresponding components, respectively. The Refine refers to the refinement module in DIM\cite{Xu2017Deep}.}
	\label{tab:ablation_networks}
	\small
	\centering
	\setlength{\tabcolsep}{1.8mm}{
		\begin{tabular}{l|ccccc}
			\hline
			Methods & SAD$\downarrow$ & MSE$\downarrow$  & Grad$\downarrow$  & 
			Conn$\downarrow$  \\
			\hline
			DIM~\cite{Xu2017Deep} & 54.6 & 0.017 & 36.7 & 55.3 \\
			\rowcolor{mygray}
			DIM-IA~\cite{Xu2017Deep} & 39.9 & 0.015 & 36.6 & 49.5 \\
			DIM-MSR~\cite{Xu2017Deep} & 49.2 & 0.015 & 34.5 & 51.7 \\
			\hline
			\rowcolor{mygray}
			Ours-w/IA & 43.2 & 0.012 & 22.6 & 44.9\\
			\rowcolor{mygray}
			Ours-wo/IA & 53.8 & 0.015 & 33.1 & 52.2\\
			Ours-wo/MSR & 40.3 & 0.010 & 20.4 & 37.1\\
			Ours-Refine & 40.3 & 0.010 & 19.5 & 36.0\\
		    Ours (w/MSR) & 36.4 & 0.009 & 16.9 & 31.5\\
			\hline

	\end{tabular}}
	
\end{table}

\textbf{Analysis on MSR.~}
In what follows, we  evaluate the effectiveness of the Multi-Scale Refinement module compared to the Refinement module in DIM. Attention, all the following experiments are conducted on the Composition-1K dataset. We first use the refinement module of DIM to replace our MSR and denote it as Our-Refine. As shown in Tab.~\ref{tab:ablation_networks}, our MSR can bring improvements in all four metrics compared to the original Refinement module in DIM. In terms of the SAD and Grad, Ours (w/MSR) improves the Ours-Refine by 3.9 and 2.6, proving the necessity of multi-scale information for image matting.
We further validate the availability by replacing the original refinement module in DIM with MSR, which was indicated as DIM-MSR. 
Since the original refinement module in DIM  simply concatenated the image with the result of preceding stage to refine the alpha matte and neglects the multi-scale spatial information,  the result of DIM-MSR is deservedly better than the vanilla DIM, especially the metrics increased by 0.002 and 2.2 on MSE and Grad.
\subsection{Results on Real-World  images}

Fig.~\ref{fig:visual_real} shows how our algorithm performs on some internet images with a given trimap. All the results are acquired by our PIIAMatting that train on the Composition-1k\cite{Xu2017Deep} dataset. The results show that our method can produce delicate details around the boundary regions, such as the hairs and branches. Furthermore, due to the Dynamic Gaussian Modulation mechanism, our model can pay more attention to the highly uncertain regions and contribute to the excavation of structural details on varied domains.

\section{Conclusion and future work}
In this paper, we propose a Prior-Induced Information Alignment network for image matting. It utilizes the Dynamic Gaussian Modulation mechanism to regulate the pixel-wise response according to the prior information and employs the Information Alignment strategy to match and aggregate potential valuable information efficiently.
Finally, extensive experiments demonstrate that the proposed model can enhance the boundary details significantly and achieve a new SOTA performance on Alphamatting.com, Composition-1K, and Distinctions-646 dataset.

In the future, we will explore other efficient strategies to achieve high-quality alpha mattes.
The recent evolves, such as NAS\cite{Liu_2020_CVPR} and learnable parameters\cite{kendall2018multi}, can eliminate the limitations of the hand-designed strategy and adapt to the opacity variation more efficiently. It is also appealing to explore how to apply the dynamic Gaussian Modulation mechanism to video matting.

\ifCLASSOPTIONcaptionsoff
  \newpage
\fi

\bibliographystyle{IEEEtran}
\bibliography{bare_jrnl}
\vspace{-10mm}
\begin{IEEEbiography}[{\includegraphics[width=1in,height=1.25in,clip,keepaspectratio]{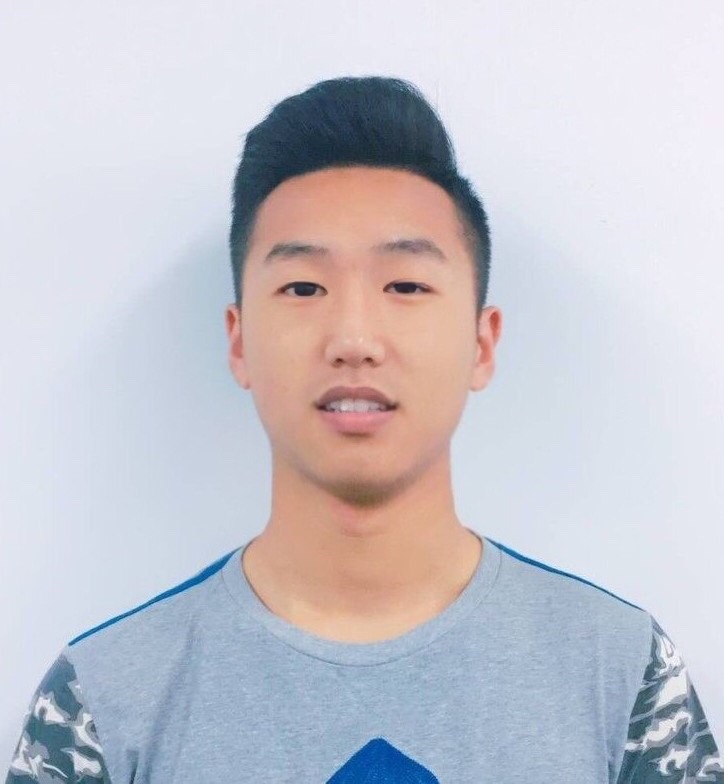}}]{Yuhao Liu}
received the B.Eng. degree from Zhengzhou University in 2019, and now he is a 
second-year master student majoring in computer science at Dalian University of 
Technology. His current research interests focus on low level computer vision 
problems and deep learning.
\end{IEEEbiography}
\vspace{-10mm}
\begin{IEEEbiography}[{\includegraphics[width=1in,height=1.25in,clip,keepaspectratio]{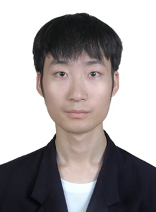}}]{Jiake Xie}
is the Computer Vision Algorithm Engineer in Hangzhou Winroad Holdings Ltd. He received his bachelor’s degree from Zhengzhou University of Light Industry. He specializes in computer vision related deep learning algorithms with research interests in object detection, semantic segmentation, salient object detection, image matting, and video matting. 
\end{IEEEbiography}
\vspace{-10mm}
\begin{IEEEbiography}[{\includegraphics[width=1in,height=1.25in,clip,keepaspectratio]{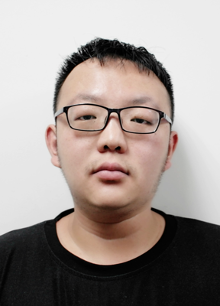}}]{Yu Qiao}
received his B.Eng from Dalian University of Technology in 2017, and takes a 
successive post-graduate and doctoral program in computer science. His research 
interest includes computer vision and image processing.
\end{IEEEbiography}

\begin{IEEEbiography}[{\includegraphics[width=1in,height=1.25in,clip,keepaspectratio]{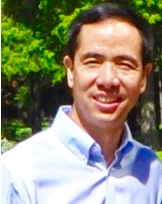}}]{Yong Tang}
	is the Vice President in Hangzhou Winroad Holdings Ltd, where he leads the Libai Artificial Intelligence Lab and the department of Human-Machine Engineering. He received his PHD degree from the Pennsylvania State University majoring in Computational Science and Civil Engineering. He obtained M.S. and B.E. degrees from Tsinghua University. He published articles in the world’s top magazines and conference journals and was awarded with the Best Paper Awards for multiple times. He also peer reviewed  articles for magazines in world-renowned professional fields. His research interests include computer vision, deep learning, text to speech synthesis, natural language processing, multi-objective optimization, and parallel computing.    
\end{IEEEbiography}
\vspace{-150mm}
\begin{IEEEbiography}[{\includegraphics[width=1in,height=1.25in,clip,keepaspectratio]{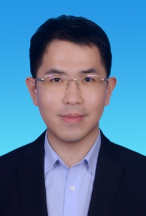}}]{Xin Yang}
is a Professor in the Department of Computer Science at Dalian University of Technology, China. Xin received his B.S. degree in Computer Science from Jilin University in 2007. From 2007 to June 2012, he was a joint Ph.D.
student in Zhejiang University and UC Davis for Graphics, and received his Ph.D. degree in July 2012. His research interests include computer graphics and robotic vision.
\end{IEEEbiography}
%








\end{document}